\documentclass[runningheads]{llncs}

\usepackage{eccv}

\usepackage{adjustbox}
\usepackage{tabularray}
\usepackage{eccvabbrv}

\usepackage{graphicx}
\usepackage{booktabs}

\usepackage[accsupp]{axessibility}  

\usepackage{import}
\usepackage{graphicx}

\usepackage{xspace} 

\usepackage{amsmath,amssymb,amsfonts}
\usepackage{bbm}
\usepackage{bm}

\usepackage{graphicx}

\usepackage{diagbox}
\usepackage[breaklinks,colorlinks]{hyperref}
\hypersetup{
    colorlinks=true, 
    linkcolor=blue,  
    citecolor=eccvblue,  
    filecolor=magenta, 
    urlcolor=blue    
}

\usepackage{enumitem}

\usepackage{color,soul}
\usepackage{xcolor}  

\usepackage{pifont}

\usepackage{tikz}
\usepackage{pgfplots}
\definecolor{lineBlue}{RGB}{57,106,177}
\definecolor{lineOrange}{RGB}{218,124,48}
\definecolor{lineGreen}{RGB}{62,150,81}
\definecolor{lineRed}{RGB}{204,37,41}
\definecolor{lineGray}{RGB}{83,81,84}
\definecolor{linePurple}{RGB}{107,76,154}
\definecolor{lineMaroon}{RGB}{146,36,40}
\definecolor{barBlue}{RGB}{114,147,203}
\definecolor{barOrange}{RGB}{225,151,76}
\definecolor{barGreen}{RGB}{132,186,91}
\definecolor{barRed}{RGB}{211,94,96}
\definecolor{barGray}{RGB}{128,133,133}
\definecolor{barPurple}{RGB}{144,103,167}
\definecolor{barMaroon}{RGB}{171,104,81}

\pgfplotsset{compat=1.18}
\usetikzlibrary{3d,decorations.text,shapes.arrows,positioning,fit,backgrounds}
\usetikzlibrary{intersections, pgfplots.fillbetween}

\usepackage{multirow}

\usetikzlibrary{arrows.meta}

\usepackage{algorithm}
\usepackage{algorithmic}

\subimport{./}{macros}

\usepackage{nicefrac}       

\usepackage{hyperref}

\renewcommand{\equationautorefname}{Eq.}
\renewcommand{\figureautorefname}{Fig.}

\newcommand{\Autoref}[1]{%
\begingroup%
\renewcommand{\equationautorefname}{Equation}%
\renewcommand{\figureautorefname}{Figure}%
\autoref{#1}%
\endgroup%
}

\usepackage{hyperref}

\usepackage{orcidlink}

\begin{document}

\title{Trainable Highly-expressive Activation Functions} 


\author{Irit Chelly$^*$\orcidlink{0009-0002-7713-7788} \and
Shahaf E. Finder$^*$\orcidlink{0000-0003-0254-1380} \and
Shira Ifergane\orcidlink{0009-0002-9682-1278} \and
Oren Freifeld\orcidlink{0000-0001-9816-9709}
}

\authorrunning{I.~Chelly et al.}

\institute{The Department of Computer Science, Ben-Gurion University of the Negev, Israel\\
\email{\{tohamy,finders,shiraif\}@post.bgu.ac.il, orenfr@cs.bgu.ac.il}}

\maketitle

\newcommand{\footnotestandalone}[1]{\let\thefootnote\relax\footnotetext{#1}}
\footnotestandalone{$^*$Equal contribution.}

\begin{abstract}
   
Nonlinear activation functions are pivotal to the success of deep neural nets, and choosing the appropriate activation function can significantly affect their performance. Most networks use fixed activation functions (\eg, ReLU, GELU, \etc), and this choice might limit their expressiveness. Furthermore, different layers may benefit from diverse activation functions. 
Consequently, there has been a growing interest in trainable activation functions. In this paper, we introduce DiTAC, a trainable highly-expressive activation function based on an efficient diffeomorphic transformation (called CPAB). Despite introducing only a negligible number of trainable parameters, DiTAC enhances model expressiveness and performance, often yielding substantial improvements. It also outperforms existing activation functions (regardless whether the latter are fixed or trainable) in tasks such as semantic segmentation, image generation, regression problems, and image classification. Our code is available at \url{https://github.com/BGU-CS-VIL/DiTAC}.

\end{abstract}
\section{Introduction}
\label{sec:intro}

\begin{figure}[t]
    \centering
    \setlength{\fboxsep}{0pt}
    \setlength{\fboxrule}{0.1pt}
    \captionsetup[sub]{font=small}
     \begin{subfigure}[b]{0.25\textwidth}
         \centering
         \framebox{\includegraphics[width=\textwidth]{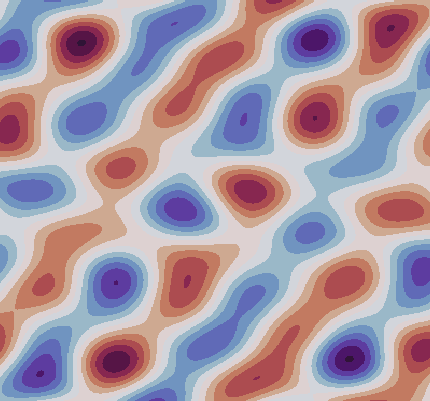}}
         \caption{Target function}
     \end{subfigure}
     \quad \quad
     \begin{subfigure}[b]{0.25\textwidth}
         \centering
         \framebox{\includegraphics[width=\textwidth]{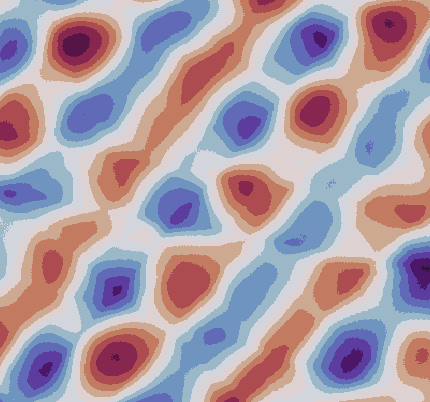}}
         \caption{MLP+DiTAC}
     \end{subfigure}
     \quad \quad
     \begin{subfigure}[b]{0.25\textwidth}
         \centering
          \framebox{\includegraphics[width=\textwidth]{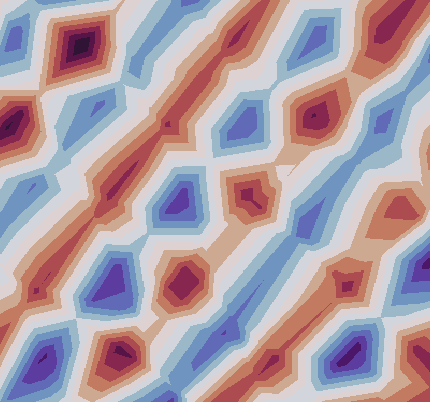}}
         \caption{MLP+PReLU}
     \end{subfigure}
        \caption{Regression-task results of reconstructing a two-dimensional function via a simple MLP, using 
        either DiTAC or (the runner-up) PReLU. Due to its expressiveness, DiTAC manages to fit a smooth function, yielding an evidently-better reconstruction.}
        \label{fig:2d_regression}
\end{figure}

Activation functions (AFs) play a major role in the success of deep neural nets, as they endow the latter with nonlinearity~\cite{Duch:1999:survey_transfer_functions,Prince:Book:2023:UDL}.
In fact, AFs are crucial to the ability of the networks to approximate almost arbitrarily-complex functions, to learn meaningful feature representations, and to achieve high predictive performance.
In addition to their nonlinearity, AFs have various characteristics which directly influence the performance of the network. 
Traditional AFs, such as the Logistic Sigmoid and Tanh Unit, map their input values into a small range, possibly causing the network gradients to become close to zero~\cite{Bengio:1994:learning_gradient_descent} and thus, impair training performance~\cite{Bishop:2006:pattern_recognition}. 
The Rectified Linear Unit (ReLU)~\cite{Nair:ICML:2010:relu} and its variants (\eg, LReLU~\cite{Maas:2013:LReLU} and PReLU~\cite{He:2015:PReLU}) partially solve this issue by mapping the input into an unbounded range in one or two directions.
Exponential AFs such as ELU~\cite{Clevert:2015:elu} inherit ReLU's benefits but also push the AF responses to zero-mean in order to improve performance~\cite{Clevert:2015:elu}. 

Generally, fixed AFs have limited nonlinearity (hence limited expressiveness) and impose a learning bias on the network. Consequently, adjusting them to varying problem types and data complexity is challenging. As a result, investigating AF design which enhances expressiveness and alleviates such a bias is an open field of research. Trainable activation functions (TAFs) such as PReLU~\cite{He:2015:PReLU}, Swish~\cite{Ramachandran:2017:Swish} and PELU~\cite{Trottier:2017:PELU} adjust the shape of a standard fixed AF by adding several learnable parameters. According to~\cite{Apicella:2021:survey_AFs_2}, this type of functions achieves a minor gain in expressiveness, as those TAFs tend to perform similarly to their base untrainable AFs.
A different AF approach is presented by the Maxout Unit~\cite{Goodfellow:2013:maxout}. Despite its remarkable improvement over ReLU in solving classification tasks~\cite{Goodfellow:2013:maxout}, the number of parameters in the Maxout layer increases with the number of the neurons in the network. 
The aforementioned limitations of existing AFs are part of the motivation behind our paper.\\

A \emph{diffeomorphism} is a differentiable invertible function with a differentiable inverse. 
In this paper we propose a \textbf{Di}ffeomorphism-based \textbf{T}rainable \textbf{Ac}tivation function (DiTAC), a differentiable parametric TAF based on highly-expressive and efficient diffeomorphisms (called CPAB~\cite{Freifeld:ICCV:2015:CPAB,Freifeld:PAMI:2017:CPAB}).
 DiTAC is highly expressive even though it adds only a negligible amount of trainable parameters. Comparing to existing TAFs which are restricted to learn either only a certain shape~\cite{He:2015:PReLU, Trottier:2017:PELU, Ramachandran:2017:Swish} or only convex functions~\cite{Goodfellow:2013:maxout}, DiTAC is capable of learning a variety of shapes (see~\autoref{fig:2d_regression} and \autoref{fig:cpab_act_phases}).
In~\cite{Dubey:2022:survey_AFs_1} it is shown that different AFs are suitable to different types of data and tasks, a fact motivating flexible TAF approaches like ours.
In particular, we show that DiTAC achieves significant improvements on various datasets, and various tasks such as semantic segmentation, image generation, image classification, and regression problems.\\

To summarize, \textbf{our contributions are as follows}: (1) 
To our knowledge, we are the first to propose using flexible diffeomorphisms within TAFs. 
(2) We present DiTAC, a novel highly-expressive AF that addresses issues of existing TAFs, and can be easily used in any model architecture. (3) We show that DiTAC outperforms existing AFs and TAFs on various tasks and datasets.

\section{Related Work}
\label{sec:relatedwork}

\subsection{Activation Functions}
AFs are a key component of artificial neural networks. They introduce nonlinearity between the linear operations of the network; \ie, without them the network is simply a linear function.
In the early days of neural networks, the most common AFs were the Logistic Sigmoid and the Tanh Unit~\cite{Lecun:IEEE:1998:lenet}, both are smooth and non-decreasing functions. However, due to their bounded response, the network gradients become generally small and quickly approach zero~\cite{Bengio:1994:learning_gradient_descent}, causing the notorious vanishing-gradient problem.
The Rectified Linear Unit (ReLU)~\cite{Nair:ICML:2010:relu} function attempts to solve this issue by mapping its positive-values input to an unbounded range. ReLU is widely used due to its computational efficiency and improved performance~\cite{Glorot:2011:deep_rec_NN, Nair:ICML:2010:relu, Maas:2013:LReLU}. To better solve the vanishing-gradient issue, other Rectified AFs such as LReLU~\cite{Maas:2013:LReLU} and PReLU~\cite{He:2015:PReLU} utilize the negative input values by returning a linear function with a fixed/trainable slope, and hence become unbounded in both directions.
Nevertheless, adjusting/learning the right slope may be more suitable for certain tasks than others, or lead to overfitting~\cite{Dubey:2022:survey_AFs_1}.
Differentiable alternatives to ReLU have also been developed, such as ELU~\cite{Clevert:2015:elu}, defined as $\mathrm{ELU}(x)=\alpha(e^x-1)$ for $x\leq 0$, and as the identity function otherwise.
Extensions to ELU were proposed~\cite{Klambauer:2017:SELU, Trottier:2017:PELU, Barron:2017:CELU, Cheng:2020:PDELU}, where in~\cite{Cheng:2020:PDELU}, PDELU learns a zero-mean AF response, allowing steepest descent of gradients and improving training speed and performance~\cite{Clevert:2015:elu}.
Other differentiable functions are Softplus~\cite{Dugas:2000:softplus},
a smooth approximation of ReLU~\cite{Apicella:2021:survey_AFs_2},
and ErfReLU~\cite{ErfReLU:2023:rajanand}, a combination of ReLU and the error function (\ie, the cumulative distribution function of a standard normal distribution).
In~\cite{PINNS:2020:jagtap, PINNS_local:2020:jagtap, KNNs:2022:jagtap}, the authors propose TAFs for physics-oriented networks.
A recent popular AF called GELU~\cite{Hendrycks:2016:gelu} utilizes the error function to weight its input. 
In~\cite{Ramachandran:2017:Swish}, 
the authors conducted an automated reinforcement learning-based search to find new AFs, and proposed Swish, defined as $\mathrm{Swish}(x)=x\cdot \mathrm{Sigmoid}(\beta x)$ where $\beta$ is a trainable parameter. Motivated by those search techniques, \cite{Misra:BMVC:2019:mish} introduced Mish, a combination of the tanh and softplus functions, and~\cite{ASH:2022:lee} introduced ASH, a generalized form of Swish.

The aforementioned AFs are either fixed AFs or based on a fixed-shape function, allowing its shape to be adapted by learning several parameters. While adaptable, their expressiveness is still limited as they implicitly  inherit the 
fixed structural form of the AF they were adapted from.
\begin{figure}[hbt!]
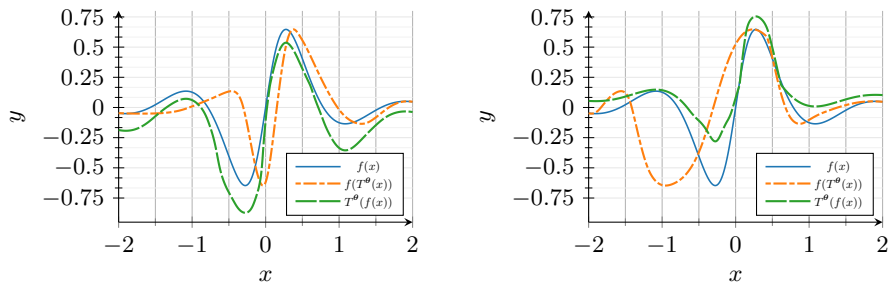
  
    \centering
    \begin{subfigure}[b]{0.45\textwidth}
    \input{./raw/cpab_diff_axes}
    \end{subfigure}
    \quad
    \quad
    \begin{subfigure}[b]{0.45\textwidth}
    \input{./raw/cpab_diff_axes_2}
    \end{subfigure}
    \caption{The CPAB transformation effect when it is applied on each axis. In each panel we use a different $T^\btheta$. When the $x$ axis is transformed (\ie, $f(T^\btheta(x))$), the intensity values are unchanged and only the $x$ values are shifted. When the $y$ axis is transformed ($T^\btheta(f(x))$), the intensity is changed while peaks and valleys' locations are kept.}
    \label{fig:cpab_axes}
\end{figure}

Some TAFs avoid this at the cost of more trainable parameters. Unfortunately, however, the number of their added trainable parameters grows with the number of neurons in the network. For instance, 
the Maxout Unit~\cite{Goodfellow:2013:maxout}, followed by its stochastic version Probabilistic Maxout~\cite{Springenberg:2013:prob_maxout}, computes the maximal value of a set of $k$ trainable linear functions of the same input, leading to a significant increase in the number of total parameters in the network. The ACON family~\cite{Ma:2021:ACON} partially mitigates this issue by approximating the general Maxout family. However, in its major proposed variants, the number of added parameters still grows with the number of channels in the network. 
Another recent TAF that increases the number of parameters is the DY-ReLU~\cite{Chen:2020:DY_ReLU}, which forms a piecewise function whose parameters are generated by a hyper function over all input elements.
Our work, in contrast to all those works, offers a more expressive nonlinear differentiable TAF using only a negligible amount of added trainable parameters.

\subsection{CPAB transformations in Deep Learning}
CPAB transformations, proposed by Freifeld~\etal~\cite{Freifeld:ICCV:2015:CPAB,Freifeld:PAMI:2017:CPAB}, 
are efficient and highly-expressive parametric diffeomorphisms. They are called CPAB, short for CPA-Based,  
as they are based on Continuous Piecewise-Affine (CPA) velocity fields (as we will explain later). 
Since their inception, those transformations found many applications in DL (\eg,~\cite{Hauberg:AISTATS:2016:DA,Detlefsen:CVPR:2018:deepdiffeomorphic,Skafte:NIPS:2019:explicit,Shapira:NeurIPS:2019:DTAN,Kaufman:ICIP:2021:Cyclic,Shacht:2021:single,Schwobel:2022:UAI:pstn,Martinez:ICML:2022:closed,Neifar:2022l:everaging,Kryeem:ICCV:2023:personalized,Shapira:ICML:2023:regularizationfreedtan,Wang:AAAI:2024:Animation})
. 
A main difference between how all these works used CPAB transformations 
and our usage of them, is that in those works the CPAB transformations were always applied
to the domain of the signals of interest (either the spatial domain in 2D images or the time domain in time series), 
typically by incorporating them in a Spatial Transformer Net (STN)~\cite{Jaderberg:NIPS:2015:spatial} or a Temporal Transformer Net (TTN)~\cite{Shapira:NeurIPS:2019:DTAN},  
while we apply them (elementwise) to the \emph{range} of feature maps 
(as an aside, among other things it also means that we do not need to preform grid resampling, a mandatory step in STNs/TTNs). \Autoref{fig:cpab_axes} illustrates this difference. 
In particular, we are the first to use CPAB transformations (or any other highly-expressive family of diffeomorphisms for that matters) for building TAFs.

\section{Method}\label{sec:method}

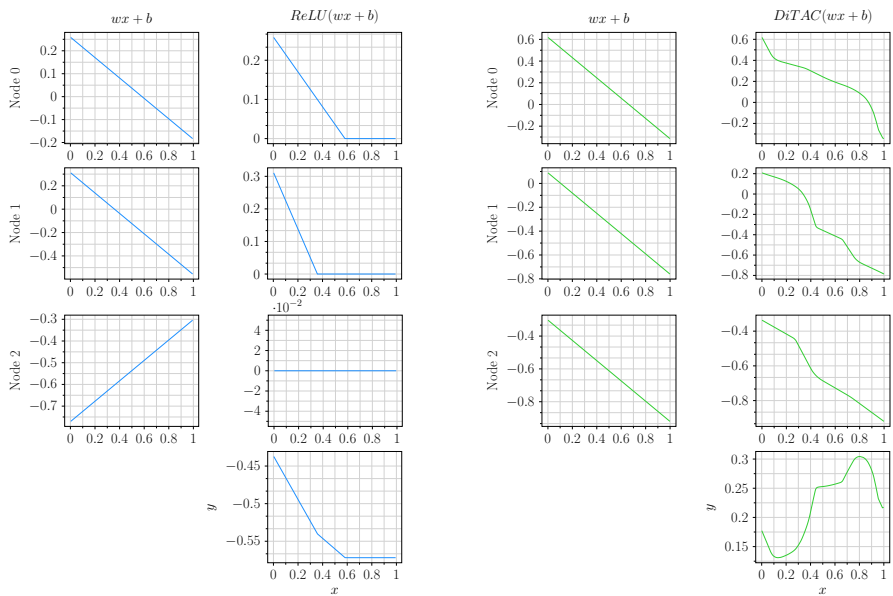
\begin{figure}[hbt!]
    \renewcommand{\arraystretch}{0.0}
    \centering
    \huge
    \begin{tabular}{p{3cm}p{3cm}p{0.05cm}p{3cm}p{3cm}}
    \scalebox{0.26}{
\begin{tikzpicture}

\definecolor{darkgray176}{RGB}{210,210,210}
\definecolor{dodgerblue}{RGB}{30,144,255}

\begin{axis}[
tick align=outside,
tick pos=left,
x grid style={darkgray176},
xmin=-0.0467265516519547, xmax=1.04324283897877,
xtick style={color=black},
y grid style={darkgray176},
ymin=-0.205006239563227, ymax=0.280196828395128,
ytick style={color=black},
ylabel={Node 0},
y label style={at={(axis description cs:-0.3,.5)},anchor=south},
title={\(\displaystyle wx + b\)},
grid=both,
grid style={line width=.1pt, draw=gray!10},
major grid style={line width=.2pt,draw=gray!20},
minor x tick num=1,
minor y tick num=1,
]
\addplot [semithick, dodgerblue]
table {%
0.00281751155853271 0.25814214348793
0.0133479833602905 0.253454476594925
0.0195685029029846 0.250685393810272
0.0220627784729004 0.249575063586235
0.0422413945198059 0.240592494606972
0.0460957884788513 0.238876700401306
0.0549092292785645 0.234953373670578
0.0579379200935364 0.23360513150692
0.059905469417572 0.232729271054268
0.0615454912185669 0.231999218463898
0.0631937384605408 0.231265500187874
0.0633630752563477 0.231190115213394
0.0704537034034729 0.22803370654583
0.0778093934059143 0.224759295582771
0.0779090523719788 0.224714934825897
0.0954298973083496 0.216915473341942
0.0959219336509705 0.216696441173553
0.0961606502532959 0.21659018099308
0.108971536159515 0.210887372493744
0.11107337474823 0.209951728582382
0.115571260452271 0.207949489355087
0.116032361984253 0.207744225859642
0.119871616363525 0.206035166978836
0.12512594461441 0.203696191310883
0.125698208808899 0.203441441059113
0.12690669298172 0.202903494238853
0.128084480762482 0.202379196882248
0.128110289573669 0.202367708086967
0.137459397315979 0.198205918073654
0.146419286727905 0.194217398762703
0.154882907867432 0.190449789166451
0.156893789768219 0.189554646611214
0.162056267261505 0.187256544828415
0.166283965110779 0.185374572873116
0.177438855171204 0.1804089397192
0.179214715957642 0.179618418216705
0.180649042129517 0.17897991836071
0.190518379211426 0.174586564302444
0.195219457149506 0.172493860125542
0.208765506744385 0.166463792324066
0.213927268981934 0.164166018366814
0.219827175140381 0.161539658904076
0.2249675989151 0.159251391887665
0.226844787597656 0.158415749669075
0.227535843849182 0.158108130097389
0.22871208190918 0.157584518194199
0.231405019760132 0.156385749578476
0.233007431030273 0.155672430992126
0.234380543231964 0.155061185359955
0.246609747409821 0.149617314338684
0.25255024433136 0.146972894668579
0.257454514503479 0.14478974044323
0.269976437091827 0.13921557366848
0.275099039077759 0.136935234069824
0.276986598968506 0.136094972491264
0.280861973762512 0.13436983525753
0.281703054904938 0.133995428681374
0.28642213344574 0.13189472258091
0.288284957408905 0.131065472960472
0.299872934818268 0.125907063484192
0.30469411611557 0.123760893940926
0.311670541763306 0.120655313134193
0.31285148859024 0.120129615068436
0.313522338867188 0.11983098089695
0.326532125473022 0.11403963714838
0.326902806758881 0.1138746291399
0.329703271389008 0.112627990543842
0.337348163127899 0.10922484844923
0.344143927097321 0.106199696660042
0.348860740661621 0.104099988937378
0.350317060947418 0.103451706469059
0.356289505958557 0.100793056190014
0.358660161495209 0.099737748503685
0.358758687973022 0.0996938869357109
0.367152631282806 0.0959573015570641
0.367277324199677 0.095901794731617
0.379245758056641 0.0905740112066269
0.386250913143158 0.0874556452035904
0.386266887187958 0.0874485373497009
0.394435822963715 0.0838121101260185
0.39991295337677 0.0813739448785782
0.409272015094757 0.0772077366709709
0.414407432079315 0.074921689927578
0.415173947811127 0.0745804682374001
0.415933728218079 0.0742422565817833
0.42482852935791 0.0702827051281929
0.428329586982727 0.068724200129509
0.429179012775421 0.0683460757136345
0.434281051158905 0.0660748854279518
0.437131762504578 0.0648058876395226
0.469995975494385 0.0501762852072716
0.472166657447815 0.049210000783205
0.472257614135742 0.049169510602951
0.480677783489227 0.0454212464392185
0.483239233493805 0.0442810095846653
0.483724355697632 0.0440650545060635
0.489208221435547 0.0416238978505135
0.493338048458099 0.0397854931652546
0.495270967483521 0.0389250479638577
0.496253669261932 0.0384875945746899
0.50495058298111 0.0346161387860775
0.520527064800262 0.0276822224259377
0.522519409656525 0.0267953239381313
0.532208681106567 0.0224821157753468
0.533660054206848 0.0218360330909491
0.537782311439514 0.0200009979307652
0.540998578071594 0.0185692682862282
0.546118080615997 0.0162903051823378
0.548947513103485 0.0150307761505246
0.552013754844666 0.0136658288538456
0.553114831447601 0.0131756821647286
0.555565059185028 0.0120849553495646
0.559368371963501 0.0103918993845582
0.560492038726807 0.00989169627428055
0.560539960861206 0.00987036339938641
0.565063297748566 0.00785678718239069
0.571642577648163 0.00492800120264292
0.572652280330658 0.00447852909564972
0.580806016921997 0.000848869152832776
0.58209490776062 0.000275115628028288
0.591409504413605 -0.00387130444869399
0.600167989730835 -0.00777016999199986
0.60407030582428 -0.00950729753822088
0.60964959859848 -0.0119909355416894
0.61403900384903 -0.0139448922127485
0.614174604415894 -0.0140052558854222
0.615768313407898 -0.0147147001698613
0.622929513454437 -0.0179025288671255
0.632022619247437 -0.021950351074338
0.63678252696991 -0.0240692384541035
0.639056444168091 -0.025081479921937
0.654363334178925 -0.031895387917757
0.663265824317932 -0.0358583554625511
0.666122615337372 -0.0371300652623177
0.668011546134949 -0.0379709266126156
0.669496238231659 -0.0386318452656269
0.688584446907043 -0.0471290163695812
0.689998388290405 -0.0477584339678288
0.697598814964294 -0.0511417873203754
0.69837874174118 -0.0514889732003212
0.702065110206604 -0.0531299710273743
0.70631605386734 -0.0550222918391228
0.706624805927277 -0.0551597326993942
0.724156439304352 -0.062963992357254
0.728997051715851 -0.0651188045740128
0.738590002059937 -0.069389134645462
0.749205708503723 -0.0741147473454475
0.749558568000793 -0.0742718279361725
0.764341175556183 -0.0808523446321487
0.76554661989212 -0.0813889503479004
0.76944637298584 -0.0831249430775642
0.769515991210938 -0.0831559300422668
0.772613525390625 -0.0845348089933395
0.773476898670197 -0.0849191397428513
0.779029428958893 -0.0873908624053001
0.779123663902283 -0.0874328166246414
0.779822468757629 -0.0877438858151436
0.781309485435486 -0.0884058400988579
0.783272564411163 -0.0892797112464905
0.783661901950836 -0.0894530266523361
0.7949178814888 -0.0944636538624763
0.795761823654175 -0.0948393419384956
0.803122997283936 -0.0981161892414093
0.811211287975311 -0.101716712117195
0.815272748470306 -0.103524684906006
0.820028066635132 -0.105641528964043
0.826399981975555 -0.108478009700775
0.833688616752625 -0.111722566187382
0.841201722621918 -0.115067042410374
0.848510444164276 -0.118320539593697
0.853233754634857 -0.120423138141632
0.85508781671524 -0.121248483657837
0.857133150100708 -0.122158966958523
0.85740339756012 -0.122279264032841
0.857532501220703 -0.122336737811565
0.859283447265625 -0.123116180300713
0.877039909362793 -0.13102051615715
0.885247409343719 -0.134674116969109
0.89163339138031 -0.137516856193542
0.892618536949158 -0.137955397367477
0.895047187805176 -0.139036506414413
0.895302593708038 -0.139150202274323
0.906564593315125 -0.144163519144058
0.90956848859787 -0.145500704646111
0.914022088050842 -0.147483244538307
0.91419130563736 -0.147558569908142
0.915676832199097 -0.148219853639603
0.915914237499237 -0.148325532674789
0.942246556282043 -0.160047441720963
0.95756596326828 -0.166866928339005
0.962188065052032 -0.16892446577549
0.965132892131805 -0.170235365629196
0.965466856956482 -0.170384034514427
0.968339085578918 -0.171662613749504
0.984445214271545 -0.178832307457924
0.985623776912689 -0.179356947541237
0.98764044046402 -0.18025466799736
0.988257527351379 -0.180529370903969
0.988281011581421 -0.180539816617966
0.99369877576828 -0.182951554656029
};
\end{axis}

\end{tikzpicture}
    } & 
    \scalebox{0.26}{
\begin{tikzpicture}

\definecolor{darkgray176}{RGB}{210,210,210}
\definecolor{dodgerblue}{RGB}{30,144,255}

\begin{axis}[
tick align=outside,
tick pos=left,
x grid style={darkgray176},
xmin=-0.0467265516519547, xmax=1.04324283897877,
xtick style={color=black},
y grid style={darkgray176},
ymin=-0.0129071071743965, ymax=0.271049250662327,
ytick style={color=black},
yticklabel style={font=\huge},
xticklabel style={font=\huge},
tick label style={/pgf/number format/fixed},
title={\(\displaystyle ReLU(wx+b)\)},
grid=both,
grid style={line width=.1pt, draw=gray!10},
major grid style={line width=.2pt,draw=gray!20},
minor x tick num=1,
minor y tick num=2,
ytick={0,0.1,0.2,0.3,0.4},
]
\addplot [semithick, dodgerblue]
table {%
0.00281751155853271 0.25814214348793
0.0133479833602905 0.253454476594925
0.0195685029029846 0.250685393810272
0.0220627784729004 0.249575063586235
0.0422413945198059 0.240592494606972
0.0460957884788513 0.238876700401306
0.0549092292785645 0.234953373670578
0.0579379200935364 0.23360513150692
0.059905469417572 0.232729271054268
0.0615454912185669 0.231999218463898
0.0631937384605408 0.231265500187874
0.0633630752563477 0.231190115213394
0.0704537034034729 0.22803370654583
0.0778093934059143 0.224759295582771
0.0779090523719788 0.224714934825897
0.0954298973083496 0.216915473341942
0.0959219336509705 0.216696441173553
0.0961606502532959 0.21659018099308
0.108971536159515 0.210887372493744
0.11107337474823 0.209951728582382
0.115571260452271 0.207949489355087
0.116032361984253 0.207744225859642
0.119871616363525 0.206035166978836
0.12512594461441 0.203696191310883
0.125698208808899 0.203441441059113
0.12690669298172 0.202903494238853
0.128084480762482 0.202379196882248
0.128110289573669 0.202367708086967
0.137459397315979 0.198205918073654
0.146419286727905 0.194217398762703
0.154882907867432 0.190449789166451
0.156893789768219 0.189554646611214
0.162056267261505 0.187256544828415
0.166283965110779 0.185374572873116
0.177438855171204 0.1804089397192
0.179214715957642 0.179618418216705
0.180649042129517 0.17897991836071
0.190518379211426 0.174586564302444
0.195219457149506 0.172493860125542
0.208765506744385 0.166463792324066
0.213927268981934 0.164166018366814
0.219827175140381 0.161539658904076
0.2249675989151 0.159251391887665
0.226844787597656 0.158415749669075
0.227535843849182 0.158108130097389
0.22871208190918 0.157584518194199
0.231405019760132 0.156385749578476
0.233007431030273 0.155672430992126
0.234380543231964 0.155061185359955
0.246609747409821 0.149617314338684
0.25255024433136 0.146972894668579
0.257454514503479 0.14478974044323
0.269976437091827 0.13921557366848
0.275099039077759 0.136935234069824
0.276986598968506 0.136094972491264
0.280861973762512 0.13436983525753
0.281703054904938 0.133995428681374
0.28642213344574 0.13189472258091
0.288284957408905 0.131065472960472
0.299872934818268 0.125907063484192
0.30469411611557 0.123760893940926
0.311670541763306 0.120655313134193
0.31285148859024 0.120129615068436
0.313522338867188 0.11983098089695
0.326532125473022 0.11403963714838
0.326902806758881 0.1138746291399
0.329703271389008 0.112627990543842
0.337348163127899 0.10922484844923
0.344143927097321 0.106199696660042
0.348860740661621 0.104099988937378
0.350317060947418 0.103451706469059
0.356289505958557 0.100793056190014
0.358660161495209 0.099737748503685
0.358758687973022 0.0996938869357109
0.367152631282806 0.0959573015570641
0.367277324199677 0.095901794731617
0.379245758056641 0.0905740112066269
0.386250913143158 0.0874556452035904
0.386266887187958 0.0874485373497009
0.394435822963715 0.0838121101260185
0.39991295337677 0.0813739448785782
0.409272015094757 0.0772077366709709
0.414407432079315 0.074921689927578
0.415173947811127 0.0745804682374001
0.415933728218079 0.0742422565817833
0.42482852935791 0.0702827051281929
0.428329586982727 0.068724200129509
0.429179012775421 0.0683460757136345
0.434281051158905 0.0660748854279518
0.437131762504578 0.0648058876395226
0.469995975494385 0.0501762852072716
0.472166657447815 0.049210000783205
0.472257614135742 0.049169510602951
0.480677783489227 0.0454212464392185
0.483239233493805 0.0442810095846653
0.483724355697632 0.0440650545060635
0.489208221435547 0.0416238978505135
0.493338048458099 0.0397854931652546
0.495270967483521 0.0389250479638577
0.496253669261932 0.0384875945746899
0.50495058298111 0.0346161387860775
0.520527064800262 0.0276822224259377
0.522519409656525 0.0267953239381313
0.532208681106567 0.0224821157753468
0.533660054206848 0.0218360330909491
0.537782311439514 0.0200009979307652
0.540998578071594 0.0185692682862282
0.546118080615997 0.0162903051823378
0.548947513103485 0.0150307761505246
0.552013754844666 0.0136658288538456
0.553114831447601 0.0131756821647286
0.555565059185028 0.0120849553495646
0.559368371963501 0.0103918993845582
0.560492038726807 0.00989169627428055
0.560539960861206 0.00987036339938641
0.565063297748566 0.00785678718239069
0.571642577648163 0.00492800120264292
0.572652280330658 0.00447852909564972
0.580806016921997 0.000848869152832776
0.58209490776062 0.000275115628028288
0.591409504413605 0
0.600167989730835 0
0.60407030582428 0
0.60964959859848 0
0.61403900384903 0
0.614174604415894 0
0.615768313407898 0
0.622929513454437 0
0.632022619247437 0
0.63678252696991 0
0.639056444168091 0
0.654363334178925 0
0.663265824317932 0
0.666122615337372 0
0.668011546134949 0
0.669496238231659 0
0.688584446907043 0
0.689998388290405 0
0.697598814964294 0
0.69837874174118 0
0.702065110206604 0
0.70631605386734 0
0.706624805927277 0
0.724156439304352 0
0.728997051715851 0
0.738590002059937 0
0.749205708503723 0
0.749558568000793 0
0.764341175556183 0
0.76554661989212 0
0.76944637298584 0
0.769515991210938 0
0.772613525390625 0
0.773476898670197 0
0.779029428958893 0
0.779123663902283 0
0.779822468757629 0
0.781309485435486 0
0.783272564411163 0
0.783661901950836 0
0.7949178814888 0
0.795761823654175 0
0.803122997283936 0
0.811211287975311 0
0.815272748470306 0
0.820028066635132 0
0.826399981975555 0
0.833688616752625 0
0.841201722621918 0
0.848510444164276 0
0.853233754634857 0
0.85508781671524 0
0.857133150100708 0
0.85740339756012 0
0.857532501220703 0
0.859283447265625 0
0.877039909362793 0
0.885247409343719 0
0.89163339138031 0
0.892618536949158 0
0.895047187805176 0
0.895302593708038 0
0.906564593315125 0
0.90956848859787 0
0.914022088050842 0
0.91419130563736 0
0.915676832199097 0
0.915914237499237 0
0.942246556282043 0
0.95756596326828 0
0.962188065052032 0
0.965132892131805 0
0.965466856956482 0
0.968339085578918 0
0.984445214271545 0
0.985623776912689 0
0.98764044046402 0
0.988257527351379 0
0.988281011581421 0
0.99369877576828 0
};
\end{axis}

\end{tikzpicture}
    } & 
    &
    \scalebox{0.26}{
\begin{tikzpicture}

\definecolor{darkgray176}{RGB}{210,210,210}
\definecolor{limegreen}{RGB}{50,205,50}

\begin{axis}[
tick align=outside,
tick pos=left,
x grid style={darkgray176},
xmin=-0.0489364236593247, xmax=1.04846346080303,
xtick style={color=black},
y grid style={darkgray176},
ymin=-0.361241579055786, ymax=0.664482593536377,
ytick style={color=black},
title={\(wx+b\)},
ylabel={Node 0},
y label style={at={(axis description cs:-0.3,.5)},anchor=south},
grid=both,
grid style={line width=.1pt, draw=gray!10},
major grid style={line width=.2pt,draw=gray!20},
minor x tick num=1,
minor y tick num=1,
]
\addplot [semithick, limegreen]
table {%
0.000945389270782471 0.61785876750946
0.00184035301208496 0.617022275924683
0.00233417749404907 0.616560697555542
0.0026053786277771 0.616307199001312
0.0101443529129028 0.609260618686676
0.0104448199272156 0.608979761600494
0.0107607841491699 0.608684480190277
0.026776909828186 0.593714416027069
0.0273440480232239 0.593184292316437
0.0350483655929565 0.585983216762543
0.0417994856834412 0.579672992229462
0.0420104265213013 0.579475879669189
0.0446122884750366 0.577043950557709
0.0494936108589172 0.572481453418732
0.0495138764381409 0.572462499141693
0.0672390460968018 0.555895030498505
0.0707660913467407 0.552598357200623
0.0725078582763672 0.550970315933228
0.0744094252586365 0.54919296503067
0.0861891508102417 0.538182616233826
0.0869072675704956 0.53751140832901
0.0916629433631897 0.53306633234024
0.0933713912963867 0.531469464302063
0.0958954095840454 0.529110312461853
0.104392945766449 0.521167814731598
0.106009840965271 0.519656479358673
0.113933503627777 0.512250363826752
0.116077423095703 0.510246455669403
0.122549533843994 0.504197061061859
0.126630306243896 0.500382840633392
0.131201565265656 0.496110141277313
0.134268462657928 0.49324357509613
0.134857833385468 0.492692679166794
0.155025660991669 0.473842114210129
0.157626271247864 0.471411347389221
0.161516785621643 0.467774957418442
0.167969763278961 0.46174344420433
0.170785903930664 0.459111243486404
0.178690910339355 0.451722532510757
0.179924249649048 0.450569748878479
0.187150895595551 0.443815112113953
0.188112199306488 0.442916601896286
0.199634730815887 0.432146638631821
0.200894117355347 0.430969506502151
0.207908391952515 0.424413353204727
0.21448540687561 0.418265908956528
0.220963537693024 0.412210911512375
0.224007248878479 0.409365981817245
0.224759876728058 0.408662527799606
0.233896613121033 0.400122553110123
0.233974039554596 0.400050163269043
0.235349953174591 0.398764133453369
0.246053874492645 0.388759315013885
0.248235702514648 0.38672000169754
0.251342713832855 0.383815914392471
0.251824975013733 0.383365154266357
0.253204643726349 0.382075607776642
0.262514233589172 0.373374044895172
0.274229407310486 0.362424045801163
0.279606938362122 0.357397735118866
0.280423641204834 0.356634378433228
0.282334983348846 0.354847878217697
0.293261647224426 0.344634890556335
0.293978571891785 0.343964785337448
0.300131261348724 0.338213950395584
0.303768575191498 0.334814220666885
0.3211470246315 0.318570822477341
0.325471639633179 0.31452864408493
0.330861032009125 0.309491276741028
0.338974952697754 0.301907300949097
0.343935370445251 0.297270864248276
0.353755593299866 0.288092046976089
0.363615155220032 0.278876453638077
0.370469748973846 0.27246955037117
0.375653862953186 0.267624050378799
0.377780616283417 0.265636205673218
0.382170855998993 0.261532694101334
0.388573229312897 0.255548506975174
0.394622147083282 0.249894648790359
0.40790730714798 0.237477198243141
0.416023671627045 0.229890957474709
0.417597591876984 0.228419825434685
0.424522936344147 0.221946805715561
0.42478358745575 0.221703186631203
0.426332354545593 0.220255568623543
0.429543614387512 0.217254057526588
0.437937796115875 0.209408134222031
0.441768884658813 0.205827265977859
0.445889055728912 0.20197619497776
0.448626041412354 0.199417978525162
0.453730165958405 0.194647222757339
0.45561146736145 0.192888796329498
0.467173635959625 0.182081803679466
0.471240162849426 0.178280875086784
0.473981916904449 0.175718203186989
0.477888643741608 0.172066628932953
0.482316017150879 0.167928427457809
0.482877910137177 0.167403236031532
0.483318150043488 0.166991755366325
0.483602643013 0.166725844144821
0.493298947811127 0.157662838697433
0.496066510677338 0.155076041817665
0.508895456790924 0.14308500289917
0.512529194355011 0.139688596129417
0.514082670211792 0.138236597180367
0.517603933811188 0.134945318102837
0.523982226848602 0.128983616828918
0.527999937534332 0.125228315591812
0.532123267650604 0.121374301612377
0.539165675640106 0.114791862666607
0.541703701019287 0.112419605255127
0.552089393138885 0.102712243795395
0.572090744972229 0.0840172618627548
0.572779774665833 0.0833732336759567
0.574827432632446 0.0814593210816383
0.578090310096741 0.0784095525741577
0.578102648258209 0.0783980190753937
0.583341836929321 0.0735010281205177
0.584954082965851 0.0719940811395645
0.586356401443481 0.0706833526492119
0.588969588279724 0.0682408437132835
0.592191576957703 0.0652292966842651
0.594320952892303 0.0632390007376671
0.596706688404083 0.0610090866684914
0.60359114408493 0.0545742847025394
0.615020751953125 0.0438911914825439
0.616230964660645 0.0427600219845772
0.629793763160706 0.0300830639898777
0.631997585296631 0.0280231833457947
0.632766962051392 0.0273040570318699
0.633415281772614 0.0266980826854706
0.636320650577545 0.0239824745804071
0.64380294084549 0.0169888846576214
0.646667957305908 0.0143109941855073
0.649064838886261 0.0120706623420119
0.650434970855713 0.0107900192961097
0.659766912460327 0.00206758524291217
0.662503600120544 -0.000490358041133732
0.670535087585449 -0.00799727626144886
0.681240022182465 -0.0180030278861523
0.681955397129059 -0.018671678379178
0.690125644207001 -0.0263082925230265
0.691943168640137 -0.0280071068555117
0.700154781341553 -0.0356823839247227
0.701727032661438 -0.0371519476175308
0.714230895042419 -0.0488391295075417
0.715877711772919 -0.0503783859312534
0.720097780227661 -0.054322823882103
0.721782982349396 -0.0558979585766792
0.722362816333771 -0.0564399212598801
0.724339962005615 -0.0582879334688187
0.725047469139099 -0.0589492283761501
0.728903353214264 -0.062553271651268
0.729076385498047 -0.0627150014042854
0.732455909252167 -0.0658737942576408
0.736843228340149 -0.0699745565652847
0.739347815513611 -0.0723155587911606
0.74403327703476 -0.0766949951648712
0.749413132667542 -0.0817234739661217
0.750742435455322 -0.0829659551382065
0.756429255008698 -0.0882813408970833
0.756718814373016 -0.0885519906878471
0.759423971176147 -0.0910804644227028
0.764318466186523 -0.0956552773714066
0.771147429943085 -0.102038212120533
0.774811029434204 -0.105462528765202
0.784129559993744 -0.114172428846359
0.790473163127899 -0.12010170519352
0.790486514568329 -0.120114184916019
0.792000591754913 -0.121529370546341
0.799741685390472 -0.128764852881432
0.800864815711975 -0.129814639687538
0.804708778858185 -0.133407533168793
0.805868983268738 -0.134491965174675
0.830307900905609 -0.157334670424461
0.83801543712616 -0.1645388007164
0.847992241382599 -0.173863977193832
0.852572441101074 -0.17814502120018
0.866361677646637 -0.191033631563187
0.871176600456238 -0.195534065365791
0.874660193920135 -0.19879013299942
0.884599268436432 -0.208080038428307
0.88707435131073 -0.210393473505974
0.889352321624756 -0.212522655725479
0.890990674495697 -0.214054003357887
0.891286075115204 -0.214330106973648
0.90108335018158 -0.223487481474876
0.906848967075348 -0.228876531124115
0.906959772109985 -0.228980094194412
0.909624993801117 -0.231471240520477
0.914373815059662 -0.235909894108772
0.916529655456543 -0.237924933433533
0.917276442050934 -0.238622933626175
0.933907568454742 -0.254167824983597
0.936382114887238 -0.256480753421783
0.94158673286438 -0.261345416307449
0.947788000106812 -0.267141669988632
0.954951107501984 -0.27383691072464
0.988680303096771 -0.305363118648529
0.998581647872925 -0.31461775302887
};
\end{axis}

\end{tikzpicture}
    } & 
    \scalebox{0.26}{
\begin{tikzpicture}

\definecolor{darkgray176}{RGB}{210,210,210}
\definecolor{limegreen}{RGB}{50,205,50}

\begin{axis}[
tick align=outside,
tick pos=left,
x grid style={darkgray176},
xmin=-0.0489364236593247, xmax=1.04846346080303,
xtick style={color=black},
y grid style={darkgray176},
ymin=-0.392396768927574, ymax=0.665966174006462,
ytick style={color=black},
title={\(\displaystyle DiTAC(wx+b)\)},
grid=both,
grid style={line width=.1pt, draw=gray!10},
major grid style={line width=.2pt,draw=gray!20},
minor x tick num=1,
minor y tick num=1,
]
\addplot [semithick, limegreen]
table {%
0.000945389270782471 0.61785876750946
0.00184035301208496 0.615909099578857
0.00233417749404907 0.614833116531372
0.0026053786277771 0.614242315292358
0.0101443529129028 0.597817659378052
0.0104448199272156 0.597162961959839
0.0107607841491699 0.596475005149841
0.026776909828186 0.561581969261169
0.0273440480232239 0.560346126556396
0.0350483655929565 0.543561697006226
0.0417994856834412 0.52885365486145
0.0420104265213013 0.528393983840942
0.0446122884750366 0.522725701332092
0.0494936108589172 0.512090802192688
0.0495138764381409 0.512046933174133
0.0672390460968018 0.473430514335632
0.0707660913467407 0.465746283531189
0.0725078582763672 0.46198582649231
0.0744094252586365 0.458091735839844
0.0861891508102417 0.43833315372467
0.0869072675704956 0.43732738494873
0.0916629433631897 0.431123971939087
0.0933713912963867 0.429073095321655
0.0958954095840454 0.426196932792664
0.104392945766449 0.417683601379395
0.106009840965271 0.416240930557251
0.113933503627777 0.409849762916565
0.116077423095703 0.408292531967163
0.122549533843994 0.403970241546631
0.126630306243896 0.4015052318573
0.131201565265656 0.398952722549438
0.134268462657928 0.39735209941864
0.134857833385468 0.397054076194763
0.155025660991669 0.388407588005066
0.157626271247864 0.387476444244385
0.161516785621643 0.386147737503052
0.167969763278961 0.384085297584534
0.170785903930664 0.383200407028198
0.178690910339355 0.38071608543396
0.179924249649048 0.38032853603363
0.187150895595551 0.378057479858398
0.188112199306488 0.377755403518677
0.199634730815887 0.374134302139282
0.200894117355347 0.373738527297974
0.207908391952515 0.37153422832489
0.21448540687561 0.369467377662659
0.220963537693024 0.367431640625
0.224007248878479 0.366475105285645
0.224759876728058 0.366238474845886
0.233896613121033 0.363367199897766
0.233974039554596 0.363342881202698
0.235349953174591 0.362910509109497
0.246053874492645 0.359546780586243
0.248235702514648 0.358860969543457
0.251342713832855 0.357884645462036
0.251824975013733 0.357733130455017
0.253204643726349 0.357299447059631
0.262514233589172 0.354373812675476
0.274229407310486 0.350692272186279
0.279606938362122 0.349002361297607
0.280423641204834 0.348745703697205
0.282334983348846 0.348145008087158
0.293261647224426 0.344711184501648
0.293978571891785 0.344485878944397
0.300131261348724 0.342552304267883
0.303768575191498 0.34140932559967
0.3211470246315 0.33594799041748
0.325471639633179 0.334589004516602
0.330861032009125 0.332893133163452
0.338974952697754 0.330150604248047
0.343935370445251 0.328317523002625
0.353755593299866 0.324331998825073
0.363615155220032 0.319844961166382
0.370469748973846 0.316433310508728
0.375653862953186 0.31369149684906
0.377780616283417 0.312526106834412
0.382170855998993 0.310046076774597
0.388573229312897 0.306372880935669
0.394622147083282 0.302902579307556
0.40790730714798 0.295280575752258
0.416023671627045 0.290624141693115
0.417597591876984 0.289721131324768
0.424522936344147 0.28574800491333
0.42478358745575 0.285598397254944
0.426332354545593 0.284709811210632
0.429543614387512 0.282867431640625
0.437937796115875 0.278051614761353
0.441768884658813 0.275853633880615
0.445889055728912 0.273489832878113
0.448626041412354 0.27191948890686
0.453730165958405 0.268991231918335
0.45561146736145 0.267911911010742
0.467173635959625 0.261278510093689
0.471240162849426 0.258945345878601
0.473981916904449 0.257372379302979
0.477888643741608 0.255131125450134
0.482316017150879 0.252591013908386
0.482877910137177 0.252268671989441
0.483318150043488 0.252016067504883
0.483602643013 0.251852869987488
0.493298947811127 0.246289968490601
0.496066510677338 0.244704723358154
0.508895456790924 0.23753023147583
0.512529194355011 0.235548377037048
0.514082670211792 0.234707474708557
0.517603933811188 0.232815146446228
0.523982226848602 0.229434728622437
0.527999937534332 0.227335095405579
0.532123267650604 0.225203275680542
0.539165675640106 0.221614122390747
0.541703701019287 0.220335960388184
0.552089393138885 0.215186476707458
0.572090744972229 0.205605387687683
0.572779774665833 0.205282688140869
0.574827432632446 0.204326272010803
0.578090310096741 0.202810645103455
0.578102648258209 0.202804923057556
0.583341836929321 0.200392246246338
0.584954082965851 0.199654817581177
0.586356401443481 0.199015498161316
0.588969588279724 0.197828412055969
0.592191576957703 0.196373343467712
0.594320952892303 0.195416450500488
0.596706688404083 0.194349050521851
0.60359114408493 0.191295385360718
0.615020751953125 0.186309099197388
0.616230964660645 0.185786962509155
0.629793763160706 0.180009007453918
0.631997585296631 0.17908251285553
0.632766962051392 0.178759932518005
0.633415281772614 0.178488373756409
0.636320650577545 0.177274703979492
0.64380294084549 0.174175143241882
0.646667957305908 0.172998070716858
0.649064838886261 0.172017216682434
0.650434970855713 0.171458125114441
0.659766912460327 0.167680621147156
0.662503600120544 0.166571259498596
0.670535087585449 0.163256525993347
0.681240022182465 0.158697366714478
0.681955397129059 0.158386826515198
0.690125644207001 0.154778957366943
0.691943168640137 0.153957486152649
0.700154781341553 0.150150299072266
0.701727032661438 0.149403095245361
0.714230895042419 0.143241167068481
0.715877711772919 0.142399549484253
0.720097780227661 0.140209436416626
0.721782982349396 0.139321088790894
0.722362816333771 0.139013648033142
0.724339962005615 0.137958288192749
0.725047469139099 0.137577891349792
0.728903353214264 0.13547956943512
0.729076385498047 0.135384440422058
0.732455909252167 0.13350784778595
0.736843228340149 0.131020545959473
0.739347815513611 0.129573941230774
0.74403327703476 0.126814365386963
0.749413132667542 0.123557448387146
0.750742435455322 0.122737765312195
0.756429255008698 0.119162201881409
0.756718814373016 0.118977069854736
0.759423971176147 0.117233276367188
0.764318466186523 0.114009976387024
0.771147429943085 0.109361052513123
0.774811029434204 0.106791257858276
0.784129559993744 0.100004553794861
0.790473163127899 0.0951684713363647
0.790486514568329 0.0951581001281738
0.792000591754913 0.0939767360687256
0.799741685390472 0.0877673625946045
0.800864815711975 0.0868422985076904
0.804708778858185 0.0836282968521118
0.805868983268738 0.0826433897018433
0.830307900905609 0.0600346922874451
0.83801543712616 0.0516775250434875
0.847992241382599 0.0395054221153259
0.852572441101074 0.0333139896392822
0.866361677646637 0.0117853283882141
0.871176600456238 0.00301617383956909
0.874660193920135 -0.00371038913726807
0.884599268436432 -0.0241129994392395
0.88707435131073 -0.0295074582099915
0.889352321624756 -0.0345958471298218
0.890990674495697 -0.0383327603340149
0.891286075115204 -0.0390136241912842
0.90108335018158 -0.0629679560661316
0.906848967075348 -0.0785158276557922
0.906959772109985 -0.0788266658782959
0.909624993801117 -0.0864636898040771
0.914373815059662 -0.10087126493454
0.916529655456543 -0.107793211936951
0.917276442050934 -0.11025208234787
0.933907568454742 -0.17445957660675
0.936382114887238 -0.184770584106445
0.94158673286438 -0.206456899642944
0.947788000106812 -0.232296466827393
0.954951107501984 -0.262143492698669
0.988680303096771 -0.338301002979279
0.998581647872925 -0.344289362430573
};
\end{axis}

\end{tikzpicture}
    }\\ 
    \scalebox{0.26}{
\begin{tikzpicture}

\definecolor{darkgray176}{RGB}{210,210,210}
\definecolor{dodgerblue}{RGB}{30,144,255}

\begin{axis}[
tick align=outside,
tick pos=left,
x grid style={darkgray176},
xmin=-0.0467265516519547, xmax=1.04324283897877,
xtick style={color=black},
y grid style={darkgray176},
ymin=-0.59839686602354, ymax=0.353904186189175,
ytick style={color=black},
ylabel={Node 1},
y label style={at={(axis description cs:-0.3,.5)},anchor=south},
grid=both,
grid style={line width=.1pt, draw=gray!10},
major grid style={line width=.2pt,draw=gray!20},
minor x tick num=1,
minor y tick num=1,
]
\addplot [semithick, dodgerblue]
table {%
0.00281751155853271 0.31061777472496
0.0133479833602905 0.301417350769043
0.0195685029029846 0.295982509851456
0.0220627784729004 0.2938032746315
0.0422413945198059 0.276173323392868
0.0460957884788513 0.272805750370026
0.0549092292785645 0.265105485916138
0.0579379200935364 0.262459337711334
0.059905469417572 0.26074030995369
0.0615454912185669 0.259307414293289
0.0631937384605408 0.257867336273193
0.0633630752563477 0.257719397544861
0.0704537034034729 0.251524358987808
0.0778093934059143 0.245097726583481
0.0779090523719788 0.245010659098625
0.0954298973083496 0.229702770709991
0.0959219336509705 0.22927288711071
0.0961606502532959 0.22906431555748
0.108971536159515 0.217871502041817
0.11107337474823 0.216035142540932
0.115571260452271 0.212105363607407
0.116032361984253 0.21170249581337
0.119871616363525 0.208348155021667
0.12512594461441 0.203757479786873
0.125698208808899 0.203257501125336
0.12690669298172 0.202201649546623
0.128084480762482 0.20117262005806
0.128110289573669 0.201150074601173
0.137459397315979 0.192981794476509
0.146419286727905 0.185153588652611
0.154882907867432 0.17775896191597
0.156893789768219 0.176002070307732
0.162056267261505 0.171491637825966
0.166283965110779 0.167797908186913
0.177438855171204 0.158051937818527
0.179214715957642 0.15650038421154
0.180649042129517 0.155247211456299
0.190518379211426 0.146624431014061
0.195219457149506 0.142517119646072
0.208765506744385 0.130681991577148
0.213927268981934 0.126172184944153
0.219827175140381 0.121017470955849
0.2249675989151 0.116526305675507
0.226844787597656 0.114886216819286
0.227535843849182 0.114282444119453
0.22871208190918 0.113254770636559
0.231405019760132 0.110901959240437
0.233007431030273 0.10950194299221
0.234380543231964 0.108302265405655
0.246609747409821 0.0976176634430885
0.25255024433136 0.092427484691143
0.257454514503479 0.0881426408886909
0.269976437091827 0.0772022977471352
0.275099039077759 0.0727267116308212
0.276986598968506 0.0710775554180145
0.280861973762512 0.0676916614174843
0.281703054904938 0.0669568106532097
0.28642213344574 0.0628337785601616
0.288284957408905 0.0612062364816666
0.299872934818268 0.0510818734765053
0.30469411611557 0.0468696318566799
0.311670541763306 0.0407743602991104
0.31285148859024 0.039742574095726
0.313522338867188 0.0391564555466175
0.326532125473022 0.027789868414402
0.326902806758881 0.0274660047143698
0.329703271389008 0.0250192526727915
0.337348163127899 0.0183399468660355
0.344143927097321 0.0124025214463472
0.348860740661621 0.00828146375715733
0.350317060947418 0.00700908387079835
0.356289505958557 0.00179098744411021
0.358660161495209 -0.000280242878943682
0.358758687973022 -0.00036632499541156
0.367152631282806 -0.00770007306709886
0.367277324199677 -0.00780901638790965
0.379245758056641 -0.0182657800614834
0.386250913143158 -0.0243861507624388
0.386266887187958 -0.0244001056998968
0.394435822963715 -0.031537264585495
0.39991295337677 -0.036322608590126
0.409272015094757 -0.0444995760917664
0.414407432079315 -0.0489863641560078
0.415173947811127 -0.0496560670435429
0.415933728218079 -0.0503198839724064
0.42482852935791 -0.058091226965189
0.428329586982727 -0.0611500851809978
0.429179012775421 -0.0618922226130962
0.434281051158905 -0.0663498491048813
0.437131762504578 -0.068840503692627
0.469995975494385 -0.0975538119673729
0.472166657447815 -0.0994503200054169
0.472257614135742 -0.0995297878980637
0.480677783489227 -0.106886453926563
0.483239233493805 -0.109124377369881
0.483724355697632 -0.109548225998878
0.489208221435547 -0.114339455962181
0.493338048458099 -0.117947667837143
0.495270967483521 -0.119636446237564
0.496253669261932 -0.120495028793812
0.50495058298111 -0.128093481063843
0.520527064800262 -0.141702577471733
0.522519409656525 -0.143443286418915
0.532208681106567 -0.151908755302429
0.533660054206848 -0.153176814317703
0.537782311439514 -0.156778410077095
0.540998578071594 -0.159588441252708
0.546118080615997 -0.164061322808266
0.548947513103485 -0.166533380746841
0.552013754844666 -0.169212341308594
0.553114831447601 -0.170174360275269
0.555565059185028 -0.17231510579586
0.559368371963501 -0.175638034939766
0.560492038726807 -0.176619783043861
0.560539960861206 -0.176661655306816
0.565063297748566 -0.180613666772842
0.571642577648163 -0.186361953616142
0.572652280330658 -0.18724413216114
0.580806016921997 -0.194368004798889
0.58209490776062 -0.195494100451469
0.591409504413605 -0.203632220625877
0.600167989730835 -0.211284473538399
0.60407030582428 -0.214693903923035
0.60964959859848 -0.219568505883217
0.61403900384903 -0.223403513431549
0.614174604415894 -0.22352197766304
0.615768313407898 -0.224914401769638
0.622929513454437 -0.231171101331711
0.632022619247437 -0.239115715026855
0.63678252696991 -0.243274420499802
0.639056444168091 -0.245261132717133
0.654363334178925 -0.258634686470032
0.663265824317932 -0.266412734985352
0.666122615337372 -0.268908709287643
0.668011546134949 -0.270559072494507
0.669496238231659 -0.271856218576431
0.688584446907043 -0.288533508777618
0.689998388290405 -0.289768844842911
0.697598814964294 -0.29640930891037
0.69837874174118 -0.297090739011765
0.702065110206604 -0.300311505794525
0.70631605386734 -0.304025530815125
0.706624805927277 -0.304295271635056
0.724156439304352 -0.319612592458725
0.728997051715851 -0.32384181022644
0.738590002059937 -0.332223117351532
0.749205708503723 -0.341498017311096
0.749558568000793 -0.341806292533875
0.764341175556183 -0.354721784591675
0.76554661989212 -0.355774998664856
0.76944637298584 -0.359182178974152
0.769515991210938 -0.359243005514145
0.772613525390625 -0.361949294805527
0.773476898670197 -0.362703621387482
0.779029428958893 -0.367554843425751
0.779123663902283 -0.367637187242508
0.779822468757629 -0.368247717618942
0.781309485435486 -0.369546920061111
0.783272564411163 -0.371262073516846
0.783661901950836 -0.371602207422256
0.7949178814888 -0.38143652677536
0.795761823654175 -0.382173866033554
0.803122997283936 -0.388605296611786
0.811211287975311 -0.395671993494034
0.815272748470306 -0.39922046661377
0.820028066635132 -0.403375178575516
0.826399981975555 -0.40894228219986
0.833688616752625 -0.415310323238373
0.841201722621918 -0.421874493360519
0.848510444164276 -0.428260087966919
0.853233754634857 -0.432386815547943
0.85508781671524 -0.434006690979004
0.857133150100708 -0.435793697834015
0.85740339756012 -0.436029821634293
0.857532501220703 -0.43614262342453
0.859283447265625 -0.437672406435013
0.877039909362793 -0.453186124563217
0.885247409343719 -0.46035698056221
0.89163339138031 -0.4659363925457
0.892618536949158 -0.466797113418579
0.895047187805176 -0.468919008970261
0.895302593708038 -0.4691421687603
0.906564593315125 -0.478981703519821
0.90956848859787 -0.481606215238571
0.914022088050842 -0.485497295856476
0.91419130563736 -0.485645145177841
0.915676832199097 -0.486943036317825
0.915914237499237 -0.487150460481644
0.942246556282043 -0.510156869888306
0.95756596326828 -0.523541390895844
0.962188065052032 -0.527579665184021
0.965132892131805 -0.530152559280396
0.965466856956482 -0.530444324016571
0.968339085578918 -0.532953798770905
0.984445214271545 -0.547025620937347
0.985623776912689 -0.548055350780487
0.98764044046402 -0.549817264080048
0.988257527351379 -0.550356447696686
0.988281011581421 -0.550376951694489
0.99369877576828 -0.555110454559326
};
\end{axis}

\end{tikzpicture}
    } & 
    \scalebox{0.26}{
\begin{tikzpicture}

\definecolor{darkgray176}{RGB}{210,210,210}
\definecolor{dodgerblue}{RGB}{30,144,255}

\begin{axis}[
tick align=outside,
tick pos=left,
x grid style={darkgray176},
xmin=-0.0467265516519547, xmax=1.04324283897877,
xtick style={color=black},
y grid style={darkgray176},
ymin=-0.015530888736248, ymax=0.326148663461208,
ytick style={color=black},
grid=both,
grid style={line width=.1pt, draw=gray!10},
major grid style={line width=.2pt,draw=gray!20},
minor x tick num=1,
minor y tick num=2,
ytick={0,0.1,0.2,0.3,0.4},
]
\addplot [semithick, dodgerblue]
table {%
0.00281751155853271 0.31061777472496
0.0133479833602905 0.301417350769043
0.0195685029029846 0.295982509851456
0.0220627784729004 0.2938032746315
0.0422413945198059 0.276173323392868
0.0460957884788513 0.272805750370026
0.0549092292785645 0.265105485916138
0.0579379200935364 0.262459337711334
0.059905469417572 0.26074030995369
0.0615454912185669 0.259307414293289
0.0631937384605408 0.257867336273193
0.0633630752563477 0.257719397544861
0.0704537034034729 0.251524358987808
0.0778093934059143 0.245097726583481
0.0779090523719788 0.245010659098625
0.0954298973083496 0.229702770709991
0.0959219336509705 0.22927288711071
0.0961606502532959 0.22906431555748
0.108971536159515 0.217871502041817
0.11107337474823 0.216035142540932
0.115571260452271 0.212105363607407
0.116032361984253 0.21170249581337
0.119871616363525 0.208348155021667
0.12512594461441 0.203757479786873
0.125698208808899 0.203257501125336
0.12690669298172 0.202201649546623
0.128084480762482 0.20117262005806
0.128110289573669 0.201150074601173
0.137459397315979 0.192981794476509
0.146419286727905 0.185153588652611
0.154882907867432 0.17775896191597
0.156893789768219 0.176002070307732
0.162056267261505 0.171491637825966
0.166283965110779 0.167797908186913
0.177438855171204 0.158051937818527
0.179214715957642 0.15650038421154
0.180649042129517 0.155247211456299
0.190518379211426 0.146624431014061
0.195219457149506 0.142517119646072
0.208765506744385 0.130681991577148
0.213927268981934 0.126172184944153
0.219827175140381 0.121017470955849
0.2249675989151 0.116526305675507
0.226844787597656 0.114886216819286
0.227535843849182 0.114282444119453
0.22871208190918 0.113254770636559
0.231405019760132 0.110901959240437
0.233007431030273 0.10950194299221
0.234380543231964 0.108302265405655
0.246609747409821 0.0976176634430885
0.25255024433136 0.092427484691143
0.257454514503479 0.0881426408886909
0.269976437091827 0.0772022977471352
0.275099039077759 0.0727267116308212
0.276986598968506 0.0710775554180145
0.280861973762512 0.0676916614174843
0.281703054904938 0.0669568106532097
0.28642213344574 0.0628337785601616
0.288284957408905 0.0612062364816666
0.299872934818268 0.0510818734765053
0.30469411611557 0.0468696318566799
0.311670541763306 0.0407743602991104
0.31285148859024 0.039742574095726
0.313522338867188 0.0391564555466175
0.326532125473022 0.027789868414402
0.326902806758881 0.0274660047143698
0.329703271389008 0.0250192526727915
0.337348163127899 0.0183399468660355
0.344143927097321 0.0124025214463472
0.348860740661621 0.00828146375715733
0.350317060947418 0.00700908387079835
0.356289505958557 0.00179098744411021
0.358660161495209 0
0.358758687973022 0
0.367152631282806 0
0.367277324199677 0
0.379245758056641 0
0.386250913143158 0
0.386266887187958 0
0.394435822963715 0
0.39991295337677 0
0.409272015094757 0
0.414407432079315 0
0.415173947811127 0
0.415933728218079 0
0.42482852935791 0
0.428329586982727 0
0.429179012775421 0
0.434281051158905 0
0.437131762504578 0
0.469995975494385 0
0.472166657447815 0
0.472257614135742 0
0.480677783489227 0
0.483239233493805 0
0.483724355697632 0
0.489208221435547 0
0.493338048458099 0
0.495270967483521 0
0.496253669261932 0
0.50495058298111 0
0.520527064800262 0
0.522519409656525 0
0.532208681106567 0
0.533660054206848 0
0.537782311439514 0
0.540998578071594 0
0.546118080615997 0
0.548947513103485 0
0.552013754844666 0
0.553114831447601 0
0.555565059185028 0
0.559368371963501 0
0.560492038726807 0
0.560539960861206 0
0.565063297748566 0
0.571642577648163 0
0.572652280330658 0
0.580806016921997 0
0.58209490776062 0
0.591409504413605 0
0.600167989730835 0
0.60407030582428 0
0.60964959859848 0
0.61403900384903 0
0.614174604415894 0
0.615768313407898 0
0.622929513454437 0
0.632022619247437 0
0.63678252696991 0
0.639056444168091 0
0.654363334178925 0
0.663265824317932 0
0.666122615337372 0
0.668011546134949 0
0.669496238231659 0
0.688584446907043 0
0.689998388290405 0
0.697598814964294 0
0.69837874174118 0
0.702065110206604 0
0.70631605386734 0
0.706624805927277 0
0.724156439304352 0
0.728997051715851 0
0.738590002059937 0
0.749205708503723 0
0.749558568000793 0
0.764341175556183 0
0.76554661989212 0
0.76944637298584 0
0.769515991210938 0
0.772613525390625 0
0.773476898670197 0
0.779029428958893 0
0.779123663902283 0
0.779822468757629 0
0.781309485435486 0
0.783272564411163 0
0.783661901950836 0
0.7949178814888 0
0.795761823654175 0
0.803122997283936 0
0.811211287975311 0
0.815272748470306 0
0.820028066635132 0
0.826399981975555 0
0.833688616752625 0
0.841201722621918 0
0.848510444164276 0
0.853233754634857 0
0.85508781671524 0
0.857133150100708 0
0.85740339756012 0
0.857532501220703 0
0.859283447265625 0
0.877039909362793 0
0.885247409343719 0
0.89163339138031 0
0.892618536949158 0
0.895047187805176 0
0.895302593708038 0
0.906564593315125 0
0.90956848859787 0
0.914022088050842 0
0.91419130563736 0
0.915676832199097 0
0.915914237499237 0
0.942246556282043 0
0.95756596326828 0
0.962188065052032 0
0.965132892131805 0
0.965466856956482 0
0.968339085578918 0
0.984445214271545 0
0.985623776912689 0
0.98764044046402 0
0.988257527351379 0
0.988281011581421 0
0.99369877576828 0
};
\end{axis}

\end{tikzpicture}
    } & 
    &
    \scalebox{0.26}{
\begin{tikzpicture}

\definecolor{darkgray176}{RGB}{210,210,210}
\definecolor{limegreen}{RGB}{50,205,50}

\begin{axis}[
tick align=outside,
tick pos=left,
x grid style={darkgray176},
xmin=-0.0489364236593247, xmax=1.04846346080303,
xtick style={color=black},
y grid style={darkgray176},
ymin=-0.802024242281914, ymax=0.130754765868187,
ytick style={color=black},
ylabel={Node 1},
y label style={at={(axis description cs:-0.3,.5)},anchor=south},
grid=both,
grid style={line width=.1pt, draw=gray!10},
major grid style={line width=.2pt,draw=gray!20},
minor x tick num=1,
minor y tick num=1,
]
\addplot [semithick, limegreen]
table {%
0.000945389270782471 0.0883557200431824
0.00184035301208496 0.0875950157642365
0.00233417749404907 0.087175264954567
0.0026053786277771 0.0869447514414787
0.0101443529129028 0.0805366933345795
0.0104448199272156 0.0802813023328781
0.0107607841491699 0.0800127387046814
0.026776909828186 0.0663991868495941
0.0273440480232239 0.0659171268343925
0.0350483655929565 0.0593685321509838
0.0417994856834412 0.0536301471292973
0.0420104265213013 0.0534508489072323
0.0446122884750366 0.0512392930686474
0.0494936108589172 0.0470902174711227
0.0495138764381409 0.0470729917287827
0.0672390460968018 0.032006774097681
0.0707660913467407 0.0290088206529617
0.0725078582763672 0.0275283362716436
0.0744094252586365 0.0259120222181082
0.0861891508102417 0.0158993732184172
0.0869072675704956 0.0152889806777239
0.0916629433631897 0.0112467035651207
0.0933713912963867 0.00979453977197409
0.0958954095840454 0.00764914928004146
0.104392945766449 0.000426327693276107
0.106009840965271 -0.000948017172049731
0.113933503627777 -0.0076830517500639
0.116077423095703 -0.00950536224991083
0.122549533843994 -0.015006591565907
0.126630306243896 -0.0184752084314823
0.131201565265656 -0.0223607327789068
0.134268462657928 -0.0249675642699003
0.134857833385468 -0.0254685245454311
0.155025660991669 -0.042610976845026
0.157626271247864 -0.0448214709758759
0.161516785621643 -0.0481283701956272
0.167969763278961 -0.0536133348941803
0.170785903930664 -0.0560070276260376
0.178690910339355 -0.0627262070775032
0.179924249649048 -0.0637745335698128
0.187150895595551 -0.0699171051383018
0.188112199306488 -0.0707342028617859
0.199634730815887 -0.0805282443761826
0.200894117355347 -0.0815987065434456
0.207908391952515 -0.087560772895813
0.21448540687561 -0.0931511744856834
0.220963537693024 -0.0986575186252594
0.224007248878479 -0.101244643330574
0.224759876728058 -0.101884365081787
0.233896613121033 -0.109650500118732
0.233974039554596 -0.109716318547726
0.235349953174591 -0.110885828733444
0.246053874492645 -0.119984053075314
0.248235702514648 -0.121838584542274
0.251342713832855 -0.12447951734066
0.251824975013733 -0.124889433383942
0.253204643726349 -0.126062139868736
0.262514233589172 -0.133975192904472
0.274229407310486 -0.143932983279228
0.279606938362122 -0.148503825068474
0.280423641204834 -0.14919801056385
0.282334983348846 -0.150822639465332
0.293261647224426 -0.16011019051075
0.293978571891785 -0.160719573497772
0.300131261348724 -0.165949299931526
0.303768575191498 -0.169040977954865
0.3211470246315 -0.183812484145164
0.325471639633179 -0.187488362193108
0.330861032009125 -0.192069292068481
0.338974952697754 -0.198966056108475
0.343935370445251 -0.203182354569435
0.353755593299866 -0.211529448628426
0.363615155220032 -0.219909980893135
0.370469748973846 -0.225736320018768
0.375653862953186 -0.230142757296562
0.377780616283417 -0.231950476765633
0.382170855998993 -0.235682144761086
0.388573229312897 -0.241124093532562
0.394622147083282 -0.246265605092049
0.40790730714798 -0.25755786895752
0.416023671627045 -0.264456689357758
0.417597591876984 -0.265794515609741
0.424522936344147 -0.271680980920792
0.42478358745575 -0.271902531385422
0.426332354545593 -0.273218959569931
0.429543614387512 -0.275948524475098
0.437937796115875 -0.283083468675613
0.441768884658813 -0.28633987903595
0.445889055728912 -0.28984197974205
0.448626041412354 -0.292168378829956
0.453730165958405 -0.296506851911545
0.45561146736145 -0.298105925321579
0.467173635959625 -0.307933658361435
0.471240162849426 -0.311390161514282
0.473981916904449 -0.313720613718033
0.477888643741608 -0.317041307687759
0.482316017150879 -0.320804536342621
0.482877910137177 -0.321282148361206
0.483318150043488 -0.321656346321106
0.483602643013 -0.32189816236496
0.493298947811127 -0.33013990521431
0.496066510677338 -0.33249232172966
0.508895456790924 -0.343396782875061
0.512529194355011 -0.346485435962677
0.514082670211792 -0.347805857658386
0.517603933811188 -0.350798904895782
0.523982226848602 -0.35622039437294
0.527999937534332 -0.359635412693024
0.532123267650604 -0.363140195608139
0.539165675640106 -0.369126170873642
0.541703701019287 -0.37128347158432
0.552089393138885 -0.380111217498779
0.572090744972229 -0.397112160921097
0.572779774665833 -0.39769783616066
0.574827432632446 -0.399438321590424
0.578090310096741 -0.402211725711823
0.578102648258209 -0.402222216129303
0.583341836929321 -0.406675487756729
0.584954082965851 -0.40804585814476
0.586356401443481 -0.409237831830978
0.588969588279724 -0.411458998918533
0.592191576957703 -0.414197653532028
0.594320952892303 -0.416007608175278
0.596706688404083 -0.418035477399826
0.60359114408493 -0.423887193202972
0.615020751953125 -0.43360224366188
0.616230964660645 -0.434630900621414
0.629793763160706 -0.446159154176712
0.631997585296631 -0.448032379150391
0.632766962051392 -0.448686331510544
0.633415281772614 -0.449237406253815
0.636320650577545 -0.451706945896149
0.64380294084549 -0.458066821098328
0.646667957305908 -0.460502058267593
0.649064838886261 -0.462539374828339
0.650434970855713 -0.463703960180283
0.659766912460327 -0.471636027097702
0.662503600120544 -0.473962187767029
0.670535087585449 -0.480788856744766
0.681240022182465 -0.489887952804565
0.681955397129059 -0.490496009588242
0.690125644207001 -0.49744063615799
0.691943168640137 -0.498985528945923
0.700154781341553 -0.505965292453766
0.701727032661438 -0.507301688194275
0.714230895042419 -0.51792985200882
0.715877711772919 -0.51932966709137
0.720097780227661 -0.522916674613953
0.721782982349396 -0.524349093437195
0.722362816333771 -0.524841904640198
0.724339962005615 -0.52652245759964
0.725047469139099 -0.527123868465424
0.728903353214264 -0.530401289463043
0.729076385498047 -0.530548393726349
0.732455909252167 -0.533420979976654
0.736843228340149 -0.537150144577026
0.739347815513611 -0.539278984069824
0.74403327703476 -0.543261587619781
0.749413132667542 -0.547834396362305
0.750742435455322 -0.548964321613312
0.756429255008698 -0.553798079490662
0.756718814373016 -0.554044187068939
0.759423971176147 -0.556343555450439
0.764318466186523 -0.560503840446472
0.771147429943085 -0.566308379173279
0.774811029434204 -0.569422364234924
0.784129559993744 -0.577343046665192
0.790473163127899 -0.582735061645508
0.790486514568329 -0.582746386528015
0.792000591754913 -0.584033370018005
0.799741685390472 -0.590613186359406
0.800864815711975 -0.591567873954773
0.804708778858185 -0.594835162162781
0.805868983268738 -0.59582132101059
0.830307900905609 -0.616594195365906
0.83801543712616 -0.623145520687103
0.847992241382599 -0.631625711917877
0.852572441101074 -0.635518789291382
0.866361677646637 -0.647239506244659
0.871176600456238 -0.651332139968872
0.874660193920135 -0.654293179512024
0.884599268436432 -0.662741303443909
0.88707435131073 -0.664845108985901
0.889352321624756 -0.666781365871429
0.890990674495697 -0.668173909187317
0.891286075115204 -0.668425023555756
0.90108335018158 -0.676752626895905
0.906848967075348 -0.681653320789337
0.906959772109985 -0.681747496128082
0.909624993801117 -0.684012889862061
0.914373815059662 -0.688049376010895
0.916529655456543 -0.689881801605225
0.917276442050934 -0.690516591072083
0.933907568454742 -0.704652845859528
0.936382114887238 -0.706756174564362
0.94158673286438 -0.711180090904236
0.947788000106812 -0.716451108455658
0.954951107501984 -0.722539663314819
0.988680303096771 -0.751209139823914
0.998581647872925 -0.759625196456909
};
\end{axis}

\end{tikzpicture}
    } & 
    \scalebox{0.26}{
\begin{tikzpicture}

\definecolor{darkgray176}{RGB}{210,210,210}
\definecolor{limegreen}{RGB}{50,205,50}

\begin{axis}[
tick align=outside,
tick pos=left,
x grid style={darkgray176},
xmin=-0.0489364236593247, xmax=1.04846346080303,
xtick style={color=black},
y grid style={darkgray176},
ymin=-0.835970485210419, ymax=0.25749477148056,
ytick style={color=black},
grid=both,
grid style={line width=.1pt, draw=gray!10},
major grid style={line width=.2pt,draw=gray!20},
minor x tick num=1,
minor y tick num=1,
]
\addplot [semithick, limegreen]
table {%
0.000945389270782471 0.207791805267334
0.00184035301208496 0.207406759262085
0.00233417749404907 0.207194805145264
0.0026053786277771 0.207078337669373
0.0101443529129028 0.203866720199585
0.0104448199272156 0.203739643096924
0.0107607841491699 0.203606128692627
0.026776909828186 0.196937561035156
0.0273440480232239 0.196704864501953
0.0350483655929565 0.193566799163818
0.0417994856834412 0.190850496292114
0.0420104265213013 0.190766096115112
0.0446122884750366 0.189727783203125
0.0494936108589172 0.187791466712952
0.0495138764381409 0.187783598899841
0.0672390460968018 0.180877447128296
0.0707660913467407 0.1795254945755
0.0725078582763672 0.178860545158386
0.0744094252586365 0.178136467933655
0.0861891508102417 0.17369556427002
0.0869072675704956 0.17342734336853
0.0916629433631897 0.17165732383728
0.0933713912963867 0.171024441719055
0.0958954095840454 0.170091986656189
0.104392945766449 0.166970014572144
0.106009840965271 0.166371703147888
0.113933503627777 0.163397192955017
0.116077423095703 0.162580013275146
0.122549533843994 0.160080194473267
0.126630306243896 0.158478140830994
0.131201565265656 0.156659364700317
0.134268462657928 0.15542209148407
0.134857833385468 0.155182242393494
0.155025660991669 0.146574258804321
0.157626271247864 0.145404458045959
0.161516785621643 0.143627405166626
0.167969763278961 0.140606880187988
0.170785903930664 0.139259338378906
0.178690910339355 0.135377883911133
0.179924249649048 0.134758710861206
0.187150895595551 0.13105583190918
0.188112199306488 0.130553245544434
0.199634730815887 0.124340295791626
0.200894117355347 0.123639464378357
0.207908391952515 0.119653463363647
0.21448540687561 0.115785360336304
0.220963537693024 0.111845970153809
0.224007248878479 0.109948873519897
0.224759876728058 0.109475255012512
0.233896613121033 0.103573799133301
0.233974039554596 0.103522539138794
0.235349953174591 0.102608680725098
0.246053874492645 0.0952662229537964
0.248235702514648 0.0937172174453735
0.251342713832855 0.0914796590805054
0.251824975013733 0.091128945350647
0.253204643726349 0.0901204347610474
0.262514233589172 0.0831135511398315
0.274229407310486 0.073766827583313
0.279606938362122 0.0692660212516785
0.280423641204834 0.0685703158378601
0.282334983348846 0.0669294595718384
0.293261647224426 0.0569120049476624
0.293978571891785 0.056210458278656
0.300131261348724 0.0499410629272461
0.303768575191498 0.0460094213485718
0.3211470246315 0.024430513381958
0.325471639633179 0.0181958675384521
0.330861032009125 0.00983268022537231
0.338974952697754 -0.00407952070236206
0.343935370445251 -0.0131197571754456
0.353755593299866 -0.0322070717811584
0.363615155220032 -0.0532743334770203
0.370469748973846 -0.0693087577819824
0.375653862953186 -0.0823538303375244
0.377780616283417 -0.0879680514335632
0.382170855998993 -0.100104808807373
0.388573229312897 -0.119343519210815
0.394622147083282 -0.139608979225159
0.40790730714798 -0.189572334289551
0.416023671627045 -0.220326960086823
0.417597591876984 -0.226291000843048
0.424522936344147 -0.252532362937927
0.42478358745575 -0.25352019071579
0.426332354545593 -0.25938880443573
0.429543614387512 -0.271556973457336
0.437937796115875 -0.303364098072052
0.441768884658813 -0.31483519077301
0.445889055728912 -0.322406768798828
0.448626041412354 -0.326073408126831
0.453730165958405 -0.3312668800354
0.45561146736145 -0.332812786102295
0.467173635959625 -0.339964270591736
0.471240162849426 -0.342200875282288
0.473981916904449 -0.34370881319046
0.477888643741608 -0.345857560634613
0.482316017150879 -0.348292648792267
0.482877910137177 -0.348601698875427
0.483318150043488 -0.348843812942505
0.483602643013 -0.34900027513504
0.493298947811127 -0.354333281517029
0.496066510677338 -0.355855464935303
0.508895456790924 -0.362911343574524
0.512529194355011 -0.364909946918488
0.514082670211792 -0.365764319896698
0.517603933811188 -0.367701053619385
0.523982226848602 -0.37120908498764
0.527999937534332 -0.373418867588043
0.532123267650604 -0.375686705112457
0.539165675640106 -0.379560053348541
0.541703701019287 -0.380955934524536
0.552089393138885 -0.386668086051941
0.572090744972229 -0.397668838500977
0.572779774665833 -0.398047804832458
0.574827432632446 -0.39917403459549
0.578090310096741 -0.400968611240387
0.578102648258209 -0.400975406169891
0.583341836929321 -0.403856933116913
0.584954082965851 -0.404743731021881
0.586356401443481 -0.405515015125275
0.588969588279724 -0.406952202320099
0.592191576957703 -0.408724367618561
0.594320952892303 -0.409895479679108
0.596706688404083 -0.411207675933838
0.60359114408493 -0.414994120597839
0.615020751953125 -0.421280413866043
0.616230964660645 -0.42194601893425
0.629793763160706 -0.429405570030212
0.631997585296631 -0.430617690086365
0.632766962051392 -0.431040853261948
0.633415281772614 -0.431397408246994
0.636320650577545 -0.4329953789711
0.64380294084549 -0.43711069226265
0.646667957305908 -0.438758701086044
0.649064838886261 -0.440295934677124
0.650434970855713 -0.441250890493393
0.659766912460327 -0.449876308441162
0.662503600120544 -0.453465104103088
0.670535087585449 -0.467489451169968
0.681240022182465 -0.487003147602081
0.681955397129059 -0.48830708861351
0.690125644207001 -0.50320029258728
0.691943168640137 -0.506513476371765
0.700154781341553 -0.521481990814209
0.701727032661438 -0.524348020553589
0.714230895042419 -0.547140777111053
0.715877711772919 -0.550142824649811
0.720097780227661 -0.557835340499878
0.721782982349396 -0.560907304286957
0.722362816333771 -0.561964154243469
0.724339962005615 -0.565568208694458
0.725047469139099 -0.566857933998108
0.728903353214264 -0.573886632919312
0.729076385498047 -0.574202060699463
0.732455909252167 -0.580362498760223
0.736843228340149 -0.588359951972961
0.739347815513611 -0.592925429344177
0.74403327703476 -0.601466417312622
0.749413132667542 -0.611273050308228
0.750742435455322 -0.613680243492126
0.756429255008698 -0.62321412563324
0.756718814373016 -0.623668432235718
0.759423971176147 -0.627782642841339
0.764318466186523 -0.634675741195679
0.771147429943085 -0.643276810646057
0.774811029434204 -0.647474229335785
0.784129559993744 -0.657062828540802
0.790473163127899 -0.662836670875549
0.790486514568329 -0.662848174571991
0.792000591754913 -0.664148092269897
0.799741685390472 -0.670379102230072
0.800864815711975 -0.671229898929596
0.804708778858185 -0.674048602581024
0.805868983268738 -0.674872279167175
0.830307900905609 -0.689898073673248
0.83801543712616 -0.694233775138855
0.847992241382599 -0.699846088886261
0.852572441101074 -0.702422559261322
0.866361677646637 -0.710179448127747
0.871176600456238 -0.71288800239563
0.874660193920135 -0.71484762430191
0.884599268436432 -0.720438718795776
0.88707435131073 -0.721831023693085
0.889352321624756 -0.723112404346466
0.890990674495697 -0.724034070968628
0.891286075115204 -0.724200248718262
0.90108335018158 -0.729711532592773
0.906848967075348 -0.732954859733582
0.906959772109985 -0.733017146587372
0.909624993801117 -0.734516441822052
0.914373815059662 -0.737187802791595
0.916529655456543 -0.738400518894196
0.917276442050934 -0.738820672035217
0.933907568454742 -0.748176157474518
0.936382114887238 -0.749568164348602
0.94158673286438 -0.752495944499969
0.947788000106812 -0.755984365940094
0.954951107501984 -0.760013818740845
0.988680303096771 -0.779787600040436
0.998581647872925 -0.786267518997192
};
\end{axis}

\end{tikzpicture}
    }\\ 
    \scalebox{0.26}{
\begin{tikzpicture}

\definecolor{darkgray176}{RGB}{210,210,210}
\definecolor{dodgerblue}{RGB}{30,144,255}

\begin{axis}[
tick align=outside,
tick pos=left,
x grid style={darkgray176},
xmin=-0.0467265516519547, xmax=1.04324283897877,
xtick style={color=black},
y grid style={darkgray176},
ymin=-0.793397356569767, ymax=-0.280890862643719,
ytick style={color=black},
ylabel={Node 2},
y label style={at={(axis description cs:-0.3,.5)},anchor=south},
grid=both,
grid style={line width=.1pt, draw=gray!10},
major grid style={line width=.2pt,draw=gray!20},
minor x tick num=1,
minor y tick num=1,
]
\addplot [semithick, dodgerblue]
table {%
0.00281751155853271 -0.770101606845856
0.0133479833602905 -0.765150129795074
0.0195685029029846 -0.762225210666656
0.0220627784729004 -0.761052429676056
0.0422413945198059 -0.751564383506775
0.0460957884788513 -0.749752044677734
0.0549092292785645 -0.745607912540436
0.0579379200935364 -0.744183838367462
0.059905469417572 -0.743258655071259
0.0615454912185669 -0.742487549781799
0.0631937384605408 -0.741712510585785
0.0633630752563477 -0.741632878780365
0.0704537034034729 -0.738298892974854
0.0778093934059143 -0.734840214252472
0.0779090523719788 -0.734793365001678
0.0954298973083496 -0.726554989814758
0.0959219336509705 -0.726323664188385
0.0961606502532959 -0.726211369037628
0.108971536159515 -0.720187664031982
0.11107337474823 -0.719199359416962
0.115571260452271 -0.717084467411041
0.116032361984253 -0.716867685317993
0.119871616363525 -0.715062439441681
0.12512594461441 -0.712591826915741
0.125698208808899 -0.712322771549225
0.12690669298172 -0.711754500865936
0.128084480762482 -0.711200714111328
0.128110289573669 -0.711188614368439
0.137459397315979 -0.706792593002319
0.146419286727905 -0.702579617500305
0.154882907867432 -0.698600053787231
0.156893789768219 -0.697654485702515
0.162056267261505 -0.695227086544037
0.166283965110779 -0.693239212036133
0.177438855171204 -0.687994182109833
0.179214715957642 -0.68715912103653
0.180649042129517 -0.686484754085541
0.190518379211426 -0.681844115257263
0.195219457149506 -0.679633677005768
0.208765506744385 -0.673264265060425
0.213927268981934 -0.670837223529816
0.219827175140381 -0.668063044548035
0.2249675989151 -0.665646016597748
0.226844787597656 -0.664763391017914
0.227535843849182 -0.664438426494598
0.22871208190918 -0.663885354995728
0.231405019760132 -0.662619113922119
0.233007431030273 -0.661865651607513
0.234380543231964 -0.661220014095306
0.246609747409821 -0.655469834804535
0.25255024433136 -0.652676582336426
0.257454514503479 -0.650370597839355
0.269976437091827 -0.644482731819153
0.275099039077759 -0.642074108123779
0.276986598968506 -0.641186535358429
0.280861973762512 -0.639364361763
0.281703054904938 -0.638968884944916
0.28642213344574 -0.636749923229218
0.288284957408905 -0.635874032974243
0.299872934818268 -0.630425333976746
0.30469411611557 -0.628158390522003
0.311670541763306 -0.62487804889679
0.31285148859024 -0.624322772026062
0.313522338867188 -0.624007344245911
0.326532125473022 -0.617890119552612
0.326902806758881 -0.617715835571289
0.329703271389008 -0.616399049758911
0.337348163127899 -0.612804412841797
0.344143927097321 -0.609609007835388
0.348860740661621 -0.607391119003296
0.350317060947418 -0.606706380844116
0.356289505958557 -0.603898108005524
0.358660161495209 -0.602783441543579
0.358758687973022 -0.602737128734589
0.367152631282806 -0.598790228366852
0.367277324199677 -0.598731637001038
0.379245758056641 -0.593104004859924
0.386250913143158 -0.589810192584991
0.386266887187958 -0.58980268239975
0.394435822963715 -0.585961639881134
0.39991295337677 -0.583386242389679
0.409272015094757 -0.578985631465912
0.414407432079315 -0.576570928096771
0.415173947811127 -0.576210498809814
0.415933728218079 -0.575853228569031
0.42482852935791 -0.571670889854431
0.428329586982727 -0.57002466917038
0.429179012775421 -0.56962525844574
0.434281051158905 -0.56722629070282
0.437131762504578 -0.565885901451111
0.469995975494385 -0.550433039665222
0.472166657447815 -0.549412369728088
0.472257614135742 -0.549369633197784
0.480677783489227 -0.545410394668579
0.483239233493805 -0.544206023216248
0.483724355697632 -0.543977916240692
0.489208221435547 -0.541399419307709
0.493338048458099 -0.539457559585571
0.495270967483521 -0.538548648357391
0.496253669261932 -0.538086593151093
0.50495058298111 -0.533997297286987
0.520527064800262 -0.526673197746277
0.522519409656525 -0.525736391544342
0.532208681106567 -0.52118045091629
0.533660054206848 -0.520498037338257
0.537782311439514 -0.518559753894806
0.540998578071594 -0.51704740524292
0.546118080615997 -0.514640212059021
0.548947513103485 -0.513309836387634
0.552013754844666 -0.511868059635162
0.553114831447601 -0.511350333690643
0.555565059185028 -0.51019823551178
0.559368371963501 -0.508409917354584
0.560492038726807 -0.50788152217865
0.560539960861206 -0.507858991622925
0.565063297748566 -0.505732119083405
0.571642577648163 -0.502638518810272
0.572652280330658 -0.502163767814636
0.580806016921997 -0.498329877853394
0.58209490776062 -0.497723817825317
0.591409504413605 -0.493344068527222
0.600167989730835 -0.489225804805756
0.60407030582428 -0.48739093542099
0.60964959859848 -0.484767526388168
0.61403900384903 -0.482703626155853
0.614174604415894 -0.482639878988266
0.615768313407898 -0.481890499591827
0.622929513454437 -0.478523284196854
0.632022619247437 -0.474247694015503
0.63678252696991 -0.472009569406509
0.639056444168091 -0.470940351486206
0.654363334178925 -0.463743031024933
0.663265824317932 -0.459557056427002
0.666122615337372 -0.458213776350021
0.668011546134949 -0.457325607538223
0.669496238231659 -0.456627488136292
0.688584446907043 -0.447652161121368
0.689998388290405 -0.446987330913544
0.697598814964294 -0.443413585424423
0.69837874174118 -0.443046867847443
0.702065110206604 -0.441313534975052
0.70631605386734 -0.439314723014832
0.706624805927277 -0.439169555902481
0.724156439304352 -0.430926114320755
0.728997051715851 -0.428650051355362
0.738590002059937 -0.424139410257339
0.749205708503723 -0.419147878885269
0.749558568000793 -0.418981969356537
0.764341175556183 -0.412031143903732
0.76554661989212 -0.411464363336563
0.76944637298584 -0.409630686044693
0.769515991210938 -0.409597933292389
0.772613525390625 -0.40814146399498
0.773476898670197 -0.407735526561737
0.779029428958893 -0.405124694108963
0.779123663902283 -0.405080378055573
0.779822468757629 -0.404751807451248
0.781309485435486 -0.40405261516571
0.783272564411163 -0.403129577636719
0.783661901950836 -0.402946501970291
0.7949178814888 -0.39765390753746
0.795761823654175 -0.397257089614868
0.803122997283936 -0.393795847892761
0.811211287975311 -0.389992713928223
0.815272748470306 -0.388082981109619
0.820028066635132 -0.38584703207016
0.826399981975555 -0.38285094499588
0.833688616752625 -0.379423797130585
0.841201722621918 -0.375891119241714
0.848510444164276 -0.372454553842545
0.853233754634857 -0.370233625173569
0.85508781671524 -0.369361847639084
0.857133150100708 -0.368400126695633
0.85740339756012 -0.368273049592972
0.857532501220703 -0.368212342262268
0.859283447265625 -0.367389053106308
0.877039909362793 -0.359039902687073
0.885247409343719 -0.355180710554123
0.89163339138031 -0.352178007364273
0.892618536949158 -0.351714789867401
0.895047187805176 -0.350572854280472
0.895302593708038 -0.350452750921249
0.906564593315125 -0.345157325267792
0.90956848859787 -0.343744903802872
0.914022088050842 -0.341650784015656
0.91419130563736 -0.341571241617203
0.915676832199097 -0.34087273478508
0.915914237499237 -0.340761095285416
0.942246556282043 -0.328379571437836
0.95756596326828 -0.32117635011673
0.962188065052032 -0.319003015756607
0.965132892131805 -0.317618370056152
0.965466856956482 -0.317461341619492
0.968339085578918 -0.316110789775848
0.984445214271545 -0.308537662029266
0.985623776912689 -0.307983487844467
0.98764044046402 -0.307035237550735
0.988257527351379 -0.306745082139969
0.988281011581421 -0.306734055280685
0.99369877576828 -0.30418661236763
};
\end{axis}

\end{tikzpicture}
    } & 
    \scalebox{0.26}{
\begin{tikzpicture}

\definecolor{darkgray176}{RGB}{210,210,210}
\definecolor{dodgerblue}{RGB}{30,144,255}

\begin{axis}[
tick align=outside,
tick pos=left,
x grid style={darkgray176},
xmin=-0.0467265516519547, xmax=1.04324283897877,
xtick style={color=black},
y grid style={darkgray176},
ymin=-0.055, ymax=0.055,
ytick style={color=black},
grid=both,
grid style={line width=.1pt, draw=gray!10},
major grid style={line width=.2pt,draw=gray!20},
minor x tick num=1,
minor y tick num=1,
]
\addplot [semithick, dodgerblue]
table {%
0.00281751155853271 0
0.0133479833602905 0
0.0195685029029846 0
0.0220627784729004 0
0.0422413945198059 0
0.0460957884788513 0
0.0549092292785645 0
0.0579379200935364 0
0.059905469417572 0
0.0615454912185669 0
0.0631937384605408 0
0.0633630752563477 0
0.0704537034034729 0
0.0778093934059143 0
0.0779090523719788 0
0.0954298973083496 0
0.0959219336509705 0
0.0961606502532959 0
0.108971536159515 0
0.11107337474823 0
0.115571260452271 0
0.116032361984253 0
0.119871616363525 0
0.12512594461441 0
0.125698208808899 0
0.12690669298172 0
0.128084480762482 0
0.128110289573669 0
0.137459397315979 0
0.146419286727905 0
0.154882907867432 0
0.156893789768219 0
0.162056267261505 0
0.166283965110779 0
0.177438855171204 0
0.179214715957642 0
0.180649042129517 0
0.190518379211426 0
0.195219457149506 0
0.208765506744385 0
0.213927268981934 0
0.219827175140381 0
0.2249675989151 0
0.226844787597656 0
0.227535843849182 0
0.22871208190918 0
0.231405019760132 0
0.233007431030273 0
0.234380543231964 0
0.246609747409821 0
0.25255024433136 0
0.257454514503479 0
0.269976437091827 0
0.275099039077759 0
0.276986598968506 0
0.280861973762512 0
0.281703054904938 0
0.28642213344574 0
0.288284957408905 0
0.299872934818268 0
0.30469411611557 0
0.311670541763306 0
0.31285148859024 0
0.313522338867188 0
0.326532125473022 0
0.326902806758881 0
0.329703271389008 0
0.337348163127899 0
0.344143927097321 0
0.348860740661621 0
0.350317060947418 0
0.356289505958557 0
0.358660161495209 0
0.358758687973022 0
0.367152631282806 0
0.367277324199677 0
0.379245758056641 0
0.386250913143158 0
0.386266887187958 0
0.394435822963715 0
0.39991295337677 0
0.409272015094757 0
0.414407432079315 0
0.415173947811127 0
0.415933728218079 0
0.42482852935791 0
0.428329586982727 0
0.429179012775421 0
0.434281051158905 0
0.437131762504578 0
0.469995975494385 0
0.472166657447815 0
0.472257614135742 0
0.480677783489227 0
0.483239233493805 0
0.483724355697632 0
0.489208221435547 0
0.493338048458099 0
0.495270967483521 0
0.496253669261932 0
0.50495058298111 0
0.520527064800262 0
0.522519409656525 0
0.532208681106567 0
0.533660054206848 0
0.537782311439514 0
0.540998578071594 0
0.546118080615997 0
0.548947513103485 0
0.552013754844666 0
0.553114831447601 0
0.555565059185028 0
0.559368371963501 0
0.560492038726807 0
0.560539960861206 0
0.565063297748566 0
0.571642577648163 0
0.572652280330658 0
0.580806016921997 0
0.58209490776062 0
0.591409504413605 0
0.600167989730835 0
0.60407030582428 0
0.60964959859848 0
0.61403900384903 0
0.614174604415894 0
0.615768313407898 0
0.622929513454437 0
0.632022619247437 0
0.63678252696991 0
0.639056444168091 0
0.654363334178925 0
0.663265824317932 0
0.666122615337372 0
0.668011546134949 0
0.669496238231659 0
0.688584446907043 0
0.689998388290405 0
0.697598814964294 0
0.69837874174118 0
0.702065110206604 0
0.70631605386734 0
0.706624805927277 0
0.724156439304352 0
0.728997051715851 0
0.738590002059937 0
0.749205708503723 0
0.749558568000793 0
0.764341175556183 0
0.76554661989212 0
0.76944637298584 0
0.769515991210938 0
0.772613525390625 0
0.773476898670197 0
0.779029428958893 0
0.779123663902283 0
0.779822468757629 0
0.781309485435486 0
0.783272564411163 0
0.783661901950836 0
0.7949178814888 0
0.795761823654175 0
0.803122997283936 0
0.811211287975311 0
0.815272748470306 0
0.820028066635132 0
0.826399981975555 0
0.833688616752625 0
0.841201722621918 0
0.848510444164276 0
0.853233754634857 0
0.85508781671524 0
0.857133150100708 0
0.85740339756012 0
0.857532501220703 0
0.859283447265625 0
0.877039909362793 0
0.885247409343719 0
0.89163339138031 0
0.892618536949158 0
0.895047187805176 0
0.895302593708038 0
0.906564593315125 0
0.90956848859787 0
0.914022088050842 0
0.91419130563736 0
0.915676832199097 0
0.915914237499237 0
0.942246556282043 0
0.95756596326828 0
0.962188065052032 0
0.965132892131805 0
0.965466856956482 0
0.968339085578918 0
0.984445214271545 0
0.985623776912689 0
0.98764044046402 0
0.988257527351379 0
0.988281011581421 0
0.99369877576828 0
};
\end{axis}

\end{tikzpicture}
    } & 
    &
    \scalebox{0.26}{
\begin{tikzpicture}

\definecolor{darkgray176}{RGB}{210,210,210}
\definecolor{limegreen}{RGB}{50,205,50}

\begin{axis}[
tick align=outside,
tick pos=left,
x grid style={darkgray176},
xmin=-0.0489364236593247, xmax=1.04846346080303,
xtick style={color=black},
y grid style={darkgray176},
ymin=-0.950416821241379, ymax=-0.272425991296768,
ytick style={color=black},
ylabel={Node 2},
y label style={at={(axis description cs:-0.3,.5)},anchor=south},
grid=both,
grid style={line width=.1pt, draw=gray!10},
major grid style={line width=.2pt,draw=gray!20},
minor x tick num=1,
minor y tick num=2,
]
\addplot [semithick, limegreen]
table {%
0.000945389270782471 -0.30324375629425
0.00184035301208496 -0.303796678781509
0.00233417749404907 -0.304101794958115
0.0026053786277771 -0.304269343614578
0.0101443529129028 -0.308927029371262
0.0104448199272156 -0.309112668037415
0.0107607841491699 -0.309307873249054
0.026776909828186 -0.319202870130539
0.0273440480232239 -0.319553285837173
0.0350483655929565 -0.324313133955002
0.0417994856834412 -0.328484058380127
0.0420104265213013 -0.328614383935928
0.0446122884750366 -0.330221861600876
0.0494936108589172 -0.333237618207932
0.0495138764381409 -0.333250135183334
0.0672390460968018 -0.344201028347015
0.0707660913467407 -0.346380084753036
0.0725078582763672 -0.347456187009811
0.0744094252586365 -0.348630994558334
0.0861891508102417 -0.355908691883087
0.0869072675704956 -0.356352359056473
0.0916629433631897 -0.359290480613708
0.0933713912963867 -0.360345989465714
0.0958954095840454 -0.361905366182327
0.104392945766449 -0.367155283689499
0.106009840965271 -0.368154227733612
0.113933503627777 -0.373049587011337
0.116077423095703 -0.374374151229858
0.122549533843994 -0.378372699022293
0.126630306243896 -0.380893886089325
0.131201565265656 -0.383718073368073
0.134268462657928 -0.385612845420837
0.134857833385468 -0.38597697019577
0.155025660991669 -0.398436963558197
0.157626271247864 -0.400043666362762
0.161516785621643 -0.402447283267975
0.167969763278961 -0.406434029340744
0.170785903930664 -0.408173888921738
0.178690910339355 -0.413057744503021
0.179924249649048 -0.413819700479507
0.187150895595551 -0.418284446001053
0.188112199306488 -0.418878346681595
0.199634730815887 -0.425997167825699
0.200894117355347 -0.426775217056274
0.207908391952515 -0.431108742952347
0.21448540687561 -0.435172140598297
0.220963537693024 -0.43917441368103
0.224007248878479 -0.441054880619049
0.224759876728058 -0.441519856452942
0.233896613121033 -0.447164684534073
0.233974039554596 -0.447212517261505
0.235349953174591 -0.448062568902969
0.246053874492645 -0.454675644636154
0.248235702514648 -0.45602360367775
0.251342713832855 -0.457943171262741
0.251824975013733 -0.458241105079651
0.253204643726349 -0.459093481302261
0.262514233589172 -0.464845091104507
0.274229407310486 -0.472082912921906
0.279606938362122 -0.475405246019363
0.280423641204834 -0.475909799337387
0.282334983348846 -0.477090656757355
0.293261647224426 -0.483841329813004
0.293978571891785 -0.484284251928329
0.300131261348724 -0.488085478544235
0.303768575191498 -0.490332663059235
0.3211470246315 -0.501069366931915
0.325471639633179 -0.503741145133972
0.330861032009125 -0.50707083940506
0.338974952697754 -0.51208370923996
0.343935370445251 -0.515148341655731
0.353755593299866 -0.521215438842773
0.363615155220032 -0.527306854724884
0.370469748973846 -0.531541705131531
0.375653862953186 -0.534744560718536
0.377780616283417 -0.536058485507965
0.382170855998993 -0.538770854473114
0.388573229312897 -0.542726337909698
0.394622147083282 -0.546463429927826
0.40790730714798 -0.554671227931976
0.416023671627045 -0.55968564748764
0.417597591876984 -0.560658037662506
0.424522936344147 -0.564936637878418
0.42478358745575 -0.565097630023956
0.426332354545593 -0.56605452299118
0.429543614387512 -0.568038463592529
0.437937796115875 -0.573224544525146
0.441768884658813 -0.575591444969177
0.445889055728912 -0.578136920928955
0.448626041412354 -0.579827904701233
0.453730165958405 -0.582981288433075
0.45561146736145 -0.584143579006195
0.467173635959625 -0.591286897659302
0.471240162849426 -0.593799233436584
0.473981916904449 -0.595493137836456
0.477888643741608 -0.597906768321991
0.482316017150879 -0.600642085075378
0.482877910137177 -0.60098922252655
0.483318150043488 -0.60126119852066
0.483602643013 -0.601436972618103
0.493298947811127 -0.607427537441254
0.496066510677338 -0.609137356281281
0.508895456790924 -0.617063283920288
0.512529194355011 -0.619308233261108
0.514082670211792 -0.620268046855927
0.517603933811188 -0.622443497180939
0.523982226848602 -0.626384139060974
0.527999937534332 -0.628866314888
0.532123267650604 -0.631413757801056
0.539165675640106 -0.635764718055725
0.541703701019287 -0.637332737445831
0.552089393138885 -0.643749177455902
0.572090744972229 -0.656106293201447
0.572779774665833 -0.656531989574432
0.574827432632446 -0.65779709815979
0.578090310096741 -0.659812927246094
0.578102648258209 -0.659820556640625
0.583341836929321 -0.663057446479797
0.584954082965851 -0.664053499698639
0.586356401443481 -0.664919853210449
0.588969588279724 -0.666534304618835
0.592191576957703 -0.668524920940399
0.594320952892303 -0.669840514659882
0.596706688404083 -0.671314418315887
0.60359114408493 -0.675567746162415
0.615020751953125 -0.6826291680336
0.616230964660645 -0.683376848697662
0.629793763160706 -0.691756129264832
0.631997585296631 -0.693117678165436
0.632766962051392 -0.69359302520752
0.633415281772614 -0.69399356842041
0.636320650577545 -0.695788562297821
0.64380294084549 -0.700411260128021
0.646667957305908 -0.702181279659271
0.649064838886261 -0.703662097454071
0.650434970855713 -0.704508602619171
0.659766912460327 -0.710274040699005
0.662503600120544 -0.711964786052704
0.670535087585449 -0.716926753520966
0.681240022182465 -0.723540425300598
0.681955397129059 -0.723982393741608
0.690125644207001 -0.729030132293701
0.691943168640137 -0.730153024196625
0.700154781341553 -0.735226273536682
0.701727032661438 -0.736197650432587
0.714230895042419 -0.743922710418701
0.715877711772919 -0.744940161705017
0.720097780227661 -0.747547388076782
0.721782982349396 -0.748588502407074
0.722362816333771 -0.748946785926819
0.724339962005615 -0.750168263912201
0.725047469139099 -0.750605404376984
0.728903353214264 -0.752987623214722
0.729076385498047 -0.753094494342804
0.732455909252167 -0.755182445049286
0.736843228340149 -0.757892966270447
0.739347815513611 -0.759440362453461
0.74403327703476 -0.762335121631622
0.749413132667542 -0.765658855438232
0.750742435455322 -0.766480147838593
0.756429255008698 -0.769993543624878
0.756718814373016 -0.770172417163849
0.759423971176147 -0.771843731403351
0.764318466186523 -0.774867594242096
0.771147429943085 -0.779086649417877
0.774811029434204 -0.781350076198578
0.784129559993744 -0.787107229232788
0.790473163127899 -0.791026413440704
0.790486514568329 -0.791034638881683
0.792000591754913 -0.791970074176788
0.799741685390472 -0.796752631664276
0.800864815711975 -0.797446548938751
0.804708778858185 -0.799821376800537
0.805868983268738 -0.800538182258606
0.830307900905609 -0.815636932849884
0.83801543712616 -0.82039874792099
0.847992241382599 -0.826562583446503
0.852572441101074 -0.829392313957214
0.866361677646637 -0.837911486625671
0.871176600456238 -0.840886235237122
0.874660193920135 -0.843038439750671
0.884599268436432 -0.849178969860077
0.88707435131073 -0.85070812702179
0.889352321624756 -0.852115511894226
0.890990674495697 -0.853127717971802
0.891286075115204 -0.853310167789459
0.90108335018158 -0.859363079071045
0.906848967075348 -0.862925171852112
0.906959772109985 -0.862993657588959
0.909624993801117 -0.864640295505524
0.914373815059662 -0.867574155330658
0.916529655456543 -0.868906080722809
0.917276442050934 -0.869367480278015
0.933907568454742 -0.879642426967621
0.936382114887238 -0.881171226501465
0.94158673286438 -0.884386718273163
0.947788000106812 -0.888217985630035
0.954951107501984 -0.892643451690674
0.988680303096771 -0.913481891155243
0.998581647872925 -0.919599056243896
};
\end{axis}

\end{tikzpicture}
    } & 
    \scalebox{0.26}{
\begin{tikzpicture}

\definecolor{darkgray176}{RGB}{210,210,210}
\definecolor{limegreen}{RGB}{50,205,50}

\begin{axis}[
tick align=outside,
tick pos=left,
x grid style={darkgray176},
xmin=-0.0489364236593247, xmax=1.04846346080303,
xtick style={color=black},
y grid style={darkgray176},
ymin=-0.948734021186829, ymax=-0.307764792442322,
ytick style={color=black},
grid=both,
grid style={line width=.1pt, draw=gray!10},
major grid style={line width=.2pt,draw=gray!20},
minor x tick num=1,
minor y tick num=2,
]
\addplot [semithick, limegreen]
table {%
0.000945389270782471 -0.336899757385254
0.00184035301208496 -0.337276935577393
0.00233417749404907 -0.337480783462524
0.0026053786277771 -0.337591409683228
0.0101443529129028 -0.340607106685638
0.0104448199272156 -0.340727210044861
0.0107607841491699 -0.34085351228714
0.026776909828186 -0.347256243228912
0.0273440480232239 -0.347482979297638
0.0350483655929565 -0.350562989711761
0.0417994856834412 -0.353261828422546
0.0420104265213013 -0.353346109390259
0.0446122884750366 -0.354386329650879
0.0494936108589172 -0.356337666511536
0.0495138764381409 -0.356345772743225
0.0672390460968018 -0.363431751728058
0.0707660913467407 -0.364841818809509
0.0725078582763672 -0.365538060665131
0.0744094252586365 -0.366298258304596
0.0861891508102417 -0.37100738286972
0.0869072675704956 -0.371294438838959
0.0916629433631897 -0.373195648193359
0.0933713912963867 -0.373878598213196
0.0958954095840454 -0.374887645244598
0.104392945766449 -0.378284752368927
0.106009840965271 -0.378931045532227
0.113933503627777 -0.382098734378815
0.116077423095703 -0.382955849170685
0.122549533843994 -0.385543167591095
0.126630306243896 -0.387174487113953
0.131201565265656 -0.389001965522766
0.134268462657928 -0.390227973461151
0.134857833385468 -0.390463650226593
0.155025660991669 -0.398526072502136
0.157626271247864 -0.399565696716309
0.161516785621643 -0.401121020317078
0.167969763278961 -0.403700768947601
0.170785903930664 -0.404826581478119
0.178690910339355 -0.407986760139465
0.179924249649048 -0.408479750156403
0.187150895595551 -0.411368727684021
0.188112199306488 -0.411753058433533
0.199634730815887 -0.416359424591064
0.200894117355347 -0.416862845420837
0.207908391952515 -0.419666945934296
0.21448540687561 -0.422296226024628
0.220963537693024 -0.424885958433151
0.224007248878479 -0.426102757453918
0.224759876728058 -0.426403641700745
0.233896613121033 -0.430056214332581
0.233974039554596 -0.430087208747864
0.235349953174591 -0.430637210607529
0.246053874492645 -0.434916347265244
0.248235702514648 -0.435788571834564
0.251342713832855 -0.437030643224716
0.251824975013733 -0.43722340464592
0.253204643726349 -0.437783092260361
0.262514233589172 -0.442247450351715
0.274229407310486 -0.450515419244766
0.279606938362122 -0.456071078777313
0.280423641204834 -0.457065850496292
0.282334983348846 -0.459558516740799
0.293261647224426 -0.474035680294037
0.293978571891785 -0.4749855697155
0.300131261348724 -0.483137577772141
0.303768575191498 -0.487956792116165
0.3211470246315 -0.510982394218445
0.325471639633179 -0.516712188720703
0.330861032009125 -0.523852944374084
0.338974952697754 -0.534603357315063
0.343935370445251 -0.541175663471222
0.353755593299866 -0.554186940193176
0.363615155220032 -0.567250430583954
0.370469748973846 -0.576332330703735
0.375653862953186 -0.583201050758362
0.377780616283417 -0.586018860340118
0.382170855998993 -0.591835737228394
0.388573229312897 -0.600318551063538
0.394622147083282 -0.608332931995392
0.40790730714798 -0.624813556671143
0.416023671627045 -0.633372128009796
0.417597591876984 -0.634918689727783
0.424522936344147 -0.641340255737305
0.42478358745575 -0.641570448875427
0.426332354545593 -0.642922639846802
0.429543614387512 -0.645641684532166
0.437937796115875 -0.652257800102234
0.441768884658813 -0.655064105987549
0.445889055728912 -0.657947897911072
0.448626041412354 -0.659791827201843
0.453730165958405 -0.663087606430054
0.45561146736145 -0.664258122444153
0.467173635959625 -0.67098081111908
0.471240162849426 -0.67317008972168
0.473981916904449 -0.674599528312683
0.477888643741608 -0.676574945449829
0.482316017150879 -0.678731143474579
0.482877910137177 -0.67899876832962
0.483318150043488 -0.679207563400269
0.483602643013 -0.679342031478882
0.493298947811127 -0.683734953403473
0.496066510677338 -0.68492466211319
0.508895456790924 -0.690208554267883
0.512529194355011 -0.691694259643555
0.514082670211792 -0.692329466342926
0.517603933811188 -0.693769216537476
0.523982226848602 -0.696377158164978
0.527999937534332 -0.698019862174988
0.532123267650604 -0.699705839157104
0.539165675640106 -0.702585339546204
0.541703701019287 -0.703623056411743
0.552089393138885 -0.707869529724121
0.572090744972229 -0.716047525405884
0.572779774665833 -0.716329336166382
0.574827432632446 -0.717166543006897
0.578090310096741 -0.718500673770905
0.578102648258209 -0.718505680561066
0.583341836929321 -0.720647931098938
0.584954082965851 -0.721307098865509
0.586356401443481 -0.721880435943604
0.588969588279724 -0.722948908805847
0.592191576957703 -0.724266290664673
0.594320952892303 -0.725136995315552
0.596706688404083 -0.726112484931946
0.60359114408493 -0.728927373886108
0.615020751953125 -0.733600676059723
0.616230964660645 -0.734095454216003
0.629793763160706 -0.739641010761261
0.631997585296631 -0.740542054176331
0.632766962051392 -0.740856647491455
0.633415281772614 -0.741121768951416
0.636320650577545 -0.74230968952179
0.64380294084549 -0.745369076728821
0.646667957305908 -0.746540427207947
0.649064838886261 -0.747520446777344
0.650434970855713 -0.748080730438232
0.659766912460327 -0.751896321773529
0.662503600120544 -0.753015279769897
0.670535087585449 -0.756299138069153
0.681240022182465 -0.760676145553589
0.681955397129059 -0.760968685150146
0.690125644207001 -0.764309287071228
0.691943168640137 -0.765052437782288
0.700154781341553 -0.768437683582306
0.701727032661438 -0.769096434116364
0.714230895042419 -0.77447247505188
0.715877711772919 -0.77519965171814
0.720097780227661 -0.777084827423096
0.721782982349396 -0.777846574783325
0.722362816333771 -0.778109967708588
0.724339962005615 -0.779012560844421
0.725047469139099 -0.779337406158447
0.728903353214264 -0.781124711036682
0.729076385498047 -0.781205654144287
0.732455909252167 -0.782797932624817
0.736843228340149 -0.784901142120361
0.739347815513611 -0.7861208319664
0.74403327703476 -0.78844165802002
0.749413132667542 -0.791172921657562
0.750742435455322 -0.791857957839966
0.756429255008698 -0.794789016246796
0.756718814373016 -0.794938266277313
0.759423971176147 -0.796332538127899
0.764318466186523 -0.798855245113373
0.771147429943085 -0.802375018596649
0.774811029434204 -0.804263353347778
0.784129559993744 -0.809066295623779
0.790473163127899 -0.812335908412933
0.790486514568329 -0.812342762947083
0.792000591754913 -0.813123166561127
0.799741685390472 -0.817113101482391
0.800864815711975 -0.81769198179245
0.804708778858185 -0.819673180580139
0.805868983268738 -0.820271193981171
0.830307900905609 -0.832867503166199
0.83801543712616 -0.836840093135834
0.847992241382599 -0.841982364654541
0.852572441101074 -0.844343066215515
0.866361677646637 -0.851450324058533
0.871176600456238 -0.853932023048401
0.874660193920135 -0.855727553367615
0.884599268436432 -0.86085033416748
0.88707435131073 -0.862126052379608
0.889352321624756 -0.863300144672394
0.890990674495697 -0.864144623279572
0.891286075115204 -0.864296853542328
0.90108335018158 -0.869346559047699
0.906848967075348 -0.872318267822266
0.906959772109985 -0.87237536907196
0.909624993801117 -0.873749136924744
0.914373815059662 -0.8761967420578
0.916529655456543 -0.877307891845703
0.917276442050934 -0.877692818641663
0.933907568454742 -0.886264801025391
0.936382114887238 -0.887540221214294
0.94158673286438 -0.890222787857056
0.947788000106812 -0.893419027328491
0.954951107501984 -0.897111058235168
0.988680303096771 -0.914495766162872
0.998581647872925 -0.919599056243896
};
\end{axis}

\end{tikzpicture}
    }
    \end{tabular}
    \begin{tabular}{p{2.53cm}p{3cm}p{0.0cm}p{3.34cm}p{3cm}}
         &
        \scalebox{0.26}{
\begin{tikzpicture}

\definecolor{darkgray176}{RGB}{210,210,210}
\definecolor{dodgerblue}{RGB}{30,144,255}

\begin{axis}[
tick align=outside,
tick pos=left,
x grid style={darkgray176},
xmin=-0.0467265516519547, xmax=1.04324283897877,
xtick style={color=black},
y grid style={darkgray176},
ymin=-0.578573340177536, ymax=-0.430642849206924,
ytick style={color=black},
ylabel={\(\displaystyle y\)},
xlabel={\(\displaystyle x\)},
grid=both,
grid style={line width=.1pt, draw=gray!10},
major grid style={line width=.2pt,draw=gray!20},
minor x tick num=1,
minor y tick num=2,
]
\addplot [semithick, dodgerblue]
table {%
0.00281751155853271 -0.437366962432861
0.0133479833602905 -0.44041496515274
0.0195685029029846 -0.442215472459793
0.0220627784729004 -0.442937433719635
0.0422413945198059 -0.448778033256531
0.0460957884788513 -0.449893653392792
0.0549092292785645 -0.452444672584534
0.0579379200935364 -0.453321278095245
0.059905469417572 -0.453890800476074
0.0615454912185669 -0.454365491867065
0.0631937384605408 -0.454842567443848
0.0633630752563477 -0.454891562461853
0.0704537034034729 -0.456943929195404
0.0778093934059143 -0.459072977304459
0.0779090523719788 -0.45910182595253
0.0954298973083496 -0.464173138141632
0.0959219336509705 -0.464315563440323
0.0961606502532959 -0.464384645223618
0.108971536159515 -0.468092709779739
0.11107337474823 -0.468701064586639
0.115571260452271 -0.470002949237823
0.116032361984253 -0.470136404037476
0.119871616363525 -0.471247673034668
0.12512594461441 -0.472768515348434
0.125698208808899 -0.472934126853943
0.12690669298172 -0.47328394651413
0.128084480762482 -0.4736248254776
0.128110289573669 -0.473632305860519
0.137459397315979 -0.476338386535645
0.146419286727905 -0.478931754827499
0.154882907867432 -0.481381505727768
0.156893789768219 -0.481963545084
0.162056267261505 -0.483457803726196
0.166283965110779 -0.484681487083435
0.177438855171204 -0.487910211086273
0.179214715957642 -0.488424211740494
0.180649042129517 -0.488839387893677
0.190518379211426 -0.491696000099182
0.195219457149506 -0.49305671453476
0.208765506744385 -0.496977537870407
0.213927268981934 -0.498471587896347
0.219827175140381 -0.500179290771484
0.2249675989151 -0.501667141914368
0.226844787597656 -0.50221049785614
0.227535843849182 -0.502410531044006
0.22871208190918 -0.502750992774963
0.231405019760132 -0.503530442714691
0.233007431030273 -0.503994226455688
0.234380543231964 -0.504391670227051
0.246609747409821 -0.507931351661682
0.25255024433136 -0.509650826454163
0.257454514503479 -0.511070311069489
0.269976437091827 -0.514694690704346
0.275099039077759 -0.516177415847778
0.276986598968506 -0.51672375202179
0.280861973762512 -0.517845451831818
0.281703054904938 -0.518088936805725
0.28642213344574 -0.519454836845398
0.288284957408905 -0.519994020462036
0.299872934818268 -0.523348093032837
0.30469411611557 -0.524743556976318
0.311670541763306 -0.526762843132019
0.31285148859024 -0.527104675769806
0.313522338867188 -0.527298867702484
0.326532125473022 -0.531064450740814
0.326902806758881 -0.53117173910141
0.329703271389008 -0.53198230266571
0.337348163127899 -0.534195125102997
0.344143927097321 -0.536162078380585
0.348860740661621 -0.53752738237381
0.350317060947418 -0.537948906421661
0.356289505958557 -0.539677560329437
0.358660161495209 -0.540316045284271
0.358758687973022 -0.540329933166504
0.367152631282806 -0.541511297225952
0.367277324199677 -0.541528820991516
0.379245758056641 -0.543213248252869
0.386250913143158 -0.544199168682098
0.386266887187958 -0.5442014336586
0.394435822963715 -0.545351088047028
0.39991295337677 -0.546121954917908
0.409272015094757 -0.547439157962799
0.414407432079315 -0.548161923885345
0.415173947811127 -0.548269808292389
0.415933728218079 -0.548376739025116
0.42482852935791 -0.549628555774689
0.428329586982727 -0.550121307373047
0.429179012775421 -0.550240874290466
0.434281051158905 -0.550958931446075
0.437131762504578 -0.551360130310059
0.469995975494385 -0.555985450744629
0.472166657447815 -0.556290924549103
0.472257614135742 -0.556303739547729
0.480677783489227 -0.557488799095154
0.483239233493805 -0.557849287986755
0.483724355697632 -0.557917594909668
0.489208221435547 -0.55868935585022
0.493338048458099 -0.559270620346069
0.495270967483521 -0.559542655944824
0.496253669261932 -0.559680938720703
0.50495058298111 -0.560904979705811
0.520527064800262 -0.563097178936005
0.522519409656525 -0.563377618789673
0.532208681106567 -0.564741253852844
0.533660054206848 -0.56494551897049
0.537782311439514 -0.565525710582733
0.540998578071594 -0.565978348255157
0.546118080615997 -0.566698849201202
0.548947513103485 -0.567097067832947
0.552013754844666 -0.567528605461121
0.553114831447601 -0.567683577537537
0.555565059185028 -0.568028450012207
0.559368371963501 -0.56856369972229
0.560492038726807 -0.568721830844879
0.560539960861206 -0.568728625774384
0.565063297748566 -0.569365203380585
0.571642577648163 -0.57029116153717
0.572652280330658 -0.57043331861496
0.580806016921997 -0.571580827236176
0.58209490776062 -0.571762263774872
0.591409504413605 -0.571849226951599
0.600167989730835 -0.571849226951599
0.60407030582428 -0.571849226951599
0.60964959859848 -0.571849226951599
0.61403900384903 -0.571849226951599
0.614174604415894 -0.571849226951599
0.615768313407898 -0.571849226951599
0.622929513454437 -0.571849226951599
0.632022619247437 -0.571849226951599
0.63678252696991 -0.571849226951599
0.639056444168091 -0.571849226951599
0.654363334178925 -0.571849226951599
0.663265824317932 -0.571849226951599
0.666122615337372 -0.571849226951599
0.668011546134949 -0.571849226951599
0.669496238231659 -0.571849226951599
0.688584446907043 -0.571849226951599
0.689998388290405 -0.571849226951599
0.697598814964294 -0.571849226951599
0.69837874174118 -0.571849226951599
0.702065110206604 -0.571849226951599
0.70631605386734 -0.571849226951599
0.706624805927277 -0.571849226951599
0.724156439304352 -0.571849226951599
0.728997051715851 -0.571849226951599
0.738590002059937 -0.571849226951599
0.749205708503723 -0.571849226951599
0.749558568000793 -0.571849226951599
0.764341175556183 -0.571849226951599
0.76554661989212 -0.571849226951599
0.76944637298584 -0.571849226951599
0.769515991210938 -0.571849226951599
0.772613525390625 -0.571849226951599
0.773476898670197 -0.571849226951599
0.779029428958893 -0.571849226951599
0.779123663902283 -0.571849226951599
0.779822468757629 -0.571849226951599
0.781309485435486 -0.571849226951599
0.783272564411163 -0.571849226951599
0.783661901950836 -0.571849226951599
0.7949178814888 -0.571849226951599
0.795761823654175 -0.571849226951599
0.803122997283936 -0.571849226951599
0.811211287975311 -0.571849226951599
0.815272748470306 -0.571849226951599
0.820028066635132 -0.571849226951599
0.826399981975555 -0.571849226951599
0.833688616752625 -0.571849226951599
0.841201722621918 -0.571849226951599
0.848510444164276 -0.571849226951599
0.853233754634857 -0.571849226951599
0.85508781671524 -0.571849226951599
0.857133150100708 -0.571849226951599
0.85740339756012 -0.571849226951599
0.857532501220703 -0.571849226951599
0.859283447265625 -0.571849226951599
0.877039909362793 -0.571849226951599
0.885247409343719 -0.571849226951599
0.89163339138031 -0.571849226951599
0.892618536949158 -0.571849226951599
0.895047187805176 -0.571849226951599
0.895302593708038 -0.571849226951599
0.906564593315125 -0.571849226951599
0.90956848859787 -0.571849226951599
0.914022088050842 -0.571849226951599
0.91419130563736 -0.571849226951599
0.915676832199097 -0.571849226951599
0.915914237499237 -0.571849226951599
0.942246556282043 -0.571849226951599
0.95756596326828 -0.571849226951599
0.962188065052032 -0.571849226951599
0.965132892131805 -0.571849226951599
0.965466856956482 -0.571849226951599
0.968339085578918 -0.571849226951599
0.984445214271545 -0.571849226951599
0.985623776912689 -0.571849226951599
0.98764044046402 -0.571849226951599
0.988257527351379 -0.571849226951599
0.988281011581421 -0.571849226951599
0.99369877576828 -0.571849226951599
};
\end{axis}

\end{tikzpicture}
        } &
         &
         &
        \scalebox{0.26}{
\begin{tikzpicture}

\definecolor{darkgray176}{RGB}{210,210,210}
\definecolor{limegreen}{RGB}{50,205,50}

\begin{axis}[
tick align=outside,
tick pos=left,
x grid style={darkgray176},
xmin=-0.0489364236593247, xmax=1.04846346080303,
xtick style={color=black},
y grid style={darkgray176},
ymin=0.122340910881758, ymax=0.312990299612284,
ytick style={color=black},
ylabel={\(\displaystyle y\)},
xlabel={\(\displaystyle x\)},
grid=both,
grid style={line width=.1pt, draw=gray!10},
major grid style={line width=.2pt,draw=gray!20},
minor x tick num=1,
minor y tick num=1,
]
\addplot [semithick, limegreen]
table {%
0.000945389270782471 0.177268981933594
0.00184035301208496 0.176829695701599
0.00233417749404907 0.176586702466011
0.0026053786277771 0.176453173160553
0.0101443529129028 0.172728210687637
0.0104448199272156 0.1725794672966
0.0107607841491699 0.172423169016838
0.026776909828186 0.164467453956604
0.0273440480232239 0.164184719324112
0.0350483655929565 0.160338878631592
0.0417994856834412 0.156959682703018
0.0420104265213013 0.156853944063187
0.0446122884750366 0.155549347400665
0.0494936108589172 0.15309838950634
0.0495138764381409 0.153088271617889
0.0672390460968018 0.144154757261276
0.0707660913467407 0.14237105846405
0.0725078582763672 0.141499638557434
0.0744094252586365 0.140610679984093
0.0861891508102417 0.136390373110771
0.0869072675704956 0.1361915320158
0.0916629433631897 0.135009229183197
0.0933713912963867 0.134636536240578
0.0958954095840454 0.134130999445915
0.104392945766449 0.132772400975227
0.106009840965271 0.132567778229713
0.113933503627777 0.131778553128242
0.116077423095703 0.131619557738304
0.122549533843994 0.131261214613914
0.126630306243896 0.13111974298954
0.131201565265656 0.131029978394508
0.134268462657928 0.131007671356201
0.134857833385468 0.131006792187691
0.155025660991669 0.131547480821609
0.157626271247864 0.131688162684441
0.161516785621643 0.131925076246262
0.167969763278961 0.132379978895187
0.170785903930664 0.132590994238853
0.178690910339355 0.133210077881813
0.179924249649048 0.133310362696648
0.187150895595551 0.133918330073357
0.188112199306488 0.134001940488815
0.199634730815887 0.135055258870125
0.200894117355347 0.135176301002502
0.207908391952515 0.135872930288315
0.21448540687561 0.136561632156372
0.220963537693024 0.137275189161301
0.224007248878479 0.137622982263565
0.224759876728058 0.13771016895771
0.233896613121033 0.138810396194458
0.233974039554596 0.138820067048073
0.235349953174591 0.138992592692375
0.246053874492645 0.14039808511734
0.248235702514648 0.140698745846748
0.251342713832855 0.14113561809063
0.251824975013733 0.141204357147217
0.253204643726349 0.141403138637543
0.262514233589172 0.142865568399429
0.274229407310486 0.145104140043259
0.279606938362122 0.146357893943787
0.280423641204834 0.146566107869148
0.282334983348846 0.14707262814045
0.293261647224426 0.150163277983665
0.293978571891785 0.150378093123436
0.300131261348724 0.152289271354675
0.303768575191498 0.153480380773544
0.3211470246315 0.159930661320686
0.325471639633179 0.161770850419998
0.330861032009125 0.16422475874424
0.338974952697754 0.168221905827522
0.343935370445251 0.170764446258545
0.353755593299866 0.176015317440033
0.363615155220032 0.181660115718842
0.370469748973846 0.185874804854393
0.375653862953186 0.189263850450516
0.377780616283417 0.190713495016098
0.382170855998993 0.193832606077194
0.388573229312897 0.198782846331596
0.394622147083282 0.204027563333511
0.40790730714798 0.216921269893646
0.416023671627045 0.224716156721115
0.417597591876984 0.226216897368431
0.424522936344147 0.232783287763596
0.42478358745575 0.23302935063839
0.426332354545593 0.234489798545837
0.429543614387512 0.237509816884995
0.437937796115875 0.245356842875481
0.441768884658813 0.248089969158173
0.445889055728912 0.249721318483353
0.448626041412354 0.250427603721619
0.453730165958405 0.251284211874008
0.45561146736145 0.251495569944382
0.467173635959625 0.252110809087753
0.471240162849426 0.252234578132629
0.473981916904449 0.252313613891602
0.477888643741608 0.252420336008072
0.482316017150879 0.252533316612244
0.482877910137177 0.252547085285187
0.483318150043488 0.252557784318924
0.483602643013 0.252564668655396
0.493298947811127 0.25278052687645
0.496066510677338 0.252836734056473
0.508895456790924 0.253126859664917
0.512529194355011 0.253222972154617
0.514082670211792 0.2532659471035
0.517603933811188 0.25336742401123
0.523982226848602 0.25356537103653
0.527999937534332 0.253698885440826
0.532123267650604 0.253842771053314
0.539165675640106 0.25410395860672
0.541703701019287 0.254202634096146
0.552089393138885 0.254630506038666
0.572090744972229 0.255554527044296
0.572779774665833 0.255588561296463
0.574827432632446 0.255690425634384
0.578090310096741 0.25585526227951
0.578102648258209 0.255855888128281
0.583341836929321 0.256126791238785
0.584954082965851 0.256211668252945
0.586356401443481 0.256286084651947
0.588969588279724 0.25642603635788
0.592191576957703 0.256601184606552
0.594320952892303 0.256718337535858
0.596706688404083 0.256851017475128
0.60359114408493 0.257241696119308
0.615020751953125 0.257915169000626
0.616230964660645 0.257988184690475
0.629793763160706 0.258828699588776
0.631997585296631 0.258968949317932
0.632766962051392 0.259018182754517
0.633415281772614 0.259059727191925
0.636320650577545 0.259246945381165
0.64380294084549 0.259736835956573
0.646667957305908 0.259946942329407
0.649064838886261 0.260166972875595
0.650434970855713 0.260313928127289
0.659766912460327 0.26190048456192
0.662503600120544 0.262653231620789
0.670535087585449 0.265793830156326
0.681240022182465 0.270161032676697
0.681955397129059 0.270451098680496
0.690125644207001 0.273746103048325
0.691943168640137 0.274473518133163
0.700154781341553 0.2777339220047
0.701727032661438 0.278353780508041
0.714230895042419 0.283231228590012
0.715877711772919 0.283866494894028
0.720097780227661 0.285486549139023
0.721782982349396 0.286130279302597
0.722362816333771 0.286351323127747
0.724339962005615 0.28710350394249
0.725047469139099 0.287372022867203
0.728903353214264 0.288829565048218
0.729076385498047 0.28889474272728
0.732455909252167 0.290163546800613
0.736843228340149 0.291798979043961
0.739347815513611 0.29272648692131
0.74403327703476 0.294449478387833
0.749413132667542 0.296407908201218
0.750742435455322 0.296884030103683
0.756429255008698 0.298692673444748
0.756718814373016 0.298775404691696
0.759423971176147 0.299508541822433
0.764318466186523 0.300665110349655
0.771147429943085 0.301957339048386
0.774811029434204 0.302514761686325
0.784129559993744 0.303562492132187
0.790473163127899 0.304006606340408
0.790486514568329 0.304007321596146
0.792000591754913 0.304084002971649
0.799741685390472 0.304312884807587
0.800864815711975 0.304324418306351
0.804708778858185 0.304324358701706
0.805868983268738 0.30431255698204
0.830307900905609 0.302878201007843
0.83801543712616 0.301951169967651
0.847992241382599 0.300348252058029
0.852572441101074 0.299432814121246
0.866361677646637 0.295817226171494
0.871176600456238 0.294182270765305
0.874660193920135 0.292885690927505
0.884599268436432 0.288826018571854
0.88707435131073 0.2877217233181
0.889352321624756 0.286668598651886
0.890990674495697 0.285888195037842
0.891286075115204 0.28574538230896
0.90108335018158 0.280600160360336
0.906848967075348 0.277140527963638
0.906959772109985 0.277070432901382
0.909624993801117 0.275337308645248
0.914373815059662 0.272011190652847
0.916529655456543 0.2703877389431
0.917276442050934 0.269807249307632
0.933907568454742 0.254067182540894
0.936382114887238 0.251499801874161
0.94158673286438 0.246100053191185
0.947788000106812 0.239666134119034
0.954951107501984 0.232234418392181
0.988680303096771 0.216616094112396
0.998581647872925 0.217085272073746
};
\end{axis}

\end{tikzpicture}
        }
    \end{tabular}
    \caption{DiTAC's expressiveness reflected in a 3-node hidden-layer regression network. The first three rows match the three hidden nodes. Two left columns (blue): each node's value before and after \emph{the} ReLU function. Two right columns (green): each node's value before and after \emph{a} DiTAC function. Bottom row: learned 1D regression using ReLU (blue) versus using DiTAC (green).}
    \label{fig:cpab_act_phases}
\end{figure}

\subsection{Preliminaries: 1D CPAB Transformations}  
  Let $T^\btheta$ be a diffeomorphism parameterized by $\btheta$.
 Working with diffeomorphisms usually involves expensive computations.
 In our case, since we use diffeomorphisms directly
 within DL architectures, it is even more important (than in non-DL applications) to reduce the 
associated computational burden because during training, the quantities
$x\mapsto T^\btheta(x)$ and $x\mapsto \nabla_\btheta T^\btheta( x)$
are computed at multiple values of $x$  and for multiple values of $\btheta$. 

The main reason why we chose CPAB transformations~\cite{Freifeld:ICCV:2015:CPAB,Freifeld:PAMI:2017:CPAB} as the diffeomorphism family to be used within our method is that they are both expressive and efficient.
Throughout the remainder of this paper all the CPAB transformations are assumed to be in 1D (for the more general case, see~\cite{Freifeld:PAMI:2017:CPAB}).
In a presentation based on~\cite{Freifeld:PAMI:2017:CPAB}, we now briefly explain what {CPAB transformations} are, as well as their name. 

Let $\Omega=[a,b]\subset \RR$ be a finite interval and let $\Vcal$ be a space 
of continuous functions, from $\Omega$ to $\RR$, that are also piecewise-affine
\wrt some fixed partition of $\Omega$ into sub-intervals. 
Note that $\Vcal$ is a finite-dimensional linear space.  
Let $d=\dim(\Vcal)$, let $\btheta\in\Rd$, and let $v^\btheta\in \Vcal$ denote the generic element of $\Vcal$, parameterized by $\btheta$. 
The space of CPAB transformations obtained via the integration of elements of $\Vcal$,  is 
defined as 
\begin{align}
&\Tcal\triangleq
   \Bigl \{
 T^\btheta:
   x\mapsto \phi^\btheta( x;1)
  \text{ s.t. } \phi^\btheta( x;t) \text{ solves the integral equation} \nonumber \\ 
 & \phi^\btheta(x;t) = x+\int_{0}^t  v^\btheta(\phi^\btheta(x;\tau))\, 
 \mathrm{d}\tau \text{ where }  v^\btheta\in \Vcal\, 
  \Bigr
 \}\, .
 \label{Eqn:IntegralEquation}
\end{align}
 It can be shown that every $T^\btheta\in\Tcal$ is an order-preserving transformation (\ie, it is monotonically increasing) and a diffeomorphism~\cite{Freifeld:PAMI:2017:CPAB}.
Note that while $v^\btheta\in\Vcal$ is CPA, the CPAB $T^\btheta\in\Tcal$ is not (\eg, $T^\btheta$ is differentiable, unlike any non-trivial CPA function). 
\Autoref{Eqn:IntegralEquation}
 also implies that the elements of $\Vcal$ are viewed as velocity fields.

Particularly useful for us are the following facts: 
1) The finer the partition of $\Omega$ is, the more expressive the CPAB family becomes (which also means that $d$ increases). 
2) CPAB transformations lend themselves to fast and accurate computations in closed form 
of both $x\mapsto T^\btheta(x)$~\cite{Freifeld:PAMI:2017:CPAB}
and the gradient, $x\mapsto\nabla_\btheta T^\btheta(x)$~\cite{Freifeld:TR_CPAB_Derivaitive:2017,Martinez:ICML:2022:closed}.
Together, these facts mean that \emph{CPAB transformations provide us with a convenient and an efficient way 
to parameterize, and optimize over, nonlinear monotonically-increasing functions}.

\subsection{The DiTAC Activation Function}\label{Sec:Method:Subsec:CPABAct}
Our proposed TAF, called DiTAC,  is a TAF derived from a CPAB transformation. 
DiTAC includes a negligible amount of trainable parameters, and yet it is highly expressive.
Unlike existing TAFs which dedicate a parameter for each input's channel, DiTAC's expressiveness stems from the expressiveness of CPAB transformations.
For illustration, in~\autoref{fig:cpab_act_phases} we show how nonlinearity evolves step-by-step in a regression MLP with a 3-node hidden layer, when using either ReLU or DiTAC.
While in the ReLU case, the expressiveness is mostly reflected only after summing all of the activation responses (and the resulting function is non-differentiable in several locations), DiTAC's expressiveness (and differentiability) is distinctly shown already in the very first data transformation each neuron goes through.
Importantly, the availability of closed-form expressions for the CPAB transformation and its gradient makes it easy to use DiTAC as a drop-in replacement instead of any AF in any DL architecture. 
%


We now explain how the DiTAC is built. 
Recall that a CPAB transformation, $T^\btheta$, is defined on a finite interval, $[a,b]$. Its co-domain is also a finite interval, which may or may not coincide with $\Omega$ (depending whether one imposes zero-boundary conditions
on $\bv^\btheta$ or not; see~\cite{Freifeld:PAMI:2017:CPAB}).
As some of the AF's input can fall outside of $[a,b]$, our main version of DiTAC combines $T^\btheta$ 
with GELU, a recent widely-used AF in state-of-the-art models. 
%
Recall that GELU~\cite{Hendrycks:2016:gelu} is given by $\mathrm{GELU}(x) = x\cdot \Phi(x)$
where $\Phi$ is the cumulative distribution function of a standard normal distribution.

The GELU-like DiTAC function is  
\begin{align}
    \mathrm{DiTAC}(x) = \Tilde{x}\cdot\Phi(x)\,,
   \qquad
      \Tilde{x}   =    \begin{cases}
       T^\btheta(x) & \quad \text{If } a \leq x \leq b\\
       x & \text{Otherwise}\\
     \end{cases}\, 
\end{align}
where $T^\btheta$ is a (learnable) CPAB transformation while $\Omega=[a,b]$, the domain of $T^\btheta$, is user-defined. 
This main DiTAC version is the one used in the experiments 
in the paper (while the appendix also contains other versions;
see below).

Other versions of DiTAC can also be built by combining $T^\btheta$ with a variety of other AFs, 
not just with GELU. For instance, we present Leaky-DiTAC, where $T^\btheta$ is applied on 
$[a,b]$ while the rest of the data goes through a Leaky-ReLU (LReLU) function. That is, 
\begin{align}	  
\mathrm{Leaky \ DiTAC}(x) = 
     \begin{cases}
        T^\btheta(x) & \quad \text{If } a \leq x \leq b\\ \mathrm{LReLU}(x) & \text{Otherwise}\\
     \end{cases}
\end{align}
For an illustration of both these DiTAC types, see \autoref{fig:cpab_act_method}.
Additional DiTAC's versions and their illustrations can be found in the Appendix.\\

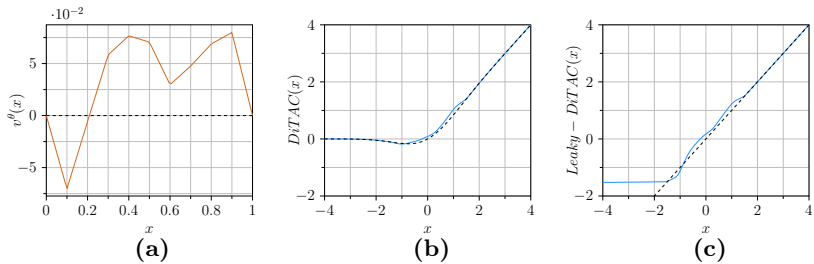
\begin{figure}[t]
    \renewcommand{\arraystretch}{0.04}
    \centering
    \Large
    \begin{tabular}{ccc}
    \scalebox{0.40}{

\begin{tikzpicture}

\definecolor{chocolate}{RGB}{210,105,30}
\definecolor{darkgray176}{RGB}{190,190,190}

\begin{axis}[
tick align=outside,
tick pos=left,
x grid style={darkgray176},
xlabel={\(\displaystyle x\)},
xmin=0, xmax=1,
xtick style={color=black},
y grid style={darkgray176},
ylabel={\(\displaystyle v^{\theta}(x)\)},
ymin=-0.0772934064269066, ymax=0.0872411534190178,
ytick style={color=black},
y label style={at={(axis description cs:-0.08,.5)},anchor=south},
x label style={at={(axis description cs:.5,-0.15)}},
grid=both,
grid style={line width=.1pt, draw=gray!10},
major grid style={line width=.2pt,draw=gray!20},
minor x tick num=1,
minor y tick num=1,
]
\addplot [semithick, black, dashed]
table {%
0 0
0.0101010101010101 0
0.0202020202020202 0
0.0303030303030303 0
0.0404040404040404 0
0.0505050505050505 0
0.0606060606060606 0
0.0707070707070707 0
0.0808080808080808 0
0.0909090909090909 0
0.101010101010101 0
0.111111111111111 0
0.121212121212121 0
0.131313131313131 0
0.141414141414141 0
0.151515151515152 0
0.161616161616162 0
0.171717171717172 0
0.181818181818182 0
0.191919191919192 0
0.202020202020202 0
0.212121212121212 0
0.222222222222222 0
0.232323232323232 0
0.242424242424242 0
0.252525252525253 0
0.262626262626263 0
0.272727272727273 0
0.282828282828283 0
0.292929292929293 0
0.303030303030303 0
0.313131313131313 0
0.323232323232323 0
0.333333333333333 0
0.343434343434343 0
0.353535353535354 0
0.363636363636364 0
0.373737373737374 0
0.383838383838384 0
0.393939393939394 0
0.404040404040404 0
0.414141414141414 0
0.424242424242424 0
0.434343434343434 0
0.444444444444444 0
0.454545454545455 0
0.464646464646465 0
0.474747474747475 0
0.484848484848485 0
0.494949494949495 0
0.505050505050505 0
0.515151515151515 0
0.525252525252525 0
0.535353535353535 0
0.545454545454546 0
0.555555555555556 0
0.565656565656566 0
0.575757575757576 0
0.585858585858586 0
0.595959595959596 0
0.606060606060606 0
0.616161616161616 0
0.626262626262626 0
0.636363636363636 0
0.646464646464647 0
0.656565656565657 0
0.666666666666667 0
0.676767676767677 0
0.686868686868687 0
0.696969696969697 0
0.707070707070707 0
0.717171717171717 0
0.727272727272727 0
0.737373737373737 0
0.747474747474748 0
0.757575757575758 0
0.767676767676768 0
0.777777777777778 0
0.787878787878788 0
0.797979797979798 0
0.808080808080808 0
0.818181818181818 0
0.828282828282828 0
0.838383838383838 0
0.848484848484849 0
0.858585858585859 0
0.868686868686869 0
0.878787878787879 0
0.888888888888889 0
0.898989898989899 0
0.909090909090909 0
0.919191919191919 0
0.929292929292929 0
0.939393939393939 0
0.94949494949495 0
0.95959595959596 0
0.96969696969697 0
0.97979797979798 0
0.98989898989899 0
1 0
};
\addplot [semithick, chocolate, opacity=1.0]
table {%
0 -1.22143507570275e-17
0.0101010101010101 -0.00711806723847985
0.0202020202020202 -0.0142361344769597
0.0303030303030303 -0.0213542021811008
0.0404040404040404 -0.0284722689539194
0.0505050505050505 -0.035590335726738
0.0606060606060606 -0.0427084043622017
0.0707070707070707 -0.0498264692723751
0.0808080808080808 -0.0569445379078388
0.0909090909090909 -0.0640626102685928
0.101010101010101 -0.0698145627975464
0.111111111111111 -0.0632712692022324
0.121212121212121 -0.0567279830574989
0.131313131313131 -0.0501847006380558
0.141414141414141 -0.0436414182186127
0.151515151515152 -0.0370981246232986
0.161616161616162 -0.0305548403412104
0.171717171717172 -0.0240115560591221
0.181818181818182 -0.0174682624638081
0.191919191919192 -0.0109249800443649
0.202020202020202 -0.00439568981528282
0.212121212121212 0.00207754597067833
0.222222222222222 0.00855077244341373
0.232323232323232 0.0150239989161491
0.242424242424242 0.0214972347021103
0.252525252525253 0.0279704518616199
0.262626262626263 0.0344436876475811
0.272727272727273 0.0409169234335423
0.282828282828283 0.0473901405930519
0.292929292929293 0.0538633763790131
0.303030303030303 0.0589487515389919
0.313131313131313 0.0607957653701305
0.323232323232323 0.0626427829265594
0.333333333333333 0.0644898042082787
0.343434343434343 0.0663368180394173
0.353535353535354 0.0681838393211365
0.363636363636364 0.0700308606028557
0.373737373737374 0.0718778744339943
0.383838383838384 0.0737248957157135
0.393939393939394 0.0755719169974327
0.404040404040404 0.0764340609312057
0.414141414141414 0.0758189037442207
0.424242424242424 0.0752037465572357
0.434343434343434 0.0745885893702507
0.444444444444444 0.0739734321832657
0.454545454545455 0.0733582675457001
0.464646464646465 0.0727431103587151
0.474747474747475 0.07212795317173
0.484848484848485 0.071512795984745
0.494949494949495 0.07089763879776
0.505050505050505 0.0685262903571129
0.515151515151515 0.0643988102674484
0.525252525252525 0.0602713339030743
0.535353535353535 0.0561438538134098
0.545454545454546 0.0520163774490356
0.555555555555556 0.0478888973593712
0.565656565656566 0.0437614433467388
0.575757575757576 0.0396339632570744
0.585858585858586 0.0355064831674099
0.595959595959596 0.0313790068030357
0.606060606060606 0.0308113675564528
0.616161616161616 0.0326169095933437
0.626262626262626 0.034422442317009
0.636363636363636 0.0362279824912548
0.646464646464647 0.0380335263907909
0.656565656565657 0.0398390702903271
0.666666666666667 0.0416446104645729
0.676767676767677 0.043450154364109
0.686868686868687 0.0452556870877743
0.696969696969697 0.0470612272620201
0.707070707070707 0.0491014085710049
0.717171717171717 0.0512421056628227
0.727272727272727 0.0533828027546406
0.737373737373737 0.0555234886705875
0.747474747474748 0.0576641820371151
0.757575757575758 0.059804879128933
0.767676767676768 0.0619455762207508
0.777777777777778 0.0640862733125687
0.787878787878788 0.0662269666790962
0.797979797979798 0.0683676525950432
0.808080808080808 0.0696910098195076
0.818181818181818 0.0708100497722626
0.828282828282828 0.071929082274437
0.838383838383838 0.0730481147766113
0.848484848484849 0.0741671472787857
0.858585858585859 0.0752861723303795
0.868686868686869 0.0764052048325539
0.878787878787879 0.0775242373347282
0.888888888888889 0.0786432772874832
0.898989898989899 0.0797623097896576
0.909090909090909 0.0726128965616226
0.919191919191919 0.0645448341965675
0.929292929292929 0.0564767234027386
0.939393939393939 0.0484086163341999
0.94949494949495 0.0403405055403709
0.95959595959596 0.032272394746542
0.96969696969697 0.024204283952713
0.97979797979798 0.0161362215876579
0.98989898989899 0.00806811079382896
1 0
};
\end{axis}

\end{tikzpicture}
    } & 
    \scalebox{0.40}{
\begin{tikzpicture}

\definecolor{darkgray176}{RGB}{190,190,190}
\definecolor{dodgerblue}{RGB}{30,144,255}

\begin{axis}[
tick align=outside,
tick pos=left,
x grid style={darkgray176},
xlabel={\(\displaystyle x\)},
xmin=-4, xmax=4,
xtick style={color=black},
y grid style={darkgray176},
ylabel={\(\displaystyle DiTAC(x)\)},
ymin=-2, ymax=4,
ytick style={color=black},
y label style={at={(axis description cs:-0.08,.5)},anchor=south},
x label style={at={(axis description cs:.5,-0.15)}},
grid=both,
grid style={line width=.1pt, draw=gray!10},
major grid style={line width=.2pt,draw=gray!20},
minor x tick num=1,
minor y tick num=1,
]
\addplot [semithick, dodgerblue]
table {%
-4 -0.00012671947479248
-3.91919183731079 -0.00017415032198187
-3.83838391304016 -0.000237708140048198
-3.75757575035095 -0.000322291336487979
-3.67676758766174 -0.000434250541729853
-3.59595966339111 -0.000581064610742033
-3.5151515007019 -0.000772497849538922
-3.4343433380127 -0.00102003407664597
-3.35353541374207 -0.00133783894125372
-3.27272725105286 -0.00174294819589704
-3.19191908836365 -0.00225554686039686
-3.11111116409302 -0.00289930240251124
-3.03030300140381 -0.00370162911713123
-2.9494948387146 -0.00469404365867376
-2.86868691444397 -0.0059120487421751
-2.78787875175476 -0.00739550217986107
-2.70707082748413 -0.00918773841112852
-2.62626266479492 -0.011336050927639
-2.54545450210571 -0.0138899739831686
-2.46464657783508 -0.0169011298567057
-2.38383841514587 -0.0204212907701731
-2.30303025245667 -0.0245009362697601
-2.22222232818604 -0.0291870050132275
-2.14141416549683 -0.0345202684402466
-2.06060600280762 -0.0405327528715134
-1.9797979593277 -0.047244131565094
-1.89898991584778 -0.0546584464609623
-1.81818187236786 -0.0627603307366371
-1.73737370967865 -0.0715113654732704
-1.65656566619873 -0.0808464661240578
-1.57575762271881 -0.0906704068183899
-1.4949494600296 -0.101049117743969
-1.41414141654968 -0.1146285161376
-1.33333337306976 -0.129284739494324
-1.25252521038055 -0.144906774163246
-1.17171716690063 -0.161238014698029
-1.09090912342072 -0.174740299582481
-1.01010096073151 -0.177348062396049
-0.929292917251587 -0.173045620322227
-0.848484873771667 -0.163398012518883
-0.767676770687103 -0.148717030882835
-0.686868667602539 -0.129508212208748
-0.60606062412262 -0.112991750240326
-0.525252521038055 -0.0953695848584175
-0.444444447755814 -0.0737346410751343
-0.363636374473572 -0.0499008297920227
-0.282828271389008 -0.0233196951448917
-0.202020198106766 0.00656703300774097
-0.121212124824524 0.0384282879531384
-0.0404040440917015 0.0707218796014786
0.0404040440917015 0.103545248508453
0.121212124824524 0.139358043670654
0.202020198106766 0.178634628653526
0.282828271389008 0.230754271149635
0.363636374473572 0.304538398981094
0.444444447755814 0.383364230394363
0.525252521038055 0.467576086521149
0.60606062412262 0.557827889919281
0.686868667602539 0.652990579605103
0.767676770687103 0.751203179359436
0.848484873771667 0.849390983581543
0.929292917251587 0.94853812456131
1.01010096073151 1.0450918674469
1.09090912342072 1.12460565567017
1.17171716690063 1.18851780891418
1.25252521038055 1.24250757694244
1.33333337306976 1.29487359523773
1.41414141654968 1.34616816043854
1.4949494600296 1.39695978164673
1.57575762271881 1.48508715629578
1.65656566619873 1.57571911811829
1.73737370967865 1.66586244106293
1.81818187236786 1.75542151927948
1.89898991584778 1.84433150291443
1.9797979593277 1.93255388736725
2.06060600280762 2.0200731754303
2.14141416549683 2.10689377784729
2.22222232818604 2.193035364151
2.30303025245667 2.27852916717529
2.38383841514587 2.36341714859009
2.46464657783508 2.44774556159973
2.54545450210571 2.53156447410583
2.62626266479492 2.61492657661438
2.70707082748413 2.69788289070129
2.78787875175476 2.78048324584961
2.86868691444397 2.86277484893799
2.9494948387146 2.94480085372925
3.03030300140381 3.02660131454468
3.11111116409302 3.10821175575256
3.19191908836365 3.18966341018677
3.27272725105286 3.27098441123962
3.35353541374207 3.35219764709473
3.4343433380127 3.4333233833313
3.5151515007019 3.51437902450562
3.59595966339111 3.59537863731384
3.67676758766174 3.67633318901062
3.75757575035095 3.75725340843201
3.83838391304016 3.8381462097168
3.91919183731079 3.91901755332947
4 3.99987316131592
};
\addplot [semithick, black, dashed]
table {%
-4 -0.00012671947479248
-3.91919183731079 -0.00017415032198187
-3.83838391304016 -0.000237708140048198
-3.75757575035095 -0.000322291336487979
-3.67676758766174 -0.000434250541729853
-3.59595966339111 -0.000581064610742033
-3.5151515007019 -0.000772497849538922
-3.4343433380127 -0.00102003407664597
-3.35353541374207 -0.00133783894125372
-3.27272725105286 -0.00174294819589704
-3.19191908836365 -0.00225554686039686
-3.11111116409302 -0.00289930240251124
-3.03030300140381 -0.00370162911713123
-2.9494948387146 -0.00469404365867376
-2.86868691444397 -0.0059120487421751
-2.78787875175476 -0.00739550217986107
-2.70707082748413 -0.00918773841112852
-2.62626266479492 -0.011336050927639
-2.54545450210571 -0.0138899739831686
-2.46464657783508 -0.0169011298567057
-2.38383841514587 -0.0204212907701731
-2.30303025245667 -0.0245009362697601
-2.22222232818604 -0.0291870050132275
-2.14141416549683 -0.0345202684402466
-2.06060600280762 -0.0405327528715134
-1.9797979593277 -0.047244131565094
-1.89898991584778 -0.0546584464609623
-1.81818187236786 -0.0627603307366371
-1.73737370967865 -0.0715113654732704
-1.65656566619873 -0.0808464661240578
-1.57575762271881 -0.0906704068183899
-1.4949494600296 -0.100854992866516
-1.41414141654968 -0.11123663187027
-1.33333337306976 -0.121614933013916
-1.25252521038055 -0.131752207875252
-1.17171716690063 -0.141373917460442
-1.09090912342072 -0.150170683860779
-1.01010096073151 -0.157801449298859
-0.929292917251587 -0.163898140192032
-0.848484873771667 -0.168071269989014
-0.767676770687103 -0.169917285442352
-0.686868667602539 -0.169026538729668
-0.60606062412262 -0.164992272853851
-0.525252521038055 -0.157420203089714
-0.444444447755814 -0.145938068628311
-0.363636374473572 -0.130205377936363
-0.282828271389008 -0.109922401607037
-0.202020198106766 -0.084838479757309
-0.121212124824524 -0.0547589734196663
-0.0404040440917015 -0.0195509307086468
0.0404040440917015 0.0208531115204096
0.121212124824524 0.0664531588554382
0.202020198106766 0.117181718349457
0.282828271389008 0.17290586233139
0.363636374473572 0.233430996537209
0.444444447755814 0.298506379127502
0.525252521038055 0.367832332849503
0.60606062412262 0.441068351268768
0.686868667602539 0.51784211397171
0.767676770687103 0.597759485244751
0.848484873771667 0.680413603782654
0.929292917251587 0.765394747257233
1.01010096073151 0.852299511432648
1.09090912342072 0.940738439559937
1.17171716690063 1.03034329414368
1.25252521038055 1.12077295780182
1.33333337306976 1.21171844005585
1.41414141654968 1.30290472507477
1.4949494600296 1.39409446716309
1.57575762271881 1.48508715629578
1.65656566619873 1.57571911811829
1.73737370967865 1.66586244106293
1.81818187236786 1.75542151927948
1.89898991584778 1.84433150291443
1.9797979593277 1.93255388736725
2.06060600280762 2.0200731754303
2.14141416549683 2.10689377784729
2.22222232818604 2.193035364151
2.30303025245667 2.27852916717529
2.38383841514587 2.36341714859009
2.46464657783508 2.44774556159973
2.54545450210571 2.53156447410583
2.62626266479492 2.61492657661438
2.70707082748413 2.69788289070129
2.78787875175476 2.78048324584961
2.86868691444397 2.86277484893799
2.9494948387146 2.94480085372925
3.03030300140381 3.02660131454468
3.11111116409302 3.10821175575256
3.19191908836365 3.18966341018677
3.27272725105286 3.27098441123962
3.35353541374207 3.35219764709473
3.4343433380127 3.4333233833313
3.5151515007019 3.51437902450562
3.59595966339111 3.59537863731384
3.67676758766174 3.67633318901062
3.75757575035095 3.75725340843201
3.83838391304016 3.8381462097168
3.91919183731079 3.91901755332947
4 3.99987316131592
};
\end{axis}

\end{tikzpicture}
    } & 
    \scalebox{0.40}{
\begin{tikzpicture}

\definecolor{darkgray176}{RGB}{190,190,190}
\definecolor{dodgerblue}{RGB}{30,144,255}

\begin{axis}[
tick align=outside,
tick pos=left,
x grid style={darkgray176},
xlabel={\(\displaystyle x\)},
xmin=-4, xmax=4,
xtick style={color=black},
y grid style={darkgray176},
ylabel={\(\displaystyle Leaky-DiTAC(x)\)},
ymin=-2, ymax=4,
ytick style={color=black},
y label style={at={(axis description cs:-0.08,.5)},anchor=south},
x label style={at={(axis description cs:.5,-0.15)}},
grid=both,
grid style={line width=.1pt, draw=gray!10},
major grid style={line width=.2pt,draw=gray!20},
minor x tick num=1,
minor y tick num=1,
]
\addplot [semithick, dodgerblue]
table {%
-4 -1.52499997615814
-3.91919183731079 -1.52419197559357
-3.83838391304016 -1.5233838558197
-3.75757575035095 -1.52257573604584
-3.67676758766174 -1.52176773548126
-3.59595966339111 -1.5209596157074
-3.5151515007019 -1.52015149593353
-3.4343433380127 -1.51934349536896
-3.35353541374207 -1.51853537559509
-3.27272725105286 -1.51772725582123
-3.19191908836365 -1.51691925525665
-3.11111116409302 -1.51611113548279
-3.03030300140381 -1.51530301570892
-2.9494948387146 -1.51449501514435
-2.86868691444397 -1.51368689537048
-2.78787875175476 -1.51287877559662
-2.70707082748413 -1.51207077503204
-2.62626266479492 -1.51126265525818
-2.54545450210571 -1.51045453548431
-2.46464657783508 -1.50964653491974
-2.38383841514587 -1.50883841514587
-2.30303025245667 -1.50803029537201
-2.22222232818604 -1.50722229480743
-2.14141416549683 -1.50641417503357
-2.06060600280762 -1.5056060552597
-1.9797979593277 -1.50479793548584
-1.89898991584778 -1.50398993492126
-1.81818187236786 -1.5031818151474
-1.73737370967865 -1.50237369537354
-1.65656566619873 -1.50156569480896
-1.57575762271881 -1.5007575750351
-1.4949494600296 -1.49782693386078
-1.41414141654968 -1.45726215839386
-1.33333337306976 -1.41742169857025
-1.25252521038055 -1.37758135795593
-1.17171716690063 -1.33635222911835
-1.09090912342072 -1.26939415931702
-1.01010096073151 -1.13522052764893
-0.929292917251587 -0.981158554553986
-0.848484873771667 -0.824892520904541
-0.767676770687103 -0.671895265579224
-0.686868667602539 -0.526279091835022
-0.60606062412262 -0.415048837661743
-0.525252521038055 -0.318212747573853
-0.444444447755814 -0.224553823471069
-0.363636374473572 -0.139362573623657
-0.282828271389008 -0.0600011348724365
-0.202020198106766 0.0156376361846924
-0.121212124824524 0.0850632190704346
-0.0404040440917015 0.146154165267944
0.0404040440917015 0.200624585151672
0.121212124824524 0.254192352294922
0.202020198106766 0.307964444160461
0.282828271389008 0.377452969551086
0.363636374473572 0.474406719207764
0.444444447755814 0.570788860321045
0.525252521038055 0.667683362960815
0.60606062412262 0.766496896743774
0.686868667602539 0.866130352020264
0.767676770687103 0.964737892150879
0.848484873771667 1.05920195579529
0.929292917251587 1.15165376663208
1.01010096073151 1.23858833312988
1.09090912342072 1.30412721633911
1.17171716690063 1.35159492492676
1.25252521038055 1.3885703086853
1.33333337306976 1.42483448982239
1.41414141654968 1.46109843254089
1.4949494600296 1.49802207946777
1.57575762271881 1.57575762271881
1.65656566619873 1.65656566619873
1.73737370967865 1.73737370967865
1.81818187236786 1.81818187236786
1.89898991584778 1.89898991584778
1.9797979593277 1.9797979593277
2.06060600280762 2.06060600280762
2.14141416549683 2.14141416549683
2.22222232818604 2.22222232818604
2.30303025245667 2.30303025245667
2.38383841514587 2.38383841514587
2.46464657783508 2.46464657783508
2.54545450210571 2.54545450210571
2.62626266479492 2.62626266479492
2.70707082748413 2.70707082748413
2.78787875175476 2.78787875175476
2.86868691444397 2.86868691444397
2.9494948387146 2.9494948387146
3.03030300140381 3.03030300140381
3.11111116409302 3.11111116409302
3.19191908836365 3.19191908836365
3.27272725105286 3.27272725105286
3.35353541374207 3.35353541374207
3.4343433380127 3.4343433380127
3.5151515007019 3.5151515007019
3.59595966339111 3.59595966339111
3.67676758766174 3.67676758766174
3.75757575035095 3.75757575035095
3.83838391304016 3.83838391304016
3.91919183731079 3.91919183731079
4 4
};
\addplot [semithick, black, dashed]
table {%
-4 -4
-3.91919183731079 -3.91919183731079
-3.83838391304016 -3.83838391304016
-3.75757575035095 -3.75757575035095
-3.67676758766174 -3.67676758766174
-3.59595966339111 -3.59595966339111
-3.5151515007019 -3.5151515007019
-3.4343433380127 -3.4343433380127
-3.35353541374207 -3.35353541374207
-3.27272725105286 -3.27272725105286
-3.19191908836365 -3.19191908836365
-3.11111116409302 -3.11111116409302
-3.03030300140381 -3.03030300140381
-2.9494948387146 -2.9494948387146
-2.86868691444397 -2.86868691444397
-2.78787875175476 -2.78787875175476
-2.70707082748413 -2.70707082748413
-2.62626266479492 -2.62626266479492
-2.54545450210571 -2.54545450210571
-2.46464657783508 -2.46464657783508
-2.38383841514587 -2.38383841514587
-2.30303025245667 -2.30303025245667
-2.22222232818604 -2.22222232818604
-2.14141416549683 -2.14141416549683
-2.06060600280762 -2.06060600280762
-1.9797979593277 -1.9797979593277
-1.89898991584778 -1.89898991584778
-1.81818187236786 -1.81818187236786
-1.73737370967865 -1.73737370967865
-1.65656566619873 -1.65656566619873
-1.57575762271881 -1.57575762271881
-1.4949494600296 -1.4949494600296
-1.41414141654968 -1.41414141654968
-1.33333337306976 -1.33333337306976
-1.25252521038055 -1.25252521038055
-1.17171716690063 -1.17171716690063
-1.09090912342072 -1.09090912342072
-1.01010096073151 -1.01010096073151
-0.929292917251587 -0.929292917251587
-0.848484873771667 -0.848484873771667
-0.767676770687103 -0.767676770687103
-0.686868667602539 -0.686868667602539
-0.60606062412262 -0.60606062412262
-0.525252521038055 -0.525252521038055
-0.444444447755814 -0.444444447755814
-0.363636374473572 -0.363636374473572
-0.282828271389008 -0.282828271389008
-0.202020198106766 -0.202020198106766
-0.121212124824524 -0.121212124824524
-0.0404040440917015 -0.0404040440917015
0.0404040440917015 0.0404040440917015
0.121212124824524 0.121212124824524
0.202020198106766 0.202020198106766
0.282828271389008 0.282828271389008
0.363636374473572 0.363636374473572
0.444444447755814 0.444444447755814
0.525252521038055 0.525252521038055
0.60606062412262 0.60606062412262
0.686868667602539 0.686868667602539
0.767676770687103 0.767676770687103
0.848484873771667 0.848484873771667
0.929292917251587 0.929292917251587
1.01010096073151 1.01010096073151
1.09090912342072 1.09090912342072
1.17171716690063 1.17171716690063
1.25252521038055 1.25252521038055
1.33333337306976 1.33333337306976
1.41414141654968 1.41414141654968
1.4949494600296 1.4949494600296
1.57575762271881 1.57575762271881
1.65656566619873 1.65656566619873
1.73737370967865 1.73737370967865
1.81818187236786 1.81818187236786
1.89898991584778 1.89898991584778
1.9797979593277 1.9797979593277
2.06060600280762 2.06060600280762
2.14141416549683 2.14141416549683
2.22222232818604 2.22222232818604
2.30303025245667 2.30303025245667
2.38383841514587 2.38383841514587
2.46464657783508 2.46464657783508
2.54545450210571 2.54545450210571
2.62626266479492 2.62626266479492
2.70707082748413 2.70707082748413
2.78787875175476 2.78787875175476
2.86868691444397 2.86868691444397
2.9494948387146 2.9494948387146
3.03030300140381 3.03030300140381
3.11111116409302 3.11111116409302
3.19191908836365 3.19191908836365
3.27272725105286 3.27272725105286
3.35353541374207 3.35353541374207
3.4343433380127 3.4343433380127
3.5151515007019 3.5151515007019
3.59595966339111 3.59595966339111
3.67676758766174 3.67676758766174
3.75757575035095 3.75757575035095
3.83838391304016 3.83838391304016
3.91919183731079 3.91919183731079
4 4
};
\end{axis}

\end{tikzpicture}
    }\\ 
    \quad \textbf{\small(a)} & \quad \textbf{\small(b)} & \quad \textbf{\small(c)}
    \end{tabular}
    \caption{Here we display (a) $\bv^\btheta$, a CPA velocity field, (b) DiTAC (solid line) versus ~GELU (dashed line), and (c) a Leaky DiTAC (solid line) versus the identity function (dashed line).
    The two DiTAC functions were derived from the CPAB transformation $T^\btheta$ that corresponds
    to $\bv^\btheta$ in (a).}    
    \label{fig:cpab_act_method}
\end{figure}

To stabilize the training and to prevent learning too-extreme transformations, we also use a regularization  over the velocity fields. Concretely, our regularization term (which was also used in,  \eg,~\cite{Skafte:CVPR:2018:DDTN,Shapira:NeurIPS:2019:DTAN,Martinez:ICML:2022:closed}) is
\begin{align}	  
\Lcal_{\mathrm{reg}} = \sum\nolimits_{l=1}^{L} {\btheta_l}^T \bSigma_{\mathrm{CPA}}^{-1} \btheta_l
\end{align}

\noindent where $L$ is the number of activation layers in the network, $\btheta \in \RR^d$ is the DiTAC parameters, and $\bSigma_{\mathrm{CPA}}^{-1}$ is a $d \times d$ covariance matrix associated with a Gaussian smoothness prior (proposed in~\cite{Freifeld:ICCV:2015:CPAB}) over CPA velocity fields. 
That matrix has two hyperparameters: $\lambda_{var}$, which controls the variance of the velocity fields, and $\lambda_{smooth}$, which controls the similarity of the velocities in different subintervals and hence the smoothness (in the machine-learning sense) of the field.

\subsection{How to Drastically Reduce the Computational Cost}

As is usual in DL, training involves a very high number of calls to the AFs.
Consider a tensor of size $(b,c,h,w)$, where $b$ is the batch size, $c$ is the number of channels, and $(h,w)$ is the height and width. Applying a CPAB transformation to each entry in the tensor will naturally require evaluating it $b\cdot c\cdot h\cdot w$ times. \Eg, the AF of the last bottleneck block of ResNet-50~\cite{He:2016:resnet}, with a batch size of 32, operates over $\sim$800K entries.
Thus, and although CPAB transformations provide an efficient solution for representing diffeomorphisms, 
using such transformations naively here could still incur a
significant computational cost during training and be too slow.
Luckily, there is a better way. Below we provide a solution that, during the learning, dramatically alleviates this cost. 
Moreover, during inference that solution renders DiTAC just as efficient as other AFs.

To drastically reduce the cost during learning, we quantize the interval $[a,b]$ (on which the CPAB transformation is applied) to $n$ discrete values uniformly. Although losing some information, quantizing activations in a neural network can show little to no impact on accuracy when using a large enough number of elements (in our case, usually $2^8$ suffice)~\cite{Nagel:2021:whitepaperquant,Finder:NIPS:2022:wcc}.
In this approach, we use the CPAB transformation on the set of quantized elements and create a lookup table, which we can then use to transform the values of all the entries in the input tensor. That is, we output $y_i=T^\btheta(Q(x_i))$ where $Q(\cdot)$ is the quantization function and $Q(x_i)\in\{ a+k\Delta \}_{k=0}^n$ for $\Delta=\tfrac{b-a}{n}$. For backpropagation, we utilize a variation of the Straight Through Estimator \cite{Bengio:2013:STE}. 
We calculate the CPAB derivatives only for the outputs of the quantized values, which we then broadcast as derivative estimations for the $x_i$'s,  
\begin{align}
    \frac{\partial T^{\btheta}(x_i)}{\partial x_i} \approx \frac{\partial T^{\btheta}(q_j)}{\partial q_j}, \enspace
    \frac{\partial T^{\btheta}(x_i)}{\partial \theta_\ell} \approx \frac{\partial T^{\btheta}(q_j)}{\partial \theta_\ell}, \quad\text{where } q_j=Q(x_i).
\end{align}

Revisiting that ResNet-50 example, we need to transform only a much smaller set of entries (\eg, $2^{10}=1024\ll$ 800K) for the same input while achieving, empirically, nearly identical results.
During the learning, we (quickly) build such a lookup table every time $\btheta$ changes. 
Once the learning is done and before the inference, a single lookup table (per each DiTAC function) is calculated
and then reused during inference as many times as needed.
An experiment measuring the computational costs during inference is provided in the Appendix.

\section{Results}
\label{sec:results}

We present an extensive evaluation of DiTAC on several challenging tasks using state-of-the-art architectures. In \autoref{subsec:toy_data} we analyze DiTAC's performance in classification and regression tasks on synthetic and toy data. In \autoref{subsec:real_data} we continue evaluating DiTAC on real-data (\eg: ImageNet-1K~\cite{Russakovsky:IJCV:2015:imagenet};  ADE20K~\cite{Zhou:IJCV:2019:ade20k}), on image classification, semantic segmentation, and image generation tasks.
In addition we show comprehensive comparisons between DiTAC and the following fixed AFs: ReLU, LReLU, GELU, ELU, Softplus, Mish, and the following TAFs: Swish, PReLU, PDELU and Meta-ACON, on various tasks. 
In all of our experiments, the trainable parameters of DiTAC (and of other TAF competitors) are trained without weight decay to avoid pushing the learned values to zero during training.
We found that a 10-cell partition of $\Omega$ 
(the domain of the CPAB function) suffices for good results. This translates to an addition of only 9 trainable parameters in each DiTAC instance in the recipient network.

\subsection{Toy Data}\label{subsec:toy_data}

In this section, we report the results of several experiments in classification and regression tasks on simple datasets. We use simple MLPs with at most two dense layers, changing only the AF between configurations.
We run each configuration with several learning rates and select the best performance for each competitor. In all toy-data analyses, we omit Meta-ACON (as it works in a channel-wise manner, which is impossible to apply in a fully-connected network)
while for PDELU we use its channel-shared version. 

\begin{figure}[t] 
    \centering
    \setlength{\fboxsep}{0pt}
    \setlength{\fboxrule}{0.1pt}
    \captionsetup[sub]{font=small}
     \begin{subfigure}[b]{0.28\textwidth}
         \centering
         \framebox{\includegraphics[width=\textwidth]{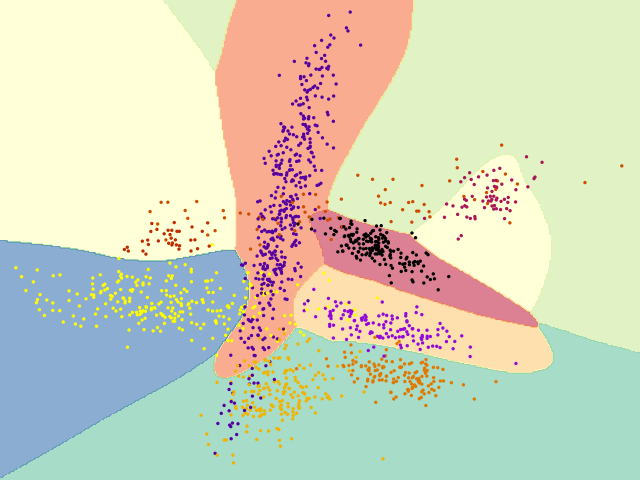}}
         \caption{DiTAC}
     \end{subfigure}
     \quad \quad
     \begin{subfigure}[b]{0.28\textwidth}
         \centering
         \framebox{\includegraphics[width=\textwidth]{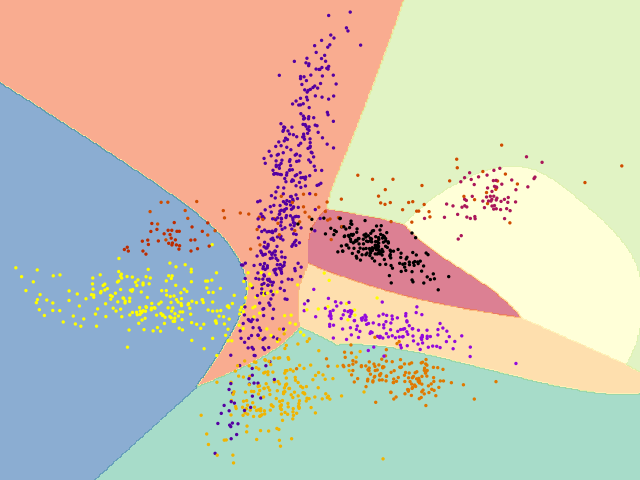}}
         \caption{Swish}
     \end{subfigure}
     \quad \quad
     \begin{subfigure}[b]{0.28\textwidth}
         \centering
         \framebox{\includegraphics[width=\textwidth]{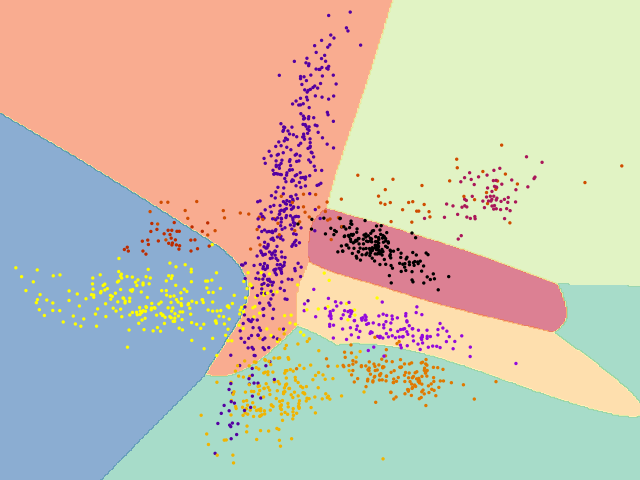}}
         \caption{GELU}
     \end{subfigure}
        \caption{Decision boundaries learned by DiTAC, Swish, and GELU in a 2D-GMM classification, along with test-batch data points colored by their ground-truth classes. Evidently, DiTAC learns more accurate boundaries and identifies more classes.}
        \label{fig:2d_gauss_db}
\end{figure}

\begin{table}[hbt!]  
    \centering
    \setlength{\tabcolsep}{5pt}
    \caption{A comparison of DiTAC with existing AFs and TAFs. We report the top-1 classification accuracy (in \%) along with the number of model parameters used by each AF,
    on the 2D-GMM and MNIST datasets. The network is a simple MLP.}
    \label{tab:classification_2d_gaussian_mnist}
    \begin{tabular}{lcccc}
         \toprule
         AF/TAF & \multicolumn{2}{c}{2D-GMM} & \multicolumn{2}{c}{MNIST} \\
                & Param. & Top-1 & Param. & Top-1 \\
         \midrule
         ReLU & 11,410 & 84.0 & 109,386 & 97.75 \\
         LReLU & 11,410 & 84.0  & 109,386 & 97.77 \\
         GELU & 11,410 & 84.0  & 109,386 & 97.99 \\
         ELU & 11,410 & 84.0  & 109,386 & 97.82 \\
         Softplus & 11,410 & 81.4  & 109,386 & 97.8  \\
         Mish & 11,410 & 84.0  & 109,386 & 97.97 \\
         Swish & 11,412 & 86.8  & 109,388 & 97.93 \\
         PReLU & 11,412 & 83.93  & 109,388 & 97.84 \\
         PDELU & 11,412 & 83.86  & 109,388 & 97.68 \\
         \midrule
         DiTAC & 11,428 & \textbf{89.06} & 109,404 & \textbf{98.29}  \\
         \bottomrule
    \end{tabular}
\end{table}


\subsubsection{Classification.}
For this task, we use two datasets, a two-dimensional Gaussian-Mixture-Model (2D-GMM) and MNIST.
We construct a 2D GMM as a 10-component mixture, where the parameters of each Gaussian component
(\ie, the mean vector and covariance matrix) are drawn from a normal-inverse-Wishart distribution. We then draw $5\cdot10^3$ points from the GMM, with a split of $70\%/30\%$ for train/test data.

For each dataset, we use a simple MLP with two hidden layers. For MNIST we set their sizes to 128 and 64, and for 2D-GMM we set both to 100. Note that the number of trained parameters in DiTAC is negligible \wrt the network size. \Eg, the 2D-GMM network has $11,410$ trainable parameters, and by using DiTAC, we add only $18$ parameters (9 per each of the two AFs in that MLP) to the model. 
More details about the training procedure and data generation can be found in the Appendix.

The results appear in \autoref{tab:classification_2d_gaussian_mnist} and show a significant advantage to using DiTAC. In this comparison, we report the top-1 accuracy along with network size generated by each AF. 
In addition, in \autoref{fig:2d_gauss_db} we visualize the learned decision boundaries of DiTAC, Swish, and GELU in the 2D-GMM classification task. DiTAC outperforms the two state-of-the-art competitors by learning a better distinction between the classes. 


\subsubsection{Regression.}
Continuing the comparison, we now experiment with regression tasks of reconstructing one- and two-dimensional functions using an MLP with one hidden layer containing $30$ and $50$ neurons respectively. The first function is $\sin(\exp 6x)$, and the second is a sum of sines with various frequencies.
We train the model for 40K iterations, ensuring training convergence of
all of the competitors. More training details can be found in the Appendix. Here, we had to omit PDELU from the results table due to its unstable training process. 

\begin{table}[hbt!] 
    \centering
    \setlength{\tabcolsep}{5pt}
    \caption{Regression-task results of learning the one- and two-dimensional functions mentioned in the text, along with the number of model parameters used by each AF.}
    \begin{tabular}{lcccccc}
         \toprule
         AF/TAF & \multicolumn{3}{c}{\textbf{1D Func.}}
              & \multicolumn{3}{c}{\textbf{2D Func.}} \\
              & Param. & MSE $\downarrow$ & R2 $\uparrow$ & Param. & MSE $\downarrow$ & R2 $\uparrow$\\
         \midrule
         ReLU & 91 & 0.117  & 76.47 & 201 & 0.023  & 88.86 \\
         LReLU & 91 & 0.051  & 89.78 & 201 & 0.011  & 93.13 \\
         GELU & 91 & 0.138  & 72.73 & 201 & 0.054  & 65.27 \\
         ELU & 91 & 0.207 & 58.30 & 201 & 0.118  & 25.11 \\
         Softplus & 91 & 0.206  & 59.56 & 201 & 0.112  & 24.96  \\
         Mish & 91 & 0.175  & 66.71 & 201 & 0.114  & 27.28 \\
         Swish & 92 & 0.061  & 88.33 & 202 & 0.130  & 25.47 \\
         PReLU & 92 & 0.062  & 88.30 & 202 & 0.008  & 96.09 \\
         \midrule
         DiTAC & 100 & \textbf{0.020} & \textbf{96.26} & 210 & \textbf{0.004} & \textbf{98.37} \\
         \bottomrule
    \end{tabular}
    \label{tab:regressionn_1d_2d_functions}
\end{table}

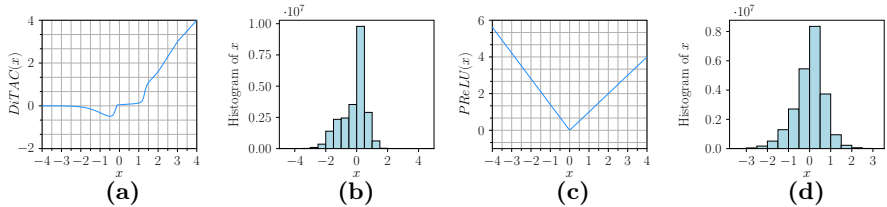
\begin{figure}[hbt!]  
    \renewcommand{\arraystretch}{0.0}
    \centering
    \LARGE
    \begin{tabular}{cccc}
    \scalebox{0.3}{
\begin{tikzpicture}

\definecolor{darkgray176}{RGB}{176,176,176}
\definecolor{dodgerblue}{RGB}{30,144,255}

\begin{axis}[
tick align=outside,
tick pos=left,
x grid style={darkgray176},
xlabel={\(\displaystyle x\)},
x label style={at={(axis description cs:0.5,-0.14)},anchor=north},
y label style={at={(axis description cs:-0.1,.5)},anchor=south},
xmajorgrids,
xmin=-4, xmax=4,
xtick style={color=black},
grid=both,
grid style={line width=.1pt, draw=gray!10},
major grid style={line width=.2pt,draw=gray!20},
minor x tick num=1,
minor y tick num=2,
x grid style={darkgray176},
xmin=-4, xmax=4,
xtick={-4,-3,-2,-1,0,1,2,3,4},
y grid style={darkgray176},
ylabel style={rotate=0.0},
ylabel={\(\displaystyle DiTAC(x)\)},
ymajorgrids,
ymin=-2, ymax=4,
ytick style={color=black}
]
\addplot [semithick, dodgerblue]
table {%
-4 -0.00012671947479248
-3.91919183731079 -0.00017415032198187
-3.83838391304016 -0.000237708140048198
-3.75757575035095 -0.000322291336487979
-3.67676758766174 -0.000434250541729853
-3.59595966339111 -0.000581064610742033
-3.5151515007019 -0.000772497849538922
-3.4343433380127 -0.00102003407664597
-3.35353541374207 -0.00133783894125372
-3.27272725105286 -0.00174294819589704
-3.19191908836365 -0.00225554686039686
-3.11111116409302 -0.00289930240251124
-3.03030300140381 -0.00370162911713123
-2.9494948387146 -0.00468567665666342
-2.86868691444397 -0.00587846897542477
-2.78787875175476 -0.00733170472085476
-2.70707082748413 -0.00906865671277046
-2.62626266479492 -0.0111369686201215
-2.54545450210571 -0.0135959731414914
-2.46464657783508 -0.016455989331007
-2.38383841514587 -0.0202601570636034
-2.30303025245667 -0.0250044818967581
-2.22222232818604 -0.0306842625141144
-2.14141416549683 -0.0374241657555103
-2.06060600280762 -0.0453874692320824
-1.9797979593277 -0.0547119863331318
-1.89898991584778 -0.0655696764588356
-1.81818187236786 -0.0781468972563744
-1.73737370967865 -0.0925633534789085
-1.65656566619873 -0.108973160386086
-1.57575762271881 -0.127442196011543
-1.4949494600296 -0.148139029741287
-1.41414141654968 -0.1710434705019
-1.33333337306976 -0.196223214268684
-1.25252521038055 -0.223760649561882
-1.17171716690063 -0.253400921821594
-1.09090912342072 -0.2849982380867
-1.01010096073151 -0.317771285772324
-0.929292917251587 -0.351463615894318
-0.848484873771667 -0.384633004665375
-0.767676770687103 -0.416173726320267
-0.686868667602539 -0.444970995187759
-0.60606062412262 -0.468328922986984
-0.525252521038055 -0.485284119844437
-0.444444447755814 -0.485423564910889
-0.363636374473572 -0.457756042480469
-0.282828271389008 -0.368344873189926
-0.202020198106766 -0.165120393037796
-0.121212124824524 0.0361363627016544
-0.0404040440917015 0.0482251420617104
0.0404040440917015 0.0530183650553226
0.121212124824524 0.057759553194046
0.202020198106766 0.0625813379883766
0.282828271389008 0.0675646960735321
0.363636374473572 0.072632908821106
0.444444447755814 0.0776963457465172
0.525252521038055 0.0828523561358452
0.60606062412262 0.0879477635025978
0.686868667602539 0.0934498384594917
0.767676770687103 0.0996963977813721
0.848484873771667 0.107520841062069
0.929292917251587 0.118034273386002
1.01010096073151 0.133306428790092
1.09090912342072 0.160954192280769
1.17171716690063 0.22110541164875
1.25252521038055 0.405051499605179
1.33333337306976 0.731617510318756
1.41414141654968 0.953819394111633
1.4949494600296 1.08440935611725
1.57575762271881 1.16661202907562
1.65656566619873 1.24693524837494
1.73737370967865 1.32451844215393
1.81818187236786 1.4044201374054
1.89898991584778 1.48607671260834
1.9797979593277 1.57809042930603
2.06060600280762 1.67962288856506
2.14141416549683 1.788694024086
2.22222232818604 1.90582406520844
2.30303025245667 2.01862740516663
2.38383841514587 2.13507318496704
2.46464657783508 2.24709916114807
2.54545450210571 2.36260342597961
2.62626266479492 2.47361993789673
2.70707082748413 2.58832836151123
2.78787875175476 2.70275402069092
2.86868691444397 2.81290555000305
2.9494948387146 2.92684888839722
3.03030300140381 3.02660131454468
3.11111116409302 3.10821175575256
3.19191908836365 3.18966341018677
3.27272725105286 3.27098441123962
3.35353541374207 3.35219764709473
3.4343433380127 3.4333233833313
3.5151515007019 3.51437902450562
3.59595966339111 3.59537863731384
3.67676758766174 3.67633318901062
3.75757575035095 3.75725340843201
3.83838391304016 3.8381462097168
3.91919183731079 3.91901755332947
4 3.99987316131592
};
\end{axis}

\end{tikzpicture}
    } &
    \scalebox{0.3}{
\begin{tikzpicture}

\definecolor{darkgray176}{RGB}{176,176,176}
\definecolor{lightblue}{RGB}{173,216,230}

\begin{axis}[
tick align=outside,
tick pos=left,
x grid style={darkgray176},
xlabel={\(\displaystyle x\)},
x label style={at={(axis description cs:0.5,-0.14)},anchor=north},
y label style={at={(axis description cs:-0.22,.5)},anchor=south},
xmin=-5, xmax=5,
xtick style={color=black},
y grid style={darkgray176},
ylabel={Histogram of \(\displaystyle x\)},
ymin=0, ymax=20552182.35,
ytick style={color=black},
ytick={0,5000000,10000000,15000000,20000000},
yticklabels={0.00,0.25,0.50,0.75,1.00,1.25,1.50,1.75,2.00,2.25}
]
\draw[draw=black,fill=lightblue] (axis cs:-5,0) rectangle (axis cs:-4.5,90);
\draw[draw=black,fill=lightblue] (axis cs:-4.5,0) rectangle (axis cs:-4,91);
\draw[draw=black,fill=lightblue] (axis cs:-4,0) rectangle (axis cs:-3.5,1693);
\draw[draw=black,fill=lightblue] (axis cs:-3.5,0) rectangle (axis cs:-3,35679);
\draw[draw=black,fill=lightblue] (axis cs:-3,0) rectangle (axis cs:-2.5,185429);
\draw[draw=black,fill=lightblue] (axis cs:-2.5,0) rectangle (axis cs:-2,715905);
\draw[draw=black,fill=lightblue] (axis cs:-2,0) rectangle (axis cs:-1.5,2790702);
\draw[draw=black,fill=lightblue] (axis cs:-1.5,0) rectangle (axis cs:-1,4691509);
\draw[draw=black,fill=lightblue] (axis cs:-1,0) rectangle (axis cs:-0.5,4904677);
\draw[draw=black,fill=lightblue] (axis cs:-0.5,0) rectangle (axis cs:0,7069204);
\draw[draw=black,fill=lightblue] (axis cs:0,0) rectangle (axis cs:0.5,19573507);
\draw[draw=black,fill=lightblue] (axis cs:0.5,0) rectangle (axis cs:1,5800784);
\draw[draw=black,fill=lightblue] (axis cs:1,0) rectangle (axis cs:1.5,1232590);
\draw[draw=black,fill=lightblue] (axis cs:1.5,0) rectangle (axis cs:2,60477);
\draw[draw=black,fill=lightblue] (axis cs:2,0) rectangle (axis cs:2.5,384);
\draw[draw=black,fill=lightblue] (axis cs:2.5,0) rectangle (axis cs:3,96);
\draw[draw=black,fill=lightblue] (axis cs:3,0) rectangle (axis cs:3.5,95);
\draw[draw=black,fill=lightblue] (axis cs:3.5,0) rectangle (axis cs:4,90);
\draw[draw=black,fill=lightblue] (axis cs:4,0) rectangle (axis cs:4.5,89);
\draw[draw=black,fill=lightblue] (axis cs:4.5,0) rectangle (axis cs:5,84);
\end{axis}

\end{tikzpicture}
    } &
    \scalebox{0.3}{
\begin{tikzpicture}

\definecolor{darkgray176}{RGB}{176,176,176}
\definecolor{dodgerblue}{RGB}{30,144,255}

\begin{axis}[
tick align=outside,
tick pos=left,
x grid style={darkgray176},
xlabel={\(\displaystyle x\)},
x label style={at={(axis description cs:0.5,-0.14)},anchor=north},
y label style={at={(axis description cs:-0.1,.5)},anchor=south},
xmajorgrids,
xmin=-4, xmax=4,
xtick style={color=black},
grid=both,
grid style={line width=.1pt, draw=gray!10},
major grid style={line width=.2pt,draw=gray!20},
minor x tick num=1,
minor y tick num=2,
x grid style={darkgray176},
xmin=-4, xmax=4,
xtick={-4,-3,-2,-1,0,1,2,3,4},
y grid style={darkgray176},
ylabel style={rotate=0.0},
ylabel={\(\displaystyle PReLU(x)\)},
ymajorgrids,
ymin=-1, ymax=6,
ytick style={color=black}
]
\addplot [semithick, dodgerblue]
table {%
-4 5.62154483795166
-3.91919183731079 5.5079779624939
-3.83838391304016 5.39441204071045
-3.75757575035095 5.28084516525269
-3.67676758766174 5.16727828979492
-3.59595966339111 5.05371189117432
-3.5151515007019 4.94014549255371
-3.4343433380127 4.82657861709595
-3.35353541374207 4.71301221847534
-3.27272725105286 4.59944581985474
-3.19191908836365 4.48587894439697
-3.11111116409302 4.37231254577637
-3.03030300140381 4.25874614715576
-2.9494948387146 4.145179271698
-2.86868691444397 4.03161287307739
-2.78787875175476 3.91804623603821
-2.70707082748413 3.80448007583618
-2.62626266479492 3.690913438797
-2.54545450210571 3.57734656333923
-2.46464657783508 3.46378040313721
-2.38383841514587 3.35021352767944
-2.30303025245667 3.23664689064026
-2.22222232818604 3.12308073043823
-2.14141416549683 3.00951385498047
-2.06060600280762 2.89594721794128
-1.9797979593277 2.78238081932068
-1.89898991584778 2.66881418228149
-1.81818187236786 2.55524778366089
-1.73737370967865 2.4416811466217
-1.65656566619873 2.32811450958252
-1.57575762271881 2.21454811096191
-1.4949494600296 2.10098147392273
-1.41414141654968 1.98741483688354
-1.33333337306976 1.87384831905365
-1.25252521038055 1.76028168201447
-1.17171716690063 1.64671516418457
-1.09090912342072 1.53314864635468
-1.01010096073151 1.41958200931549
-0.929292917251587 1.3060154914856
-0.848484873771667 1.1924489736557
-0.767676770687103 1.07888233661652
-0.686868667602539 0.965315759181976
-0.60606062412262 0.851749241352081
-0.525252521038055 0.738182663917542
-0.444444447755814 0.624616086483002
-0.363636374473572 0.511049568653107
-0.282828271389008 0.397482961416245
-0.202020198106766 0.283916413784027
-0.121212124824524 0.170349851250648
-0.0404040440917015 0.0567832849919796
0.0404040440917015 0.0404040440917015
0.121212124824524 0.121212124824524
0.202020198106766 0.202020198106766
0.282828271389008 0.282828271389008
0.363636374473572 0.363636374473572
0.444444447755814 0.444444447755814
0.525252521038055 0.525252521038055
0.60606062412262 0.60606062412262
0.686868667602539 0.686868667602539
0.767676770687103 0.767676770687103
0.848484873771667 0.848484873771667
0.929292917251587 0.929292917251587
1.01010096073151 1.01010096073151
1.09090912342072 1.09090912342072
1.17171716690063 1.17171716690063
1.25252521038055 1.25252521038055
1.33333337306976 1.33333337306976
1.41414141654968 1.41414141654968
1.4949494600296 1.4949494600296
1.57575762271881 1.57575762271881
1.65656566619873 1.65656566619873
1.73737370967865 1.73737370967865
1.81818187236786 1.81818187236786
1.89898991584778 1.89898991584778
1.9797979593277 1.9797979593277
2.06060600280762 2.06060600280762
2.14141416549683 2.14141416549683
2.22222232818604 2.22222232818604
2.30303025245667 2.30303025245667
2.38383841514587 2.38383841514587
2.46464657783508 2.46464657783508
2.54545450210571 2.54545450210571
2.62626266479492 2.62626266479492
2.70707082748413 2.70707082748413
2.78787875175476 2.78787875175476
2.86868691444397 2.86868691444397
2.9494948387146 2.9494948387146
3.03030300140381 3.03030300140381
3.11111116409302 3.11111116409302
3.19191908836365 3.19191908836365
3.27272725105286 3.27272725105286
3.35353541374207 3.35353541374207
3.4343433380127 3.4343433380127
3.5151515007019 3.5151515007019
3.59595966339111 3.59595966339111
3.67676758766174 3.67676758766174
3.75757575035095 3.75757575035095
3.83838391304016 3.83838391304016
3.91919183731079 3.91919183731079
4 4
};
\end{axis}

\end{tikzpicture}
    } &
    \scalebox{0.3}{
\begin{tikzpicture}

\definecolor{darkgray176}{RGB}{176,176,176}
\definecolor{lightblue}{RGB}{173,216,230}

\begin{axis}[
tick align=outside,
tick pos=left,
x grid style={darkgray176},
xlabel={\(\displaystyle x\)},
x label style={at={(axis description cs:0.5,-0.14)},anchor=north},
y label style={at={(axis description cs:-0.22,.5)},anchor=south},
xmin=-3.79409098625183, xmax=3.53736257553101,
xtick style={color=black},
y grid style={darkgray176},
ylabel={Histogram of \(\displaystyle x\)},
ymin=0, ymax=17544551.85,
ytick style={color=black},
ytick={0,4000000,8000000,12000000,16000000},
yticklabels={0.0,0.2,0.4,0.6,0.8,1.0,1.2,1.4,1.6,1.8}
]
\draw[draw=black,fill=lightblue] (axis cs:-5,0) rectangle (axis cs:-4.5,0);
\draw[draw=black,fill=lightblue] (axis cs:-4.5,0) rectangle (axis cs:-4,0);
\draw[draw=black,fill=lightblue] (axis cs:-4,0) rectangle (axis cs:-3.5,1081);
\draw[draw=black,fill=lightblue] (axis cs:-3.5,0) rectangle (axis cs:-3,16447);
\draw[draw=black,fill=lightblue] (axis cs:-3,0) rectangle (axis cs:-2.5,92606);
\draw[draw=black,fill=lightblue] (axis cs:-2.5,0) rectangle (axis cs:-2,347671);
\draw[draw=black,fill=lightblue] (axis cs:-2,0) rectangle (axis cs:-1.5,1027380);
\draw[draw=black,fill=lightblue] (axis cs:-1.5,0) rectangle (axis cs:-1,2586206);
\draw[draw=black,fill=lightblue] (axis cs:-1,0) rectangle (axis cs:-0.5,5414469);
\draw[draw=black,fill=lightblue] (axis cs:-0.5,0) rectangle (axis cs:0,10903799);
\draw[draw=black,fill=lightblue] (axis cs:0,0) rectangle (axis cs:0.5,16709097);
\draw[draw=black,fill=lightblue] (axis cs:0.5,0) rectangle (axis cs:1,7465840);
\draw[draw=black,fill=lightblue] (axis cs:1,0) rectangle (axis cs:1.5,1896170);
\draw[draw=black,fill=lightblue] (axis cs:1.5,0) rectangle (axis cs:2,455591);
\draw[draw=black,fill=lightblue] (axis cs:2,0) rectangle (axis cs:2.5,119112);
\draw[draw=black,fill=lightblue] (axis cs:2.5,0) rectangle (axis cs:3,24176);
\draw[draw=black,fill=lightblue] (axis cs:3,0) rectangle (axis cs:3.5,3892);
\draw[draw=black,fill=lightblue] (axis cs:3.5,0) rectangle (axis cs:4,13);
\draw[draw=black,fill=lightblue] (axis cs:4,0) rectangle (axis cs:4.5,0);
\draw[draw=black,fill=lightblue] (axis cs:4.5,0) rectangle (axis cs:5,0);
\end{axis}

\end{tikzpicture}
    }\\
    \textbf{\ \ \small(a)} & \textbf{\ \ \ \small(b)} &   \textbf{\ \ \small(c)} &  \textbf{\ \ \ \small(d)}
    \end{tabular}
    \caption{We demonstrate DiTAC's expressiveness by showing the learned DiTAC (a) and its input-data histogram (b), and the learned PReLU (c) and its input-data histogram (d), in the 2D regression problem with a simple MLP.}
    \label{fig:2d_reg_learned_cpab}
\end{figure}

\autoref{tab:regressionn_1d_2d_functions} shows that when using DiTAC, the MLP fits more accurately to the functions, despite using only an insignificant increase in the number of trainable parameters.
DiTAC's expressiveness and flexibility become even more evident when visualizing the learned DiTAC function compared to PReLU (the runner-up) for the two-dimensional regression in \autoref{fig:2d_reg_learned_cpab}: while PReLU is restricted in the function shapes it can learn, DiTAC ended up learning a certain function curvature that fits to the problem.
This is also demonstrated qualitatively in \autoref{fig:2d_regression}, where it is evident that PReLU's elevation lines are more rigid than DiTAC's, which indicates its difficulty in fitting the smooth function.
A detailed analysis of the experiments above, together with evaluation on additional regression datasets, is presented in the Appendix.


\subsection{Real-World Data}\label{subsec:real_data}


\subsubsection{Small-scale Classification Experiment.}
We first conduct a small-scale classification experiment using ImageNet-50, a subset of 50 classes from ImageNet~\cite{Russakovsky:IJCV:2015:imagenet}.
We train several MobileNet-V3~\cite{Howard:2019:MobileNet-V3} and ResNet~\cite{He:2016:resnet} configurations of different sizes, and compare DiTAC to various competitors by replacing all AF instances in each architecture with each AF/TAF candidate.
As for ResNet, the training procedure is inspired by~\cite{Wightman:2021:resnet_training_params}. The models were trained for 300 epochs on 224$\times$224 resolution on a single GPU, using an AdamW optimizer. We use a batch size of 256, initial learning rate of $5\cdot 10^{-3}$, weight decay of 0.02 and a 5 epoch linear warmup.
With regards to MobileNet-V3, we follow the training setup proposed in~\cite{Howard:2019:MobileNet-V3} with a few adjustments, considering a smaller dataset and a single GPU. We trained the models for 300 epochs, using AdamW optimizer, batch size of 256, initial learning rate of $1\cdot 10^{-3}$, weight decay of $4\cdot 10^{-5}$ and a 5-epoch linear warmup.
We use the implementation of \texttt{timm}~\cite{Ross:2019:timm} for training and evaluation.

In~\autoref{tab:classification_in50_all_acts} we report the top-1 accuracy of the various AFs/TAFs. The results show that DiTAC consistently outperforms existing AFs and TAFs in a variety of model configurations and sizes, obtaining an improvement of up to $0.6\%$ in MobileNet-V3 and $0.12\%$ in ResNet.

\begin{table}[hbt!]  
    \caption{A comparison of DiTAC with existing AFs/TAFs on a small-scale classification problem. We show the classification top-1 accuracy (in \%) on ImageNet-50 on various MobileNet-V3 (MN) and ResNet (RN) configurations. 
    For PDELU, we were unable to report the accuracy in several cases due to its unstable training process.}
    \centering
    \setlength{\tabcolsep}{5pt}
    \begin{tabular}{lcccccc}
         \toprule
         AF/TAF     & MN 0.5 & MN 0.75 & MN 1.0 & RN-18 & RN-34 & RN-50 \\
         \midrule
         ReLU     & 77.36  & 79.76  & 80.83 & 89.64 & 91.96 & 93.00 \\
         LReLU    & 77.60  & 79.88  & 81.99 & 89.64 & 91.60 & \textbf{93.24} \\
         GELU     & 79.32  & 82.12  & 83.51 & 89.92 & 92.12 & 92.84 \\
         ELU      & 76.00  & 79.00  & 80.83 & 88.12 & 92.20 & 93.12 \\
         Softplus & 75.12  & 79.40  & 80.43 & 89.04 & 91.36 & 92.40 \\
         Mish     & 78.96  & 82.00  & 83.19 & 88.96 & 92.56 & 92.60 \\
         Swish    & 77.20  & 81.52  & 82.99 & 88.48 & 91.40 & 92.32 \\
         PReLU    & 77.92  & 81.16 & 81.68 & 88.96 & 91.60 & 92.56 \\
         PDELU    & 77.00  & 80.96  & 82.28 & -- & -- & -- \\
         Meta-ACON     & 79.20 & 81.96  & 84.04 & 89.64 & 92.08 & 92.20 \\
         \midrule
         DiTAC     & \textbf{79.92} & \textbf{82.48}  & \textbf{84.19} & \textbf{90.04} & \textbf{92.64} & \textbf{93.24} \\
         \bottomrule
    \end{tabular}
    \label{tab:classification_in50_all_acts}
\end{table}


\subsubsection{Classification.}
Next, we train ConvNeXt-T~\cite{Liu:CVPR:2022:convnext} and Swin-T~\cite{Liu:ICCV:2021:swin} with DiTAC on ImageNet-100 and ImageNet-1K~\cite{Russakovsky:IJCV:2015:imagenet}, and compare to its baseline version with a GELU function.
As for the ImageNet-1K experiment, we follow the regular training setup mentioned in~\cite{Liu:CVPR:2022:convnext} and~\cite{Liu:ICCV:2021:swin} for ConvNeXt and Swin  training procedures, respectively. 
Both models were trained using 300-epochs schedule and 4-GPUs setup. We use
an AdamW optimizer with a momentum of 0.9, a weight decay of 0.05 and 20 epochs of linear warmup.
In ConvNeXT, we set the initial learning rate to $4\cdot 10^{-3}$, and use an effective batch size of 4096.
As for Swin, we use an initial learning rate of $1\cdot 10^{-3}$ and an effective batch size of 1024.
The training details of ImageNet-100 appear in the Appendix, along with additional information about the training process on ImageNet-1K.
We use the implementation of \texttt{timm}~\cite{Ross:2019:timm} for training and evaluation.

In \autoref{tab:classification_in100_in1k} we show the top-1 accuracy of DiTAC compared to the baseline AF. 
In ImageNet-100 we train the baseline AF ourselves, whereas in ImageNet-1K we compare to the published results in ConvNeXT and Swin papers.
DiTAC noticeably surpasses the baseline performance, gaining an improvement of $0.3\%$ for ConvNeXT-T and $0.2\%$ for Swin-T in ImageNet-1K. DiTAC's advantage is consistent also in the ImageNet-100 experiments.

\begin{table}[hbt!]  
    \centering
    \setlength{\tabcolsep}{5pt}
    \caption{Classification top-1 accuracy (in \%) on ImageNet-100 (IN-100) and ImageNet-1K (IN-1K) using a 300-epoch training schedule, on ConvNeXt-T and Swin-T. We compare DiTAC with GELU, the baseline AF, in both architectures.}
    \begin{tabular}{lcccccc}
         \toprule
         Configuration & IN-100 & IN-1K \\
         \midrule
         ConvNeXt-T & 92.4  & 82.1 \\
         ConvNeXt-T+DiTAC & \textbf{92.5} & \textbf{82.4} \\
         \midrule
         Swin-T & 91.4  & 81.3 \\
         Swin-T+DiTAC & \textbf{91.7} & \textbf{81.5} \\
         \bottomrule
    \end{tabular}
    \label{tab:classification_in100_in1k}
\end{table}


\subsubsection{Semantic Segmentation.}
We extend our research and evaluate DiTAC on a semantic segmentation task by comparing the performance of the baseline AF (ReLU) with DiTAC's and GELU's on the segmentation-framework.
In the two first experiments we train UperNet~\cite{Xiao:2018:UperNet} as a segmentation framework on the ADE20K dataset~\cite{Zhou:IJCV:2019:ade20k}, where in one configuration we use ConvNeXT-T and in the other we use Swin-T as the backbone.
In the third experiment we train a PSPNet~\cite{Zhao:2017:PSPNet} segmentation framework using ResNet-50~\cite{He:2016:resnet} as the backbone, on the Cityscapes dataset~\cite{Cordts:2016:cityscapes}.
In all experiments in this section we use the implementation of \texttt{mmseg}~\cite{MMSegmentation:2020:mmsegmentation} for training and evaluation, and compare DiTAC's and GELU's performance to \texttt{mmseg}'s published results.
More details about the training procedures can be found in the Appendix.

In \autoref{tab:segmentation} we report the mean intersection over union (mIoU) of the described experiments. We observe an improved performance by replacing the baseline AF of each segmentation framework with DiTAC. We notice that GELU behaves differently on each configuration, as opposed to its consistent success in classification models such as ConvNeXT and Swin-Transformer.
\begin{table}[hbt!]  
    \centering
    \setlength{\tabcolsep}{5pt}
    \caption{Segmentation results on UperNet with ConvNeXT-T and Swin-T backbones on ADE20K dataset, and PSPNet with ResNet-50 backbone on Cityscapes dataset. We compare DiTAC with the baseline AF and GELU.}
    \begin{tabular}{lcc}
         \toprule
         Configuration & Dataset & mIoU \\
         \midrule
         ConvNeXT-T + UperNet & ADE20K & 46.1  \\
         ConvNeXT-T + UperNet+GELU & ADE20K & 45.9 \\
         ConvNeXT-T + UperNet+DiTAC & ADE20K & \textbf{46.2} \\
         \midrule
         Swin-T + UperNet & ADE20K & 44.4  \\
         Swin-T + UperNet+GELU & ADE20K & \textbf{44.7} \\
         Swin-T + UperNet+DiTAC & ADE20K & \textbf{44.7} \\
         \midrule
         ResNet-50 + PSPNet & Cityscapes & 77.9  \\
         ResNet-50 + PSPNet+GELU & Cityscapes & 78.0 \\
         ResNet-50 + PSPNet+DiTAC & Cityscapes & \textbf{78.4} \\
         \bottomrule
    \end{tabular}
    \label{tab:segmentation}
\end{table}

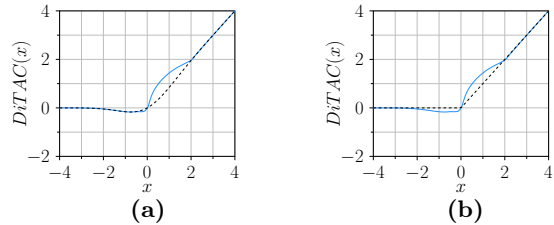
\begin{figure}[hbt!]  
    \renewcommand{\arraystretch}{0.04}
    \centering
    \huge
    \begin{tabular}{cc}
    \scalebox{0.34}{
\begin{tikzpicture}

\definecolor{darkgray176}{RGB}{190,190,190}
\definecolor{dodgerblue}{RGB}{30,144,255}

\begin{axis}[
tick align=outside,
tick pos=left,
x grid style={darkgray176},
xlabel={\(\displaystyle x\)},
xmin=-4, xmax=4,
xtick style={color=black},
y grid style={darkgray176},
ylabel={\(\displaystyle DiTAC(x)\)},
ymin=-2, ymax=4,
ytick style={color=black},
y label style={at={(axis description cs:-0.13,.5)},anchor=south},
x label style={at={(axis description cs:.5,-0.15)}},
grid=both,
grid style={line width=.1pt, draw=gray!10},
major grid style={line width=.2pt,draw=gray!20},
minor x tick num=1,
minor y tick num=1,
]
\addplot [semithick, dodgerblue]
table {%
-4 -0.00012671947479248
-3.91919183731079 -0.00017415032198187
-3.83838391304016 -0.000237708140048198
-3.75757575035095 -0.000322291336487979
-3.67676758766174 -0.000434250541729853
-3.59595966339111 -0.000581064610742033
-3.5151515007019 -0.000772497849538922
-3.4343433380127 -0.00102003407664597
-3.35353541374207 -0.00133783894125372
-3.27272725105286 -0.00174294819589704
-3.19191908836365 -0.00225554686039686
-3.11111116409302 -0.00289930240251124
-3.03030300140381 -0.00370162911713123
-2.9494948387146 -0.00469404365867376
-2.86868691444397 -0.0059120487421751
-2.78787875175476 -0.00739550217986107
-2.70707082748413 -0.00918773841112852
-2.62626266479492 -0.011336050927639
-2.54545450210571 -0.0138899739831686
-2.46464657783508 -0.0169011298567057
-2.38383841514587 -0.0204212907701731
-2.30303025245667 -0.0245009362697601
-2.22222232818604 -0.0291870050132275
-2.14141416549683 -0.0345202684402466
-2.06060600280762 -0.0405327528715134
-1.9797979593277 -0.047244131565094
-1.89898991584778 -0.0546584464609623
-1.81818187236786 -0.0627603307366371
-1.73737370967865 -0.0715113654732704
-1.65656566619873 -0.0808464661240578
-1.57575762271881 -0.0906704068183899
-1.4949494600296 -0.100854992866516
-1.41414141654968 -0.11123663187027
-1.33333337306976 -0.121614933013916
-1.25252521038055 -0.131752207875252
-1.17171716690063 -0.141373917460442
-1.09090912342072 -0.150170683860779
-1.01010096073151 -0.157801449298859
-0.929292917251587 -0.163898140192032
-0.848484873771667 -0.168071269989014
-0.767676770687103 -0.169917285442352
-0.686868667602539 -0.169026538729668
-0.60606062412262 -0.164992272853851
-0.525252521038055 -0.157420203089714
-0.444444447755814 -0.156385242938995
-0.363636374473572 -0.157876238226891
-0.282828271389008 -0.157831504940987
-0.202020198106766 -0.154180511832237
-0.121212124824524 -0.132459253072739
-0.0404040440917015 -0.0724878832697868
0.0404040440917015 0.0462098158895969
0.121212124824524 0.301163136959076
0.202020198106766 0.517434775829315
0.282828271389008 0.67666620016098
0.363636374473572 0.805009067058563
0.444444447755814 0.914732694625854
0.525252521038055 1.01143300533295
0.60606062412262 1.10082602500916
0.686868667602539 1.18120896816254
0.767676770687103 1.25385451316833
0.848484873771667 1.32152998447418
0.929292917251587 1.38503205776215
1.01010096073151 1.44422662258148
1.09090912342072 1.4997376203537
1.17171716690063 1.55112314224243
1.25252521038055 1.59924817085266
1.33333337306976 1.64440560340881
1.41414141654968 1.6868052482605
1.4949494600296 1.7264267206192
1.57575762271881 1.7643975019455
1.65656566619873 1.80149829387665
1.73737370967865 1.8386527299881
1.81818187236786 1.87579309940338
1.89898991584778 1.91151893138885
1.9797979593277 1.94590961933136
2.06060600280762 2.0200731754303
2.14141416549683 2.10689377784729
2.22222232818604 2.193035364151
2.30303025245667 2.27852916717529
2.38383841514587 2.36341714859009
2.46464657783508 2.44774556159973
2.54545450210571 2.53156447410583
2.62626266479492 2.61492657661438
2.70707082748413 2.69788289070129
2.78787875175476 2.78048324584961
2.86868691444397 2.86277484893799
2.9494948387146 2.94480085372925
3.03030300140381 3.02660131454468
3.11111116409302 3.10821175575256
3.19191908836365 3.18966341018677
3.27272725105286 3.27098441123962
3.35353541374207 3.35219764709473
3.4343433380127 3.4333233833313
3.5151515007019 3.51437902450562
3.59595966339111 3.59537863731384
3.67676758766174 3.67633318901062
3.75757575035095 3.75725340843201
3.83838391304016 3.8381462097168
3.91919183731079 3.91901755332947
4 3.99987316131592
};
\addplot [semithick, black, dashed]
table {%
-4 -0.00012671947479248
-3.91919183731079 -0.00017415032198187
-3.83838391304016 -0.000237708140048198
-3.75757575035095 -0.000322291336487979
-3.67676758766174 -0.000434250541729853
-3.59595966339111 -0.000581064610742033
-3.5151515007019 -0.000772497849538922
-3.4343433380127 -0.00102003407664597
-3.35353541374207 -0.00133783894125372
-3.27272725105286 -0.00174294819589704
-3.19191908836365 -0.00225554686039686
-3.11111116409302 -0.00289930240251124
-3.03030300140381 -0.00370162911713123
-2.9494948387146 -0.00469404365867376
-2.86868691444397 -0.0059120487421751
-2.78787875175476 -0.00739550217986107
-2.70707082748413 -0.00918773841112852
-2.62626266479492 -0.011336050927639
-2.54545450210571 -0.0138899739831686
-2.46464657783508 -0.0169011298567057
-2.38383841514587 -0.0204212907701731
-2.30303025245667 -0.0245009362697601
-2.22222232818604 -0.0291870050132275
-2.14141416549683 -0.0345202684402466
-2.06060600280762 -0.0405327528715134
-1.9797979593277 -0.047244131565094
-1.89898991584778 -0.0546584464609623
-1.81818187236786 -0.0627603307366371
-1.73737370967865 -0.0715113654732704
-1.65656566619873 -0.0808464661240578
-1.57575762271881 -0.0906704068183899
-1.4949494600296 -0.100854992866516
-1.41414141654968 -0.11123663187027
-1.33333337306976 -0.121614933013916
-1.25252521038055 -0.131752207875252
-1.17171716690063 -0.141373917460442
-1.09090912342072 -0.150170683860779
-1.01010096073151 -0.157801449298859
-0.929292917251587 -0.163898140192032
-0.848484873771667 -0.168071269989014
-0.767676770687103 -0.169917285442352
-0.686868667602539 -0.169026538729668
-0.60606062412262 -0.164992272853851
-0.525252521038055 -0.157420203089714
-0.444444447755814 -0.145938068628311
-0.363636374473572 -0.130205377936363
-0.282828271389008 -0.109922401607037
-0.202020198106766 -0.084838479757309
-0.121212124824524 -0.0547589734196663
-0.0404040440917015 -0.0195509307086468
0.0404040440917015 0.0208531115204096
0.121212124824524 0.0664531588554382
0.202020198106766 0.117181718349457
0.282828271389008 0.17290586233139
0.363636374473572 0.233430996537209
0.444444447755814 0.298506379127502
0.525252521038055 0.367832332849503
0.60606062412262 0.441068351268768
0.686868667602539 0.51784211397171
0.767676770687103 0.597759485244751
0.848484873771667 0.680413603782654
0.929292917251587 0.765394747257233
1.01010096073151 0.852299511432648
1.09090912342072 0.940738439559937
1.17171716690063 1.03034329414368
1.25252521038055 1.12077295780182
1.33333337306976 1.21171844005585
1.41414141654968 1.30290472507477
1.4949494600296 1.39409446716309
1.57575762271881 1.48508715629578
1.65656566619873 1.57571911811829
1.73737370967865 1.66586244106293
1.81818187236786 1.75542151927948
1.89898991584778 1.84433150291443
1.9797979593277 1.93255388736725
2.06060600280762 2.0200731754303
2.14141416549683 2.10689377784729
2.22222232818604 2.193035364151
2.30303025245667 2.27852916717529
2.38383841514587 2.36341714859009
2.46464657783508 2.44774556159973
2.54545450210571 2.53156447410583
2.62626266479492 2.61492657661438
2.70707082748413 2.69788289070129
2.78787875175476 2.78048324584961
2.86868691444397 2.86277484893799
2.9494948387146 2.94480085372925
3.03030300140381 3.02660131454468
3.11111116409302 3.10821175575256
3.19191908836365 3.18966341018677
3.27272725105286 3.27098441123962
3.35353541374207 3.35219764709473
3.4343433380127 3.4333233833313
3.5151515007019 3.51437902450562
3.59595966339111 3.59537863731384
3.67676758766174 3.67633318901062
3.75757575035095 3.75725340843201
3.83838391304016 3.8381462097168
3.91919183731079 3.91901755332947
4 3.99987316131592
};
\end{axis}

\end{tikzpicture}
    } & 
    \quad
    \scalebox{0.34}{
\begin{tikzpicture}

\definecolor{darkgray176}{RGB}{190,190,190}
\definecolor{dodgerblue}{RGB}{30,144,255}

\begin{axis}[
tick align=outside,
tick pos=left,
x grid style={darkgray176},
xlabel={\(\displaystyle x\)},
xmin=-4, xmax=4,
xtick style={color=black},
y grid style={darkgray176},
ylabel={\(\displaystyle DiTAC(x)\)},
ymin=-2, ymax=4,
ytick style={color=black},
y label style={at={(axis description cs:-0.13,.5)},anchor=south},
x label style={at={(axis description cs:.5,-0.15)}},
grid=both,
grid style={line width=.1pt, draw=gray!10},
major grid style={line width=.2pt,draw=gray!20},
minor x tick num=1,
minor y tick num=1,
]
\addplot [semithick, dodgerblue]
table {%
-4 -0.00012671947479248
-3.91919183731079 -0.00017415032198187
-3.83838391304016 -0.000237708140048198
-3.75757575035095 -0.000322291336487979
-3.67676758766174 -0.000434250541729853
-3.59595966339111 -0.000581064610742033
-3.5151515007019 -0.000772497849538922
-3.4343433380127 -0.00102003407664597
-3.35353541374207 -0.00133783894125372
-3.27272725105286 -0.00174294819589704
-3.19191908836365 -0.00225554686039686
-3.11111116409302 -0.00289930240251124
-3.03030300140381 -0.00370162911713123
-2.9494948387146 -0.00469404365867376
-2.86868691444397 -0.0059120487421751
-2.78787875175476 -0.00739550217986107
-2.70707082748413 -0.00918773841112852
-2.62626266479492 -0.011336050927639
-2.54545450210571 -0.0138899739831686
-2.46464657783508 -0.0169011298567057
-2.38383841514587 -0.0204212907701731
-2.30303025245667 -0.0245009362697601
-2.22222232818604 -0.0291870050132275
-2.14141416549683 -0.0345202684402466
-2.06060600280762 -0.0405327528715134
-1.9797979593277 -0.047244131565094
-1.89898991584778 -0.0546584464609623
-1.81818187236786 -0.0627603307366371
-1.73737370967865 -0.0715113654732704
-1.65656566619873 -0.0808464661240578
-1.57575762271881 -0.0906704068183899
-1.4949494600296 -0.100854992866516
-1.41414141654968 -0.11123663187027
-1.33333337306976 -0.121614933013916
-1.25252521038055 -0.131752207875252
-1.17171716690063 -0.141373917460442
-1.09090912342072 -0.150170683860779
-1.01010096073151 -0.157801449298859
-0.929292917251587 -0.163898140192032
-0.848484873771667 -0.168071269989014
-0.767676770687103 -0.169917285442352
-0.686868667602539 -0.169026538729668
-0.60606062412262 -0.164992272853851
-0.525252521038055 -0.157420203089714
-0.444444447755814 -0.156385242938995
-0.363636374473572 -0.157876238226891
-0.282828271389008 -0.157831504940987
-0.202020198106766 -0.154180511832237
-0.121212124824524 -0.132459253072739
-0.0404040440917015 -0.0724878832697868
0.0404040440917015 0.0462098158895969
0.121212124824524 0.301163136959076
0.202020198106766 0.517434775829315
0.282828271389008 0.67666620016098
0.363636374473572 0.805009067058563
0.444444447755814 0.914732694625854
0.525252521038055 1.01143300533295
0.60606062412262 1.10082602500916
0.686868667602539 1.18120896816254
0.767676770687103 1.25385451316833
0.848484873771667 1.32152998447418
0.929292917251587 1.38503205776215
1.01010096073151 1.44422662258148
1.09090912342072 1.4997376203537
1.17171716690063 1.55112314224243
1.25252521038055 1.59924817085266
1.33333337306976 1.64440560340881
1.41414141654968 1.6868052482605
1.4949494600296 1.7264267206192
1.57575762271881 1.7643975019455
1.65656566619873 1.80149829387665
1.73737370967865 1.8386527299881
1.81818187236786 1.87579309940338
1.89898991584778 1.91151893138885
1.9797979593277 1.94590961933136
2.06060600280762 2.0200731754303
2.14141416549683 2.10689377784729
2.22222232818604 2.193035364151
2.30303025245667 2.27852916717529
2.38383841514587 2.36341714859009
2.46464657783508 2.44774556159973
2.54545450210571 2.53156447410583
2.62626266479492 2.61492657661438
2.70707082748413 2.69788289070129
2.78787875175476 2.78048324584961
2.86868691444397 2.86277484893799
2.9494948387146 2.94480085372925
3.03030300140381 3.02660131454468
3.11111116409302 3.10821175575256
3.19191908836365 3.18966341018677
3.27272725105286 3.27098441123962
3.35353541374207 3.35219764709473
3.4343433380127 3.4333233833313
3.5151515007019 3.51437902450562
3.59595966339111 3.59537863731384
3.67676758766174 3.67633318901062
3.75757575035095 3.75725340843201
3.83838391304016 3.8381462097168
3.91919183731079 3.91901755332947
4 3.99987316131592
};
\addplot [semithick, black, dashed]
table {%
-4 0
-3.91919183731079 0
-3.83838391304016 0
-3.75757575035095 0
-3.67676758766174 0
-3.59595966339111 0
-3.5151515007019 0
-3.4343433380127 0
-3.35353541374207 0
-3.27272725105286 0
-3.19191908836365 0
-3.11111116409302 0
-3.03030300140381 0
-2.9494948387146 0
-2.86868691444397 0
-2.78787875175476 0
-2.70707082748413 0
-2.62626266479492 0
-2.54545450210571 0
-2.46464657783508 0
-2.38383841514587 0
-2.30303025245667 0
-2.22222232818604 0
-2.14141416549683 0
-2.06060600280762 0
-1.9797979593277 0
-1.89898991584778 0
-1.81818187236786 0
-1.73737370967865 0
-1.65656566619873 0
-1.57575762271881 0
-1.4949494600296 0
-1.41414141654968 0
-1.33333337306976 0
-1.25252521038055 0
-1.17171716690063 0
-1.09090912342072 0
-1.01010096073151 0
-0.929292917251587 0
-0.848484873771667 0
-0.767676770687103 0
-0.686868667602539 0
-0.60606062412262 0
-0.525252521038055 0
-0.444444447755814 0
-0.363636374473572 0
-0.282828271389008 0
-0.202020198106766 0
-0.121212124824524 0
-0.0404040440917015 0
0.0404040440917015 0.0404040440917015
0.121212124824524 0.121212124824524
0.202020198106766 0.202020198106766
0.282828271389008 0.282828271389008
0.363636374473572 0.363636374473572
0.444444447755814 0.444444447755814
0.525252521038055 0.525252521038055
0.60606062412262 0.60606062412262
0.686868667602539 0.686868667602539
0.767676770687103 0.767676770687103
0.848484873771667 0.848484873771667
0.929292917251587 0.929292917251587
1.01010096073151 1.01010096073151
1.09090912342072 1.09090912342072
1.17171716690063 1.17171716690063
1.25252521038055 1.25252521038055
1.33333337306976 1.33333337306976
1.41414141654968 1.41414141654968
1.4949494600296 1.4949494600296
1.57575762271881 1.57575762271881
1.65656566619873 1.65656566619873
1.73737370967865 1.73737370967865
1.81818187236786 1.81818187236786
1.89898991584778 1.89898991584778
1.9797979593277 1.9797979593277
2.06060600280762 2.06060600280762
2.14141416549683 2.14141416549683
2.22222232818604 2.22222232818604
2.30303025245667 2.30303025245667
2.38383841514587 2.38383841514587
2.46464657783508 2.46464657783508
2.54545450210571 2.54545450210571
2.62626266479492 2.62626266479492
2.70707082748413 2.70707082748413
2.78787875175476 2.78787875175476
2.86868691444397 2.86868691444397
2.9494948387146 2.9494948387146
3.03030300140381 3.03030300140381
3.11111116409302 3.11111116409302
3.19191908836365 3.19191908836365
3.27272725105286 3.27272725105286
3.35353541374207 3.35353541374207
3.4343433380127 3.4343433380127
3.5151515007019 3.5151515007019
3.59595966339111 3.59595966339111
3.67676758766174 3.67676758766174
3.75757575035095 3.75757575035095
3.83838391304016 3.83838391304016
3.91919183731079 3.91919183731079
4 4
};
\end{axis}

\end{tikzpicture}
    }\\ 
    \textbf{\ \ \small(a)} & \textbf{\ \ \ \ \ \small(b)}
    \end{tabular}
    \caption{Here we display (a) the learned DiTAC (solid line) vs. GELU (dashed line), and (b) the learned DiTAC (solid line) vs. ReLU (dashed line), on the semantic segmentation task, trained on ResNet-50+PSPNet. We associate DiTAC's improved performance to its unique function.}
    \label{fig:learned_functions_pspnet}
\end{figure}

To reveal the reason for this gap, in \autoref{fig:learned_functions_pspnet} we display GELU and ReLU (baseline) functions vs. the learned DiTAC in the last layer of the ResNet-50+PSPNet configuration, at time of convergence. It can be seen that DiTAC has learned a unique function shape which led to its improved performance.


\subsubsection{Generative Networks.}
We further evaluate DiTAC's performance on the image generation task by training two types of GAN architectures, DCGAN~\cite{Radford:2015:DCGAN} as an unconditional GAN, and BigGAN~\cite{Brock:2018:BigGAN} as a conditional GAN, on the CelebA~\cite{Liu:2015:CelebA} and CIFAR-10~\cite{Krizhevsky:2009:CIFAR} datasets, respectively.
Similar to the semantic-segmentation experiments, we compare the performance of the baseline AFs with DiTAC's and GELU's on both the Discriminator (D) and Generator (G) modules. DCGAN uses ReLU in G and LReLU in D as the baseline AFs, whereas BigGAN uses ReLU as the baseline AF in both D and G. 
We utilize the implementation of \texttt{mmgen}~\cite{MMGeneration:2021:mmgeneration} for training and evaluation, and follow their suggested training procedure. Additional training details are provided in the Appendix. 

To approximate measures of sample quality, we employ the popular standard metrics Fr\'echet Inception Distance (FID)~\cite{Heusel:2017:FID_metric} and Inception Score (IS)~\cite{Salimans:2016:IS_metric}.
In \autoref{tab:gans} we report the FID and IS results
for the above experiments. The results demonstrate a clear advantage to DiTAC on the two architecture types.

\begin{table}[hbt!]  
    \centering
    \setlength{\tabcolsep}{5pt}
    \caption{Image generation performance of DiTAC vs. baseline AFs and GELU, on DCGAN and BigGAN.}
    \begin{tabular}{lccc}
         \toprule
         Model & Dataset & FID $\downarrow$ & IS $\uparrow$ \\
         \midrule
         DCGAN & CelebA & 133.5 & 2.4  \\
         DCGAN+GELU & CelebA & 516.5 & 1.0 \\
         DCGAN+DiTAC & CelebA & \textbf{34.6} & \textbf{2.6} \\
         \midrule
         BigGAN & CIFAR-10 & 10.3 & 9.5 \\
         BigGAN+GELU & CIFAR-10 & 27.6 & 7.6 \\
         BigGAN+DiTAC & CIFAR-10 & \textbf{10.0} & \textbf{9.8} \\
         \bottomrule
    \end{tabular}
    \label{tab:gans}
\end{table}

\section{Conclusion}\label{sec:conclusion}

In this paper we presented DiTAC, a novel highly-expressive TAF that is based on efficient diffeomorphisms. 
We showed that DiTAC outperforms existing AFs and TAFs in various challenging tasks, while adding
only a negligible amount of trainable parameters.
The main limitation of DiTAC is its hypothetical computational cost during training. 
However, this limitation is mostly theoretical as we also provided a technical solution that alleviates this cost in training, and eliminates it completely in inference.

\section*{Acknowledgements} This work was supported in part by the Lynn and William Frankel Center at BGU CS, by Israel Science Foundation Personal Grant \#360/21, and by the Israeli Council for Higher Education (CHE) via the Data Science Research Center at BGU. 
I.C. and S.E.F. were also funded in part by the Kreitman School of Advanced Graduate Studies and by BGU’s Hi-Tech Scholarship.
I.C. was also supported by the Israel’s Ministry of Technology and Science Aloni Scholarship.

%
%
\bibliographystyle{splncs04}
\bibliography{refs}

\clearpage
\appendix


\section{Toy-data Experiments} \label{submat:toy_experiments}


\subsection{Classification}

Here we provide additional details about the toy-data classification task presented in section 4.1 in the paper. We train a simple MLP on MNIST dataset and a synthetic dataset generated from a two-dimensional Gaussian-Mixture-Model (2D-GMM).

\subsubsection{Data generation of 2D-GMM.}
\begin{figure}[hbt!]
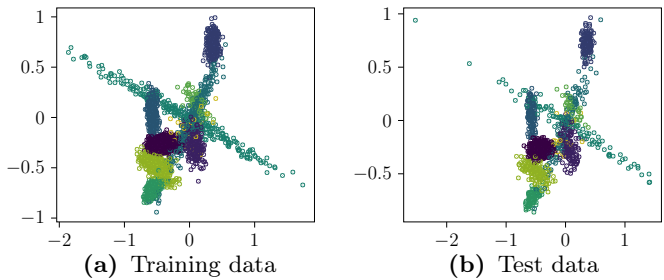

    \renewcommand{\arraystretch}{0.04}
    \centering
    \Large
    \begin{tabular}{ccc}
    \hspace{-0.6cm}
    \scalebox{0.5}{
        \input{./raw/2d_gmm_data_scatter/2d_data_train}
    } & 
    \hspace{-0.07cm}
    \scalebox{0.5}{
        \input{./raw/2d_gmm_data_scatter/2d_data_test}
    }\\ 
    \hspace{-0.09cm} \textbf{\small(a)} \small{Training data} & \hspace{0.36cm} \textbf{\small(b)} \small{Test data}
    \end{tabular}
    \caption{Here we display a 2D-GMM dataset sample separated into train and test sets.}    
    \label{fig:2d_gmm_scatter}
\end{figure}

We generate a synthetic data sampled from a 2D GMM with 10 groups (referred to as classes). The parameters of each Gaussian component (the mean vector and covariance matrix) are drawn from a normal-inverse-Wishart distribution
while the mixture weights are drawn drawn from a Dirichlet distribution. We then draw 
 $5\cdot10^3$ \iid 2D points from the GMM, with a split of $70\%/30\%$ for train/test data.  An example for such sample is shown in \autoref{fig:2d_gmm_scatter}.

\subsubsection{Training procedure.}

We use a simple MLP with two hidden layers to train on either MNIST or the 2D-GMM datasets. For MNIST we set their sizes to 128 and 64, and for the 2D-GMM we set both to 100.
We train both datasets for 150 epochs, and use Adam optimizer, a batch size of 64, and a learning rate of $1\cdot10^4$ with no weight decay.  
As for the competitors, we use the same training parameters except for the learning rate: we run each configuration with several learning rates and select the best performance for each competitor.

\subsubsection{Top-1 accuracy trend.}

\begin{figure}[hbt!] 
    \centering
    \framebox{\includegraphics[width=9cm,height=4cm]{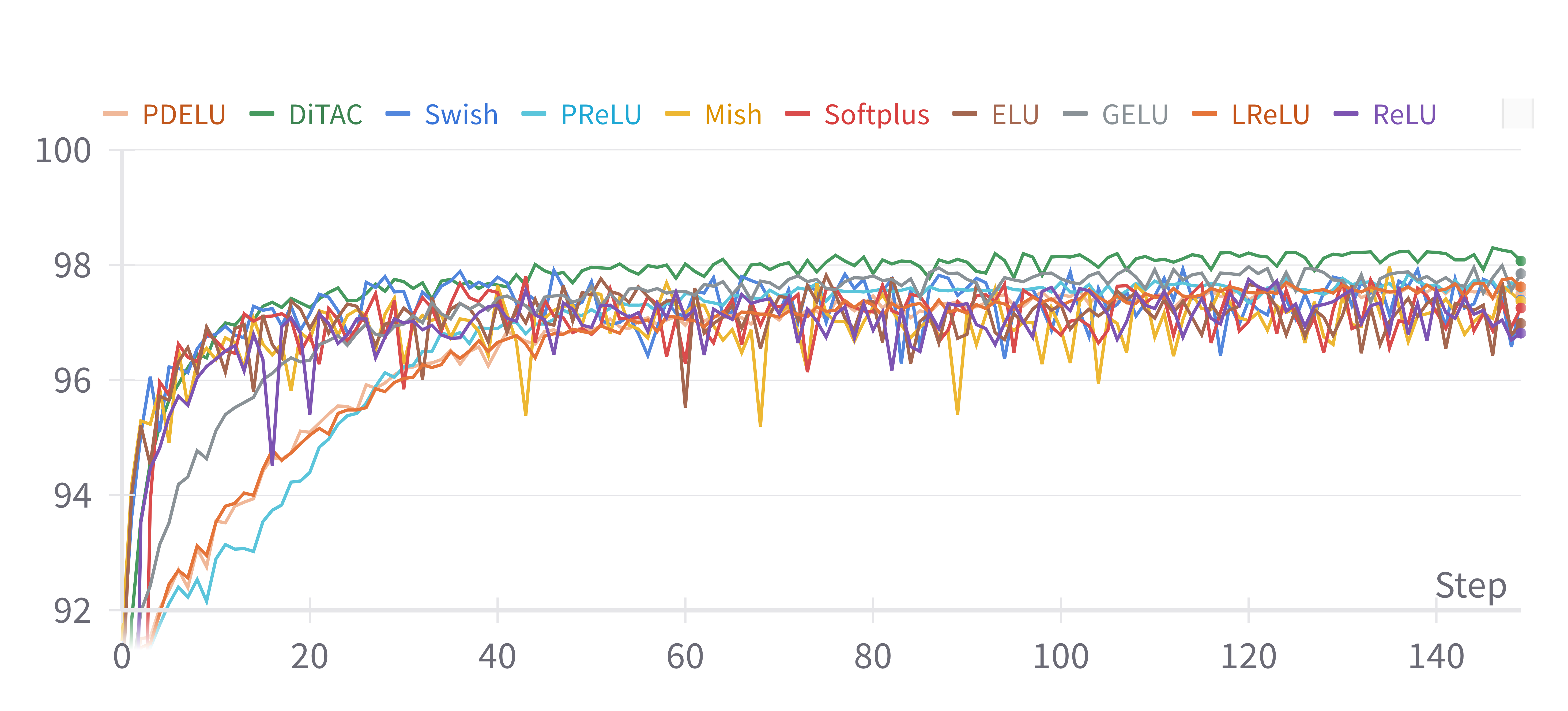}}
    \caption{Top-1 accuracy of MNIST classification on a simple MLP, using various AFs.}
    \label{fig:mnist_acc_all_acts}
\end{figure}

In \autoref{fig:mnist_acc_all_acts} we show top-1 accuracy results of DiTAC vs. existing competitors, trained on MNIST dataset. The results demonstrate a clear advantage to DiTAC.


\subsection{Regression}

Here we provide additional details about the toy-data regression task presented in section 4.1 in the paper, and display two more experiments that are not shown in the main text.


\subsubsection{Auto-MPG dataset.}
We run a standard linear regression problem on a simple MLP with two hidden layers of size 100, on the Auto-MPG dataset~\cite{Quinlan:1993:auto_mpg}, using the MPG and Horsepower values. Similar to other experiments, we compared DiTAC to existing AFs and TAFs. In \autoref{tab:regression_mpg} we report the Mean Squared Error (MSE) and the R-squared (R2) score, and show that DiTAC provides the best performance. Training details are shared later in this section. This experiment is not presented in the main text.

\begin{table}[hbt!]  
    \centering
    \setlength{\tabcolsep}{5pt}
    \caption{Regression results on Auto-MPG dataset (horsepower) using a simple MLP. We were unable to report accuracy in several cases due to an unstable training process.}
    \begin{tabular}{lcccccc}
         \toprule
         Act. & MSE $\downarrow$ & R2 $\uparrow$ \\
         \midrule
         ReLU & 409.0  & 71.8 \\
         LReLU & 404.3  & 72.2 \\
         GELU & 412.2  & 71.6 \\
         ELU & 404.3  & 72.2 \\
         Softplus & 410.0  & 71.8  \\
         Mish & 410.9  & 71.7 \\
         Swish & 405.8  & 72.1 \\
         PReLU & 406.3  & 72.0 \\
         PDELU & --  & -- \\
         \midrule
         DiTAC & \textbf{389.6} & \textbf{73.2}  \\
         \bottomrule
    \end{tabular}
    \label{tab:regression_mpg}
\end{table}

\subsubsection{Reconstruction of 1D/2D functions.}\label{Sec:Regression:Subsec:func}

We experiment with regression tasks of reconstructing one- and two-dimensional functions. Here we elaborate more details about the experiments mentioned in the paper (Table 2 in section 4.1), and show an experiment on a new 1D function. Training details are shared later in this section.\\

In Table 2 in the paper, we display performance evaluation of DiTAC vs. other AFs and TAFs in reconstructing 1D and 2D target functions, by training an MLP with one hidden layer of sizes 30 and 50 respectively.
The 1D target function is $\sin(\exp 6x)$, and the 2D target function is a sum of sines with various frequencies:
\begin{align}	  
& 0.4\sin(9xy)+0.1\sin(-9x+11y)+0.15\sin(3x+13y)\\
& +0.15\sin(9x+9y)+0.1\sin(13x+5y)+0.1\sin(3x+19y)
\end{align}

\begin{figure}[hbt!] 
    \centering
   \framebox{\includegraphics[width=9cm,height=4cm]{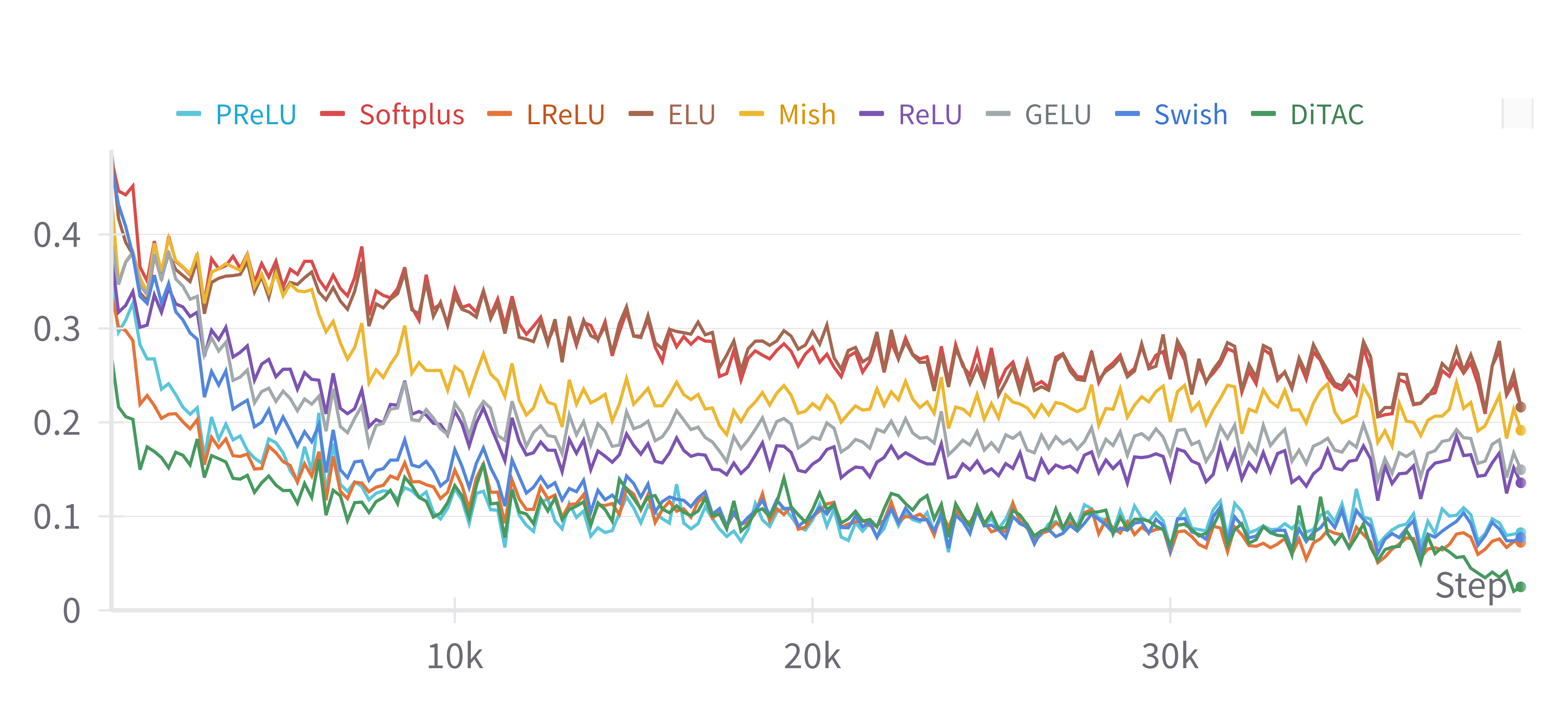}}
    \caption{MSE of the 1D-function reconstruction task, $\sin(\exp 6x)$, on a simple MLP, using various AFs.}
    \label{fig:1d_func_mse}
\end{figure}

\begin{figure}[t] 
    \centering
    \setlength{\fboxsep}{0pt}
    \setlength{\fboxrule}{0.1pt}
    \captionsetup[sub]{font=small}
    \hspace{2pt}
     \begin{subfigure}[b]{0.42\textwidth}
         \centering
         \framebox{\includegraphics[width=\textwidth]{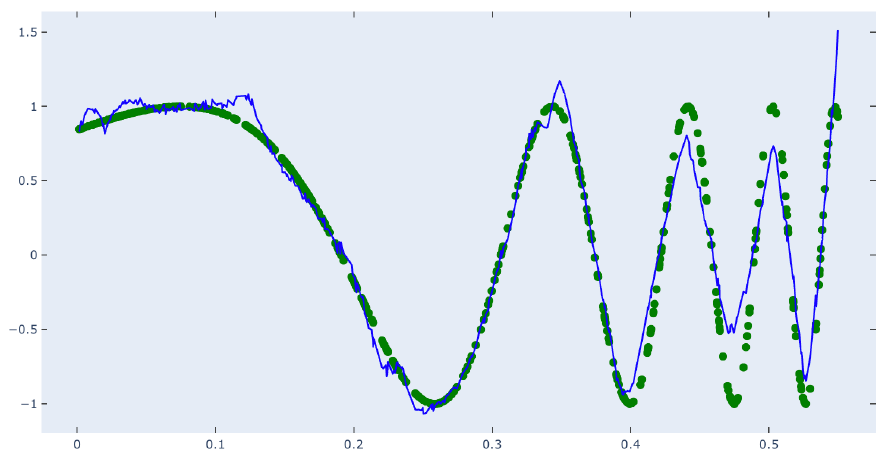}}
         \caption{DiTAC}
     \end{subfigure}
     \hspace{10pt}
     \begin{subfigure}[b]{0.42\textwidth}
         \centering
         \framebox{\includegraphics[width=\textwidth]{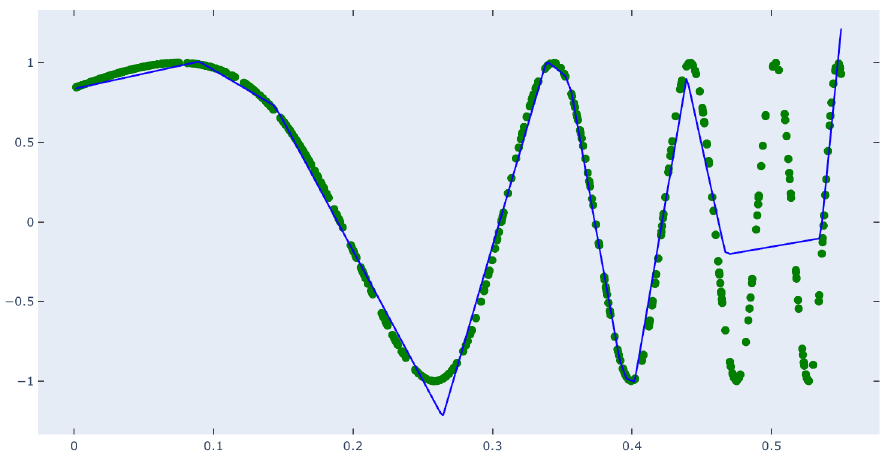}}
         \caption{LReLU}
     \end{subfigure}
    \hspace{2pt}
        \caption{1D-function reconstruction, $\sin(\exp 6x)$, learned by DiTAC and (the runner-up) LReLU. DiTAC manages to learn a smooth function that fits the data.}
        \label{fig:1d_learned_func}
\end{figure}

In \autoref{fig:1d_func_mse} we present the MSE trend of DiTAC vs. existing AFs and TAFs, on the 1D-function reconstruction experiment. 
As shown in Table 2 in the paper, DiTAC noticeably provides the best performance, gaining R2 of 96.26\% vs. 89.78\% by LReLU (second best) in the 1D-function experiment, and R2 of 98.37\% vs. 96.09\% by PReLU (second best) in the 2D-function experiment. 
This performance gap is also demonstrated in \autoref{fig:1d_learned_func}, where we compare the reconstructed 1D-function learned by DiTAC and its runner-up LReLU.\\

We show another 1D-function reconstruction experiment, where the target function is as follows:
\begin{align}	
0.4\sin(19x)+0.2\sin(23x)+0.3\sin(29x)+0.1\sin(31x)
\end{align}

\begin{figure}[hbt!] 
    \centering
   \framebox{\includegraphics[width=9cm,height=4cm]{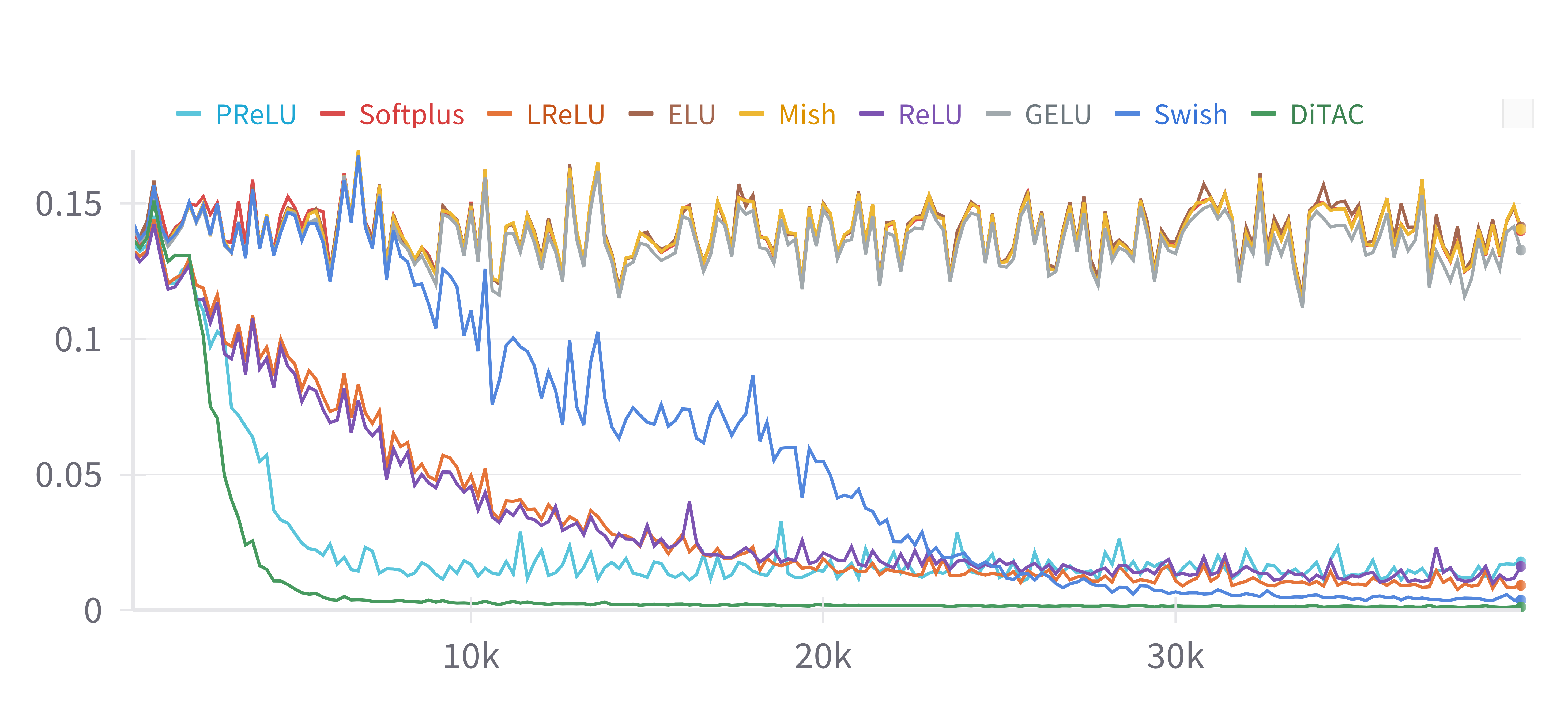}}
    \caption{MSE of the 1D-function reconstruction task on a simple MLP, using various AFs.}
    \label{fig:1d_second_func_mse}
\end{figure}

Here we used an MLP with one hidden layer of size 64. Results and model size (number of parameters) are shown in \autoref{tab:regression_1d_function} and \autoref{fig:1d_second_func_mse}. It can be shown that only part of the competitors manage to fit a reasonable function, while the others perform badly, although each AF was tested on several learning rate options.
\begin{table}[hbt!] 
    \centering
    \setlength{\tabcolsep}{5pt}
    \caption{Regression-task results of learning one-dimensional target function on a simple MLP, using various AFs and TAFs, along with the number of model parameters used by each activation.}
    \begin{tabular}{lccc}
         \toprule
         AF/TAF & Param. & MSE $\downarrow$ & R2 $\uparrow$ \\
         \midrule
         ReLU & 193 & 0.010 & 93.3  \\
         LReLU & 193 & 0.008  & 94.7 \\
         GELU & 193 & 0.111  & 9.7 \\
         ELU & 193 & 0.116 & 3.6 \\
         Softplus & 193 & 0.114  & 5.2  \\
         Mish & 193 & 0.113 & 5.5 \\
         Swish & 194 & 0.004 & 97.5 \\
         PReLU & 194 & 0.010 & 93.1 \\
         \midrule
         DiTAC & 202 & \textbf{0.001} & \textbf{99.3} \\
         \bottomrule
    \end{tabular}
    \label{tab:regression_1d_function}
\end{table}

\subsubsection{Training procedure.}
In all of the aforementioned regression experiments, we train a model for 40K iterations, using Adam optimizer, a batch size of 98, no weight decay, and a learning rate of 0.01 except of the new 1D-function experiment, in which we use a learning rate of $1\cdot10^3$.
Similar to the classification experiments, we evaluate the competitors using the same training parameters except for the learning rate, as each competitor's best performance was selected out of several learning rates options.

\section{Real-world Data Experiments} \label{submat:real_world_experiments}

Here we provide technical details about the training procedures of the semantic segmentation and the image generation tasks. 

\subsubsection{Semantic Segmentation.} 

In the paper we compare the performance of DiTAC, ReLU and GELU activation functions on the semantic segmentation task.
In the two first experiments we train UperNet~\cite{Xiao:2018:UperNet} as a segmentation framework on the ADE20K dataset~\cite{Zhou:IJCV:2019:ade20k}, where in one configuration we use ConvNeXT-T and in the other we use Swin-T as the backbone.
We follow \texttt{mmseg}'s~\cite{MMSegmentation:2020:mmsegmentation} configuration to setup the training parameters.
We use 160K iterations schedule, crop size of $512\times 512$, batch size of 16, weight decay of 0.01, and the polynomial learning rate decay policy~\cite{Chen:2017:poly_lr} where the initial learning rate is $6\cdot 10^{-5}$ and the power is 1.
In the third experiment we train a PSPNet~\cite{Zhao:2017:PSPNet} segmentation framework using ResNet-50~\cite{He:2016:resnet} as the backbone, on the Cityscapes dataset~\cite{Cordts:2016:cityscapes}.
We use a 40K-iteration schedule, crop size of $512\times 1024$, batch size of 8, weight decay of $5\cdot 10^{-4}$, and the polynomial learning rate decay policy where the initial learning rate is 0.01 and the power is 0.9.
In all configurations the backbones are pre-trained on ImageNet-1K dataset using 300 epoch schedule.

\subsubsection{Generative Networks.}
In the paper we evaluate DiTAC on the image generation task, on DCGAN~\cite{Radford:2015:DCGAN}, an unconditional GAN, and on BigGAN~\cite{Brock:2018:BigGAN}, a conditional GAN. 
We train DCGAN for 300K iterations on 4 GPUs, using image crop size of $64\times 64$, an effective batch size of 128, Adam optimizer, and a learning rate of $2\cdot 10^{-4}$ in both D and G. 
We train BigGAN for 500K iterations, using an effective batch size of 64, Adam optimizer, and a learning rate of $2\cdot 10^{-4}$ in D and $5\cdot 10^{-5}$ in G.


\section{DiTAC Versions} \label{submat:ditac_versions}

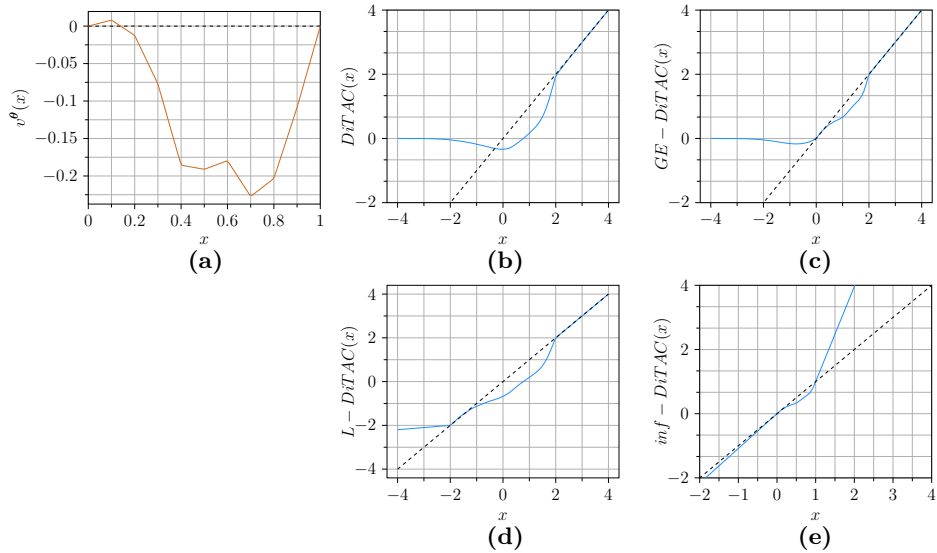
\begin{figure}[t]
    \renewcommand{\arraystretch}{0.06}
    \centering
    \Large
    \begin{tabular}{ccc}
    \hspace{-0.5cm}
    \scalebox{0.45}{
\begin{tikzpicture}

\definecolor{chocolate}{RGB}{210,105,30}
\definecolor{darkgray176}{RGB}{176,176,176}

\begin{axis}[
tick align=outside,
tick pos=left,
x grid style={darkgray176},
xlabel={\(\displaystyle x\)},
xmajorgrids,
xmin=0, xmax=1,
xtick style={color=black},
grid=both,
grid style={line width=.1pt, draw=gray!10},
major grid style={line width=.2pt,draw=gray!20},
minor x tick num=1,
minor y tick num=1,
y grid style={darkgray176},
ylabel={\(\displaystyle v^{\btheta}(x)\)},
ymajorgrids,
ymin=-0.237489585531875, ymax=0.0194867288228124,
ytick style={color=black},
y label style={at={(axis description cs:-0.23,.5)},anchor=south},
tick label style={/pgf/number format/fixed},
]
\addplot [semithick, black, dashed]
table {%
0 0
0.0101010101010101 0
0.0202020202020202 0
0.0303030303030303 0
0.0404040404040404 0
0.0505050505050505 0
0.0606060606060606 0
0.0707070707070707 0
0.0808080808080808 0
0.0909090909090909 0
0.101010101010101 0
0.111111111111111 0
0.121212121212121 0
0.131313131313131 0
0.141414141414141 0
0.151515151515152 0
0.161616161616162 0
0.171717171717172 0
0.181818181818182 0
0.191919191919192 0
0.202020202020202 0
0.212121212121212 0
0.222222222222222 0
0.232323232323232 0
0.242424242424242 0
0.252525252525253 0
0.262626262626263 0
0.272727272727273 0
0.282828282828283 0
0.292929292929293 0
0.303030303030303 0
0.313131313131313 0
0.323232323232323 0
0.333333333333333 0
0.343434343434343 0
0.353535353535354 0
0.363636363636364 0
0.373737373737374 0
0.383838383838384 0
0.393939393939394 0
0.404040404040404 0
0.414141414141414 0
0.424242424242424 0
0.434343434343434 0
0.444444444444444 0
0.454545454545455 0
0.464646464646465 0
0.474747474747475 0
0.484848484848485 0
0.494949494949495 0
0.505050505050505 0
0.515151515151515 0
0.525252525252525 0
0.535353535353535 0
0.545454545454546 0
0.555555555555556 0
0.565656565656566 0
0.575757575757576 0
0.585858585858586 0
0.595959595959596 0
0.606060606060606 0
0.616161616161616 0
0.626262626262626 0
0.636363636363636 0
0.646464646464647 0
0.656565656565657 0
0.666666666666667 0
0.676767676767677 0
0.686868686868687 0
0.696969696969697 0
0.707070707070707 0
0.717171717171717 0
0.727272727272727 0
0.737373737373737 0
0.747474747474748 0
0.757575757575758 0
0.767676767676768 0
0.777777777777778 0
0.787878787878788 0
0.797979797979798 0
0.808080808080808 0
0.818181818181818 0
0.828282828282828 0
0.838383838383838 0
0.848484848484849 0
0.858585858585859 0
0.868686868686869 0
0.878787878787879 0
0.888888888888889 0
0.898989898989899 0
0.909090909090909 0
0.919191919191919 0
0.929292929292929 0
0.939393939393939 0
0.94949494949495 0
0.95959595959596 0
0.96969696969697 0
0.97979797979798 0
0.98989898989899 0
1 0
};
\addplot [semithick, chocolate, opacity=1.0]
table {%
0 1.06626889680535e-19
0.0101010101010101 0.000809283868875355
0.0202020202020202 0.00161856773775071
0.0303030303030303 0.00242785178124905
0.0404040404040404 0.00323713547550142
0.0505050505050505 0.00404641916975379
0.0606060606060606 0.00485570356249809
0.0707070707070707 0.00566498702391982
0.0808080808080808 0.00647427095100284
0.0909090909090909 0.00728355487808585
0.101010101010101 0.00780598726123571
0.111111111111111 0.00574668310582638
0.121212121212121 0.00368737848475575
0.131313131313131 0.00162807549349964
0.141414141414141 -0.00043122743954882
0.151515151515152 -0.00249053351581097
0.161616161616162 -0.00454983627423644
0.171717171717172 -0.0066091394983232
0.181818181818182 -0.0086684450507164
0.191919191919192 -0.0107277482748032
0.202020202020202 -0.0136815570294857
0.212121212121212 -0.0202134288847446
0.222222222222222 -0.0267452895641327
0.232323232323232 -0.0332771502435207
0.242424242424242 -0.0398090220987797
0.252525252525253 -0.0463408753275871
0.262626262626263 -0.0528727471828461
0.272727272727273 -0.059404619038105
0.282828282828283 -0.0659364685416222
0.292929292929293 -0.0724683403968811
0.303030303030303 -0.080329641699791
0.313131313131313 -0.0912928655743599
0.323232323232323 -0.102256119251251
0.333333333333333 -0.113219380378723
0.343434343434343 -0.124182604253292
0.353535353535354 -0.135145872831345
0.363636363636364 -0.146109119057655
0.373737373737374 -0.157072350382805
0.383838383838384 -0.168035611510277
0.393939393939394 -0.178998872637749
0.404040404040404 -0.185800477862358
0.414141414141414 -0.186359703540802
0.424242424242424 -0.186918914318085
0.434343434343434 -0.187478125095367
0.444444444444444 -0.188037350773811
0.454545454545455 -0.188596561551094
0.464646464646465 -0.189155787229538
0.474747474747475 -0.189714998006821
0.484848484848485 -0.190274208784103
0.494949494949495 -0.190833434462547
0.505050505050505 -0.190527930855751
0.515151515151515 -0.18935763835907
0.525252525252525 -0.188187345862389
0.535353535353535 -0.187017053365707
0.545454545454546 -0.185846775770187
0.555555555555556 -0.184676483273506
0.565656565656566 -0.183506190776825
0.575757575757576 -0.182335913181305
0.585858585858586 -0.181165620684624
0.595959595959596 -0.179995328187943
0.606060606060606 -0.182419791817665
0.616161616161616 -0.187240794301033
0.626262626262626 -0.19206178188324
0.636363636363636 -0.196882799267769
0.646464646464647 -0.201703801751137
0.656565656565657 -0.206524819135666
0.666666666666667 -0.211345836520195
0.676767676767677 -0.216166839003563
0.686868686868687 -0.22098782658577
0.696969696969697 -0.225808843970299
0.707070707070707 -0.225560307502747
0.717171717171717 -0.223139062523842
0.727272727272727 -0.220717802643776
0.737373737373737 -0.218296572566032
0.747474747474748 -0.215875327587128
0.757575757575758 -0.213454082608223
0.767676767676768 -0.211032837629318
0.777777777777778 -0.208611577749252
0.787878787878788 -0.206190332770348
0.797979797979798 -0.203769102692604
0.808080808080808 -0.195632606744766
0.818181818181818 -0.18606735765934
0.828282828282828 -0.176502093672752
0.838383838383838 -0.166936844587326
0.848484848484849 -0.1573715955019
0.858585858585859 -0.147806391119957
0.868686868686869 -0.138241142034531
0.878787878787879 -0.128675878047943
0.888888888888889 -0.119110628962517
0.898989898989899 -0.10954537987709
0.909090909090909 -0.0987171828746796
0.919191919191919 -0.0877486541867256
0.929292929292929 -0.0767800658941269
0.939393939393939 -0.0658114776015282
0.94949494949495 -0.0548428855836391
0.95959595959596 -0.0438742972910404
0.96969696969697 -0.0329057052731514
0.97979797979798 -0.0219371803104877
0.98989898989899 -0.0109685901552439
1 0
};
\end{axis}

\end{tikzpicture}
    } &
    \hspace{-0.4cm}
    \scalebox{0.45}{
\begin{tikzpicture}

\definecolor{darkgray176}{RGB}{176,176,176}
\definecolor{dodgerblue}{RGB}{30,144,255}

\begin{axis}[
tick align=outside,
tick pos=left,
x grid style={darkgray176},
xlabel={\(\displaystyle x\)},
xmajorgrids,
xmin=-4.4, xmax=4.4,
xtick style={color=black},
grid=both,
grid style={line width=.1pt, draw=gray!10},
major grid style={line width=.2pt,draw=gray!20},
minor x tick num=1,
minor y tick num=2,
y grid style={darkgray176},
ylabel={\(\displaystyle DiTAC(x)\)},
y label style={at={(axis description cs:-0.1,.5)},anchor=south},
ymajorgrids,
ymin=-2, ymax=4,
ytick style={color=black}
]
\addplot [semithick, dodgerblue]
table {%
-4 -0.00012671947479248
-3.91919183731079 -0.00017415032198187
-3.83838391304016 -0.000237708140048198
-3.75757575035095 -0.000322291336487979
-3.67676758766174 -0.000434250541729853
-3.59595966339111 -0.000581064610742033
-3.5151515007019 -0.000772497849538922
-3.4343433380127 -0.00102003407664597
-3.35353541374207 -0.00133783894125372
-3.27272725105286 -0.00174294819589704
-3.19191908836365 -0.00225554686039686
-3.11111116409302 -0.00289930240251124
-3.03030300140381 -0.00370162911713123
-2.9494948387146 -0.00469404365867376
-2.86868691444397 -0.0059120487421751
-2.78787875175476 -0.00739550217986107
-2.70707082748413 -0.00918773841112852
-2.62626266479492 -0.011336050927639
-2.54545450210571 -0.0138899739831686
-2.46464657783508 -0.0169011298567057
-2.38383841514587 -0.0204212907701731
-2.30303025245667 -0.0245009362697601
-2.22222232818604 -0.0291870050132275
-2.14141416549683 -0.0345202684402466
-2.06060600280762 -0.0405327528715134
-1.9797979593277 -0.0472210124135017
-1.89898991584778 -0.0543971508741379
-1.81818187236786 -0.0622401274740696
-1.73737370967865 -0.070644423365593
-1.65656566619873 -0.079422764480114
-1.57575762271881 -0.0892691761255264
-1.4949494600296 -0.100255593657494
-1.41414141654968 -0.111629031598568
-1.33333337306976 -0.123480707406998
-1.25252521038055 -0.13536325097084
-1.17171716690063 -0.147680073976517
-1.09090912342072 -0.162371888756752
-1.01010096073151 -0.177556663751602
-0.929292917251587 -0.193051502108574
-0.848484873771667 -0.20850832760334
-0.767676770687103 -0.223558127880096
-0.686868667602539 -0.239124268293381
-0.60606062412262 -0.254935622215271
-0.525252521038055 -0.27054825425148
-0.444444447755814 -0.286224842071533
-0.363636374473572 -0.301438629627228
-0.282828271389008 -0.314743041992188
-0.202020198106766 -0.325925588607788
-0.121212124824524 -0.333650529384613
-0.0404040440917015 -0.33659303188324
0.0404040440917015 -0.334809362888336
0.121212124824524 -0.325771450996399
0.202020198106766 -0.309539169073105
0.282828271389008 -0.284073263406754
0.363636374473572 -0.246581301093102
0.444444447755814 -0.202653750777245
0.525252521038055 -0.155815541744232
0.60606062412262 -0.105402044951916
0.686868667602539 -0.0543759763240814
0.767676770687103 -0.00174257659818977
0.848484873771667 0.0536900646984577
0.929292917251587 0.111103624105453
1.01010096073151 0.171802312135696
1.09090912342072 0.236998587846756
1.17171716690063 0.303542703390121
1.25252521038055 0.373682290315628
1.33333337306976 0.456345736980438
1.41414141654968 0.55264675617218
1.4949494600296 0.672233402729034
1.57575762271881 0.816165745258331
1.65656566619873 0.984010219573975
1.73737370967865 1.19043016433716
1.81818187236786 1.41387677192688
1.89898991584778 1.65011489391327
1.9797979593277 1.89577400684357
2.06060600280762 2.0200731754303
2.14141416549683 2.10689377784729
2.22222232818604 2.193035364151
2.30303025245667 2.27852916717529
2.38383841514587 2.36341714859009
2.46464657783508 2.44774556159973
2.54545450210571 2.53156447410583
2.62626266479492 2.61492657661438
2.70707082748413 2.69788289070129
2.78787875175476 2.78048324584961
2.86868691444397 2.86277484893799
2.9494948387146 2.94480085372925
3.03030300140381 3.02660131454468
3.11111116409302 3.10821175575256
3.19191908836365 3.18966341018677
3.27272725105286 3.27098441123962
3.35353541374207 3.35219764709473
3.4343433380127 3.4333233833313
3.5151515007019 3.51437902450562
3.59595966339111 3.59537863731384
3.67676758766174 3.67633318901062
3.75757575035095 3.75725340843201
3.83838391304016 3.8381462097168
3.91919183731079 3.91901755332947
4 3.99987316131592
};
\addplot [semithick, black, dashed]
table {%
-4 -4
-3.91919183731079 -3.91919183731079
-3.83838391304016 -3.83838391304016
-3.75757575035095 -3.75757575035095
-3.67676758766174 -3.67676758766174
-3.59595966339111 -3.59595966339111
-3.5151515007019 -3.5151515007019
-3.4343433380127 -3.4343433380127
-3.35353541374207 -3.35353541374207
-3.27272725105286 -3.27272725105286
-3.19191908836365 -3.19191908836365
-3.11111116409302 -3.11111116409302
-3.03030300140381 -3.03030300140381
-2.9494948387146 -2.9494948387146
-2.86868691444397 -2.86868691444397
-2.78787875175476 -2.78787875175476
-2.70707082748413 -2.70707082748413
-2.62626266479492 -2.62626266479492
-2.54545450210571 -2.54545450210571
-2.46464657783508 -2.46464657783508
-2.38383841514587 -2.38383841514587
-2.30303025245667 -2.30303025245667
-2.22222232818604 -2.22222232818604
-2.14141416549683 -2.14141416549683
-2.06060600280762 -2.06060600280762
-1.9797979593277 -1.9797979593277
-1.89898991584778 -1.89898991584778
-1.81818187236786 -1.81818187236786
-1.73737370967865 -1.73737370967865
-1.65656566619873 -1.65656566619873
-1.57575762271881 -1.57575762271881
-1.4949494600296 -1.4949494600296
-1.41414141654968 -1.41414141654968
-1.33333337306976 -1.33333337306976
-1.25252521038055 -1.25252521038055
-1.17171716690063 -1.17171716690063
-1.09090912342072 -1.09090912342072
-1.01010096073151 -1.01010096073151
-0.929292917251587 -0.929292917251587
-0.848484873771667 -0.848484873771667
-0.767676770687103 -0.767676770687103
-0.686868667602539 -0.686868667602539
-0.60606062412262 -0.60606062412262
-0.525252521038055 -0.525252521038055
-0.444444447755814 -0.444444447755814
-0.363636374473572 -0.363636374473572
-0.282828271389008 -0.282828271389008
-0.202020198106766 -0.202020198106766
-0.121212124824524 -0.121212124824524
-0.0404040440917015 -0.0404040440917015
0.0404040440917015 0.0404040440917015
0.121212124824524 0.121212124824524
0.202020198106766 0.202020198106766
0.282828271389008 0.282828271389008
0.363636374473572 0.363636374473572
0.444444447755814 0.444444447755814
0.525252521038055 0.525252521038055
0.60606062412262 0.60606062412262
0.686868667602539 0.686868667602539
0.767676770687103 0.767676770687103
0.848484873771667 0.848484873771667
0.929292917251587 0.929292917251587
1.01010096073151 1.01010096073151
1.09090912342072 1.09090912342072
1.17171716690063 1.17171716690063
1.25252521038055 1.25252521038055
1.33333337306976 1.33333337306976
1.41414141654968 1.41414141654968
1.4949494600296 1.4949494600296
1.57575762271881 1.57575762271881
1.65656566619873 1.65656566619873
1.73737370967865 1.73737370967865
1.81818187236786 1.81818187236786
1.89898991584778 1.89898991584778
1.9797979593277 1.9797979593277
2.06060600280762 2.06060600280762
2.14141416549683 2.14141416549683
2.22222232818604 2.22222232818604
2.30303025245667 2.30303025245667
2.38383841514587 2.38383841514587
2.46464657783508 2.46464657783508
2.54545450210571 2.54545450210571
2.62626266479492 2.62626266479492
2.70707082748413 2.70707082748413
2.78787875175476 2.78787875175476
2.86868691444397 2.86868691444397
2.9494948387146 2.9494948387146
3.03030300140381 3.03030300140381
3.11111116409302 3.11111116409302
3.19191908836365 3.19191908836365
3.27272725105286 3.27272725105286
3.35353541374207 3.35353541374207
3.4343433380127 3.4343433380127
3.5151515007019 3.5151515007019
3.59595966339111 3.59595966339111
3.67676758766174 3.67676758766174
3.75757575035095 3.75757575035095
3.83838391304016 3.83838391304016
3.91919183731079 3.91919183731079
4 4
};
\end{axis}

\end{tikzpicture}
    } & 
    \hspace{-0.2cm}
    \scalebox{0.45}{
\begin{tikzpicture}

\definecolor{darkgray176}{RGB}{176,176,176}
\definecolor{dodgerblue}{RGB}{30,144,255}

\begin{axis}[
tick align=outside,
tick pos=left,
x grid style={darkgray176},
xlabel={\(\displaystyle x\)},
xmajorgrids,
xmin=-4.4, xmax=4.4,
xtick style={color=black},
grid=both,
grid style={line width=.1pt, draw=gray!10},
major grid style={line width=.2pt,draw=gray!20},
minor x tick num=1,
minor y tick num=2,
y grid style={darkgray176},
ylabel={\(\displaystyle GE-DiTAC(x)\)},
y label style={at={(axis description cs:-0.1,.5)},anchor=south},
ymajorgrids,
ymin=-2, ymax=4,
ytick style={color=black}
]
\addplot [semithick, dodgerblue]
table {%
-4 -0.00012671947479248
-3.91919183731079 -0.00017415032198187
-3.83838391304016 -0.000237708140048198
-3.75757575035095 -0.000322291336487979
-3.67676758766174 -0.000434250541729853
-3.59595966339111 -0.000581064610742033
-3.5151515007019 -0.000772497849538922
-3.4343433380127 -0.00102003407664597
-3.35353541374207 -0.00133783894125372
-3.27272725105286 -0.00174294819589704
-3.19191908836365 -0.00225554686039686
-3.11111116409302 -0.00289930240251124
-3.03030300140381 -0.00370162911713123
-2.9494948387146 -0.00469404365867376
-2.86868691444397 -0.0059120487421751
-2.78787875175476 -0.00739550217986107
-2.70707082748413 -0.00918773841112852
-2.62626266479492 -0.011336050927639
-2.54545450210571 -0.0138899739831686
-2.46464657783508 -0.0169011298567057
-2.38383841514587 -0.0204212907701731
-2.30303025245667 -0.0245009362697601
-2.22222232818604 -0.0291870050132275
-2.14141416549683 -0.0345202684402466
-2.06060600280762 -0.0405327528715134
-1.9797979593277 -0.047244131565094
-1.89898991584778 -0.0546584464609623
-1.81818187236786 -0.0627603307366371
-1.73737370967865 -0.0715113654732704
-1.65656566619873 -0.0808464661240578
-1.57575762271881 -0.0906704068183899
-1.4949494600296 -0.100854992866516
-1.41414141654968 -0.11123663187027
-1.33333337306976 -0.121614933013916
-1.25252521038055 -0.131752207875252
-1.17171716690063 -0.141373917460442
-1.09090912342072 -0.150170683860779
-1.01010096073151 -0.157801449298859
-0.929292917251587 -0.163898140192032
-0.848484873771667 -0.168071269989014
-0.767676770687103 -0.169917285442352
-0.686868667602539 -0.169026538729668
-0.60606062412262 -0.164992272853851
-0.525252521038055 -0.157420203089714
-0.444444447755814 -0.145938068628311
-0.363636374473572 -0.130205377936363
-0.282828271389008 -0.109922401607037
-0.202020198106766 -0.084838479757309
-0.121212124824524 -0.0547589734196663
-0.0404040440917015 -0.0195509307086468
0.0404040440917015 0.0434001572430134
0.121212124824524 0.131259009242058
0.202020198106766 0.216328740119934
0.282828271389008 0.281669765710831
0.363636374473572 0.347807615995407
0.444444447755814 0.405110150575638
0.525252521038055 0.447586923837662
0.60606062412262 0.489527642726898
0.686868667602539 0.527592837810516
0.767676770687103 0.560451626777649
0.848484873771667 0.590790808200836
0.929292917251587 0.6259526014328
1.01010096073151 0.669662952423096
1.09090912342072 0.725427031517029
1.17171716690063 0.796921253204346
1.25252521038055 0.879250049591064
1.33333337306976 0.955206155776978
1.41414141654968 1.02442562580109
1.4949494600296 1.09338939189911
1.57575762271881 1.1639769077301
1.65656566619873 1.23933041095734
1.73737370967865 1.34335339069366
1.81818187236786 1.49541521072388
1.89898991584778 1.70453560352325
1.9797979593277 1.93922472000122
2.06060600280762 2.06060600280762
2.14141416549683 2.14141416549683
2.22222232818604 2.22222232818604
2.30303025245667 2.30303025245667
2.38383841514587 2.38383841514587
2.46464657783508 2.46464657783508
2.54545450210571 2.54545450210571
2.62626266479492 2.62626266479492
2.70707082748413 2.70707082748413
2.78787875175476 2.78787875175476
2.86868691444397 2.86868691444397
2.9494948387146 2.9494948387146
3.03030300140381 3.03030300140381
3.11111116409302 3.11111116409302
3.19191908836365 3.19191908836365
3.27272725105286 3.27272725105286
3.35353541374207 3.35353541374207
3.4343433380127 3.4343433380127
3.5151515007019 3.5151515007019
3.59595966339111 3.59595966339111
3.67676758766174 3.67676758766174
3.75757575035095 3.75757575035095
3.83838391304016 3.83838391304016
3.91919183731079 3.91919183731079
4 4
};
\addplot [semithick, black, dashed]
table {%
-4 -4
-3.91919183731079 -3.91919183731079
-3.83838391304016 -3.83838391304016
-3.75757575035095 -3.75757575035095
-3.67676758766174 -3.67676758766174
-3.59595966339111 -3.59595966339111
-3.5151515007019 -3.5151515007019
-3.4343433380127 -3.4343433380127
-3.35353541374207 -3.35353541374207
-3.27272725105286 -3.27272725105286
-3.19191908836365 -3.19191908836365
-3.11111116409302 -3.11111116409302
-3.03030300140381 -3.03030300140381
-2.9494948387146 -2.9494948387146
-2.86868691444397 -2.86868691444397
-2.78787875175476 -2.78787875175476
-2.70707082748413 -2.70707082748413
-2.62626266479492 -2.62626266479492
-2.54545450210571 -2.54545450210571
-2.46464657783508 -2.46464657783508
-2.38383841514587 -2.38383841514587
-2.30303025245667 -2.30303025245667
-2.22222232818604 -2.22222232818604
-2.14141416549683 -2.14141416549683
-2.06060600280762 -2.06060600280762
-1.9797979593277 -1.9797979593277
-1.89898991584778 -1.89898991584778
-1.81818187236786 -1.81818187236786
-1.73737370967865 -1.73737370967865
-1.65656566619873 -1.65656566619873
-1.57575762271881 -1.57575762271881
-1.4949494600296 -1.4949494600296
-1.41414141654968 -1.41414141654968
-1.33333337306976 -1.33333337306976
-1.25252521038055 -1.25252521038055
-1.17171716690063 -1.17171716690063
-1.09090912342072 -1.09090912342072
-1.01010096073151 -1.01010096073151
-0.929292917251587 -0.929292917251587
-0.848484873771667 -0.848484873771667
-0.767676770687103 -0.767676770687103
-0.686868667602539 -0.686868667602539
-0.60606062412262 -0.60606062412262
-0.525252521038055 -0.525252521038055
-0.444444447755814 -0.444444447755814
-0.363636374473572 -0.363636374473572
-0.282828271389008 -0.282828271389008
-0.202020198106766 -0.202020198106766
-0.121212124824524 -0.121212124824524
-0.0404040440917015 -0.0404040440917015
0.0404040440917015 0.0404040440917015
0.121212124824524 0.121212124824524
0.202020198106766 0.202020198106766
0.282828271389008 0.282828271389008
0.363636374473572 0.363636374473572
0.444444447755814 0.444444447755814
0.525252521038055 0.525252521038055
0.60606062412262 0.60606062412262
0.686868667602539 0.686868667602539
0.767676770687103 0.767676770687103
0.848484873771667 0.848484873771667
0.929292917251587 0.929292917251587
1.01010096073151 1.01010096073151
1.09090912342072 1.09090912342072
1.17171716690063 1.17171716690063
1.25252521038055 1.25252521038055
1.33333337306976 1.33333337306976
1.41414141654968 1.41414141654968
1.4949494600296 1.4949494600296
1.57575762271881 1.57575762271881
1.65656566619873 1.65656566619873
1.73737370967865 1.73737370967865
1.81818187236786 1.81818187236786
1.89898991584778 1.89898991584778
1.9797979593277 1.9797979593277
2.06060600280762 2.06060600280762
2.14141416549683 2.14141416549683
2.22222232818604 2.22222232818604
2.30303025245667 2.30303025245667
2.38383841514587 2.38383841514587
2.46464657783508 2.46464657783508
2.54545450210571 2.54545450210571
2.62626266479492 2.62626266479492
2.70707082748413 2.70707082748413
2.78787875175476 2.78787875175476
2.86868691444397 2.86868691444397
2.9494948387146 2.9494948387146
3.03030300140381 3.03030300140381
3.11111116409302 3.11111116409302
3.19191908836365 3.19191908836365
3.27272725105286 3.27272725105286
3.35353541374207 3.35353541374207
3.4343433380127 3.4343433380127
3.5151515007019 3.5151515007019
3.59595966339111 3.59595966339111
3.67676758766174 3.67676758766174
3.75757575035095 3.75757575035095
3.83838391304016 3.83838391304016
3.91919183731079 3.91919183731079
4 4
};
\end{axis}

\end{tikzpicture}
    }\\ 
    \hspace{0.45cm} \textbf{\small(a)} & \hspace{0.27cm} \textbf{\small(b)} & \hspace{0.38cm} \textbf{\small(c)}\\
    & \hspace{-0.4cm}
    \scalebox{0.45}{
\begin{tikzpicture}

\definecolor{darkgray176}{RGB}{176,176,176}
\definecolor{dodgerblue}{RGB}{30,144,255}

\begin{axis}[
tick align=outside,
tick pos=left,
x grid style={darkgray176},
xlabel={\(\displaystyle x\)},
xmajorgrids,
xmin=-4.4, xmax=4.4,
xtick style={color=black},
grid=both,
grid style={line width=.1pt, draw=gray!10},
major grid style={line width=.2pt,draw=gray!20},
minor x tick num=1,
minor y tick num=1,
y grid style={darkgray176},
ylabel={\(\displaystyle L-DiTAC(x)\)},
y label style={at={(axis description cs:-0.1,.5)},anchor=south},
ymajorgrids,
ymin=-4.4, ymax=4.4,
ytick style={color=black}
]
\addplot [semithick, dodgerblue]
table {%
-4 -2.20000004768372
-3.91919183731079 -2.19191908836365
-3.83838391304016 -2.18383836746216
-3.75757575035095 -2.17575740814209
-3.67676758766174 -2.1676766872406
-3.59595966339111 -2.15959596633911
-3.5151515007019 -2.15151500701904
-3.4343433380127 -2.14343428611755
-3.35353541374207 -2.13535356521606
-3.27272725105286 -2.127272605896
-3.19191908836365 -2.11919188499451
-3.11111116409302 -2.11111116409302
-3.03030300140381 -2.10303020477295
-2.9494948387146 -2.09494948387146
-2.86868691444397 -2.08686876296997
-2.78787875175476 -2.0787878036499
-2.70707082748413 -2.07070708274841
-2.62626266479492 -2.06262612342834
-2.54545450210571 -2.05454540252686
-2.46464657783508 -2.04646468162537
-2.38383841514587 -2.0383837223053
-2.30303025245667 -2.03030300140381
-2.22222232818604 -2.02222228050232
-2.14141416549683 -2.01414132118225
-2.06060600280762 -2.00606060028076
-1.9797979593277 -1.97882914543152
-1.89898991584778 -1.88991177082062
-1.81818187236786 -1.80311143398285
-1.73737370967865 -1.71631121635437
-1.65656566619873 -1.62739372253418
-1.57575762271881 -1.55140566825867
-1.4949494600296 -1.48606467247009
-1.41414141654968 -1.41912996768951
-1.33333337306976 -1.35378885269165
-1.25252521038055 -1.28685426712036
-1.17171716690063 -1.2239830493927
-1.09090912342072 -1.17954432964325
-1.01010096073151 -1.13655579090118
-0.929292917251587 -1.0945907831192
-0.848484873771667 -1.05262577533722
-0.767676770687103 -1.01002311706543
-0.686868667602539 -0.971722960472107
-0.60606062412262 -0.93644654750824
-0.525252521038055 -0.902718663215637
-0.444444447755814 -0.871678233146667
-0.363636374473572 -0.841855049133301
-0.282828271389008 -0.809827923774719
-0.202020198106766 -0.776104807853699
-0.121212124824524 -0.73855459690094
-0.0404040440917015 -0.695604681968689
0.0404040440917015 -0.648711442947388
0.121212124824524 -0.594214797019958
0.202020198106766 -0.533642649650574
0.282828271389008 -0.464668750762939
0.363636374473572 -0.384121775627136
0.444444447755814 -0.301730036735535
0.525252521038055 -0.222499489784241
0.60606062412262 -0.144830226898193
0.686868667602539 -0.0721246004104614
0.767676770687103 -0.00223791599273682
0.848484873771667 0.0669522285461426
0.929292917251587 0.134894847869873
1.01010096073151 0.203611135482788
1.09090912342072 0.27483081817627
1.17171716690063 0.345191955566406
1.25252521038055 0.41761040687561
1.33333337306976 0.502147197723389
1.41414141654968 0.599829435348511
1.4949494600296 0.720865726470947
1.57575762271881 0.86599588394165
1.65656566619873 1.03449749946594
1.73737370967865 1.24153232574463
1.81818187236786 1.46442604064941
1.89898991584778 1.69901752471924
1.9797979593277 1.94211888313293
2.06060600280762 2.06060600280762
2.14141416549683 2.14141416549683
2.22222232818604 2.22222232818604
2.30303025245667 2.30303025245667
2.38383841514587 2.38383841514587
2.46464657783508 2.46464657783508
2.54545450210571 2.54545450210571
2.62626266479492 2.62626266479492
2.70707082748413 2.70707082748413
2.78787875175476 2.78787875175476
2.86868691444397 2.86868691444397
2.9494948387146 2.9494948387146
3.03030300140381 3.03030300140381
3.11111116409302 3.11111116409302
3.19191908836365 3.19191908836365
3.27272725105286 3.27272725105286
3.35353541374207 3.35353541374207
3.4343433380127 3.4343433380127
3.5151515007019 3.5151515007019
3.59595966339111 3.59595966339111
3.67676758766174 3.67676758766174
3.75757575035095 3.75757575035095
3.83838391304016 3.83838391304016
3.91919183731079 3.91919183731079
4 4
};
\addplot [semithick, black, dashed]
table {%
-4 -4
-3.91919183731079 -3.91919183731079
-3.83838391304016 -3.83838391304016
-3.75757575035095 -3.75757575035095
-3.67676758766174 -3.67676758766174
-3.59595966339111 -3.59595966339111
-3.5151515007019 -3.5151515007019
-3.4343433380127 -3.4343433380127
-3.35353541374207 -3.35353541374207
-3.27272725105286 -3.27272725105286
-3.19191908836365 -3.19191908836365
-3.11111116409302 -3.11111116409302
-3.03030300140381 -3.03030300140381
-2.9494948387146 -2.9494948387146
-2.86868691444397 -2.86868691444397
-2.78787875175476 -2.78787875175476
-2.70707082748413 -2.70707082748413
-2.62626266479492 -2.62626266479492
-2.54545450210571 -2.54545450210571
-2.46464657783508 -2.46464657783508
-2.38383841514587 -2.38383841514587
-2.30303025245667 -2.30303025245667
-2.22222232818604 -2.22222232818604
-2.14141416549683 -2.14141416549683
-2.06060600280762 -2.06060600280762
-1.9797979593277 -1.9797979593277
-1.89898991584778 -1.89898991584778
-1.81818187236786 -1.81818187236786
-1.73737370967865 -1.73737370967865
-1.65656566619873 -1.65656566619873
-1.57575762271881 -1.57575762271881
-1.4949494600296 -1.4949494600296
-1.41414141654968 -1.41414141654968
-1.33333337306976 -1.33333337306976
-1.25252521038055 -1.25252521038055
-1.17171716690063 -1.17171716690063
-1.09090912342072 -1.09090912342072
-1.01010096073151 -1.01010096073151
-0.929292917251587 -0.929292917251587
-0.848484873771667 -0.848484873771667
-0.767676770687103 -0.767676770687103
-0.686868667602539 -0.686868667602539
-0.60606062412262 -0.60606062412262
-0.525252521038055 -0.525252521038055
-0.444444447755814 -0.444444447755814
-0.363636374473572 -0.363636374473572
-0.282828271389008 -0.282828271389008
-0.202020198106766 -0.202020198106766
-0.121212124824524 -0.121212124824524
-0.0404040440917015 -0.0404040440917015
0.0404040440917015 0.0404040440917015
0.121212124824524 0.121212124824524
0.202020198106766 0.202020198106766
0.282828271389008 0.282828271389008
0.363636374473572 0.363636374473572
0.444444447755814 0.444444447755814
0.525252521038055 0.525252521038055
0.60606062412262 0.60606062412262
0.686868667602539 0.686868667602539
0.767676770687103 0.767676770687103
0.848484873771667 0.848484873771667
0.929292917251587 0.929292917251587
1.01010096073151 1.01010096073151
1.09090912342072 1.09090912342072
1.17171716690063 1.17171716690063
1.25252521038055 1.25252521038055
1.33333337306976 1.33333337306976
1.41414141654968 1.41414141654968
1.4949494600296 1.4949494600296
1.57575762271881 1.57575762271881
1.65656566619873 1.65656566619873
1.73737370967865 1.73737370967865
1.81818187236786 1.81818187236786
1.89898991584778 1.89898991584778
1.9797979593277 1.9797979593277
2.06060600280762 2.06060600280762
2.14141416549683 2.14141416549683
2.22222232818604 2.22222232818604
2.30303025245667 2.30303025245667
2.38383841514587 2.38383841514587
2.46464657783508 2.46464657783508
2.54545450210571 2.54545450210571
2.62626266479492 2.62626266479492
2.70707082748413 2.70707082748413
2.78787875175476 2.78787875175476
2.86868691444397 2.86868691444397
2.9494948387146 2.9494948387146
3.03030300140381 3.03030300140381
3.11111116409302 3.11111116409302
3.19191908836365 3.19191908836365
3.27272725105286 3.27272725105286
3.35353541374207 3.35353541374207
3.4343433380127 3.4343433380127
3.5151515007019 3.5151515007019
3.59595966339111 3.59595966339111
3.67676758766174 3.67676758766174
3.75757575035095 3.75757575035095
3.83838391304016 3.83838391304016
3.91919183731079 3.91919183731079
4 4
};
\end{axis}

\end{tikzpicture}
    } & 
    \hspace{-0.09cm}
    \scalebox{0.45}{
\begin{tikzpicture}

\definecolor{darkgray176}{RGB}{176,176,176}
\definecolor{dodgerblue}{RGB}{30,144,255}

\begin{axis}[
tick align=outside,
tick pos=left,
x grid style={darkgray176},
xlabel={\(\displaystyle x\)},
xmajorgrids,
xmin=-2, xmax=4,
xtick style={color=black},
grid=both,
grid style={line width=.1pt, draw=gray!10},
major grid style={line width=.2pt,draw=gray!20},
minor x tick num=1,
minor y tick num=2,
y grid style={darkgray176},
ylabel={\(\displaystyle inf-DiTAC(x)\)},
y label style={at={(axis description cs:-0.1,.5)},anchor=south},
ymajorgrids,
ymin=-2, ymax=4,
ytick style={color=black}
]
\addplot [semithick, dodgerblue]
table {%
-4 -4.33366441726685
-3.91919183731079 -4.24611568450928
-3.83838391304016 -4.15856695175171
-3.75757575035095 -4.07101821899414
-3.67676758766174 -3.98346924781799
-3.59595966339111 -3.89592051506042
-3.5151515007019 -3.80837178230286
-3.4343433380127 -3.72082281112671
-3.35353541374207 -3.63327431678772
-3.27272725105286 -3.54572534561157
-3.19191908836365 -3.458176612854
-3.11111116409302 -3.37062788009644
-3.03030300140381 -3.28307914733887
-2.9494948387146 -3.19553017616272
-2.86868691444397 -3.10798168182373
-2.78787875175476 -3.02043271064758
-2.70707082748413 -2.93288421630859
-2.62626266479492 -2.84533524513245
-2.54545450210571 -2.75778651237488
-2.46464657783508 -2.67023777961731
-2.38383841514587 -2.58268904685974
-2.30303025245667 -2.49514007568359
-2.22222232818604 -2.40759134292603
-2.14141416549683 -2.32004261016846
-2.06060600280762 -2.23249363899231
-1.9797979593277 -2.14494490623474
-1.89898991584778 -2.05739617347717
-1.81818187236786 -1.96984755992889
-1.73737370967865 -1.88229870796204
-1.65656566619873 -1.79474997520447
-1.57575762271881 -1.7072012424469
-1.4949494600296 -1.61965227127075
-1.41414141654968 -1.53210353851318
-1.33333337306976 -1.44455480575562
-1.25252521038055 -1.35700595378876
-1.17171716690063 -1.26945722103119
-1.09090912342072 -1.18190848827362
-1.01010096073151 -1.09435963630676
-0.929292917251587 -1.00681090354919
-0.848484873771667 -0.919262170791626
-0.767676770687103 -0.831713378429413
-0.686868667602539 -0.7441645860672
-0.60606062412262 -0.656615853309631
-0.525252521038055 -0.569067060947418
-0.444444447755814 -0.481518268585205
-0.363636374473572 -0.393969506025314
-0.282828271389008 -0.306420713663101
-0.202020198106766 -0.218871936202049
-0.121212124824524 -0.131323173642159
-0.0404040440917015 -0.0437743924558163
0.0404040440917015 0.0437743924558163
0.121212124824524 0.12454792112112
0.202020198106766 0.19039611518383
0.282828271389008 0.234271794557571
0.363636374473572 0.272235155105591
0.444444447755814 0.303788930177689
0.525252521038055 0.347768366336823
0.60606062412262 0.419402360916138
0.686868667602539 0.495249658823013
0.767676770687103 0.564208507537842
0.848484873771667 0.643500506877899
0.929292917251587 0.796203970909119
1.01010096073151 1.02991986274719
1.09090912342072 1.26927947998047
1.17171716690063 1.50863885879517
1.25252521038055 1.74799799919128
1.33333337306976 1.98735761642456
1.41414141654968 2.22671699523926
1.4949494600296 2.46607637405396
1.57575762271881 2.70543622970581
1.65656566619873 2.94479513168335
1.73737370967865 3.18415451049805
1.81818187236786 3.4235143661499
1.89898991584778 3.6628737449646
1.9797979593277 3.9022331237793
2.06060600280762 4.14159202575684
2.14141416549683 4.38095188140869
2.22222232818604 4.62031173706055
2.30303025245667 4.85967063903809
2.38383841514587 5.09903001785278
2.46464657783508 5.33838987350464
2.54545450210571 5.57774877548218
2.62626266479492 5.81710863113403
2.70707082748413 6.05646848678589
2.78787875175476 6.29582738876343
2.86868691444397 6.53518724441528
2.9494948387146 6.77454614639282
3.03030300140381 7.01390504837036
3.11111116409302 7.25326490402222
3.19191908836365 7.49262380599976
3.27272725105286 7.73198366165161
3.35353541374207 7.97134351730347
3.4343433380127 8.21070289611816
3.5151515007019 8.4500617980957
3.59595966339111 8.68942070007324
3.67676758766174 8.92878150939941
3.75757575035095 9.16814041137695
3.83838391304016 9.40749931335449
3.91919183731079 9.64685821533203
4 9.8862190246582
};
\addplot [semithick, black, dashed]
table {%
-4 -4
-3.91919183731079 -3.91919183731079
-3.83838391304016 -3.83838391304016
-3.75757575035095 -3.75757575035095
-3.67676758766174 -3.67676758766174
-3.59595966339111 -3.59595966339111
-3.5151515007019 -3.5151515007019
-3.4343433380127 -3.4343433380127
-3.35353541374207 -3.35353541374207
-3.27272725105286 -3.27272725105286
-3.19191908836365 -3.19191908836365
-3.11111116409302 -3.11111116409302
-3.03030300140381 -3.03030300140381
-2.9494948387146 -2.9494948387146
-2.86868691444397 -2.86868691444397
-2.78787875175476 -2.78787875175476
-2.70707082748413 -2.70707082748413
-2.62626266479492 -2.62626266479492
-2.54545450210571 -2.54545450210571
-2.46464657783508 -2.46464657783508
-2.38383841514587 -2.38383841514587
-2.30303025245667 -2.30303025245667
-2.22222232818604 -2.22222232818604
-2.14141416549683 -2.14141416549683
-2.06060600280762 -2.06060600280762
-1.9797979593277 -1.9797979593277
-1.89898991584778 -1.89898991584778
-1.81818187236786 -1.81818187236786
-1.73737370967865 -1.73737370967865
-1.65656566619873 -1.65656566619873
-1.57575762271881 -1.57575762271881
-1.4949494600296 -1.4949494600296
-1.41414141654968 -1.41414141654968
-1.33333337306976 -1.33333337306976
-1.25252521038055 -1.25252521038055
-1.17171716690063 -1.17171716690063
-1.09090912342072 -1.09090912342072
-1.01010096073151 -1.01010096073151
-0.929292917251587 -0.929292917251587
-0.848484873771667 -0.848484873771667
-0.767676770687103 -0.767676770687103
-0.686868667602539 -0.686868667602539
-0.60606062412262 -0.60606062412262
-0.525252521038055 -0.525252521038055
-0.444444447755814 -0.444444447755814
-0.363636374473572 -0.363636374473572
-0.282828271389008 -0.282828271389008
-0.202020198106766 -0.202020198106766
-0.121212124824524 -0.121212124824524
-0.0404040440917015 -0.0404040440917015
0.0404040440917015 0.0404040440917015
0.121212124824524 0.121212124824524
0.202020198106766 0.202020198106766
0.282828271389008 0.282828271389008
0.363636374473572 0.363636374473572
0.444444447755814 0.444444447755814
0.525252521038055 0.525252521038055
0.60606062412262 0.60606062412262
0.686868667602539 0.686868667602539
0.767676770687103 0.767676770687103
0.848484873771667 0.848484873771667
0.929292917251587 0.929292917251587
1.01010096073151 1.01010096073151
1.09090912342072 1.09090912342072
1.17171716690063 1.17171716690063
1.25252521038055 1.25252521038055
1.33333337306976 1.33333337306976
1.41414141654968 1.41414141654968
1.4949494600296 1.4949494600296
1.57575762271881 1.57575762271881
1.65656566619873 1.65656566619873
1.73737370967865 1.73737370967865
1.81818187236786 1.81818187236786
1.89898991584778 1.89898991584778
1.9797979593277 1.9797979593277
2.06060600280762 2.06060600280762
2.14141416549683 2.14141416549683
2.22222232818604 2.22222232818604
2.30303025245667 2.30303025245667
2.38383841514587 2.38383841514587
2.46464657783508 2.46464657783508
2.54545450210571 2.54545450210571
2.62626266479492 2.62626266479492
2.70707082748413 2.70707082748413
2.78787875175476 2.78787875175476
2.86868691444397 2.86868691444397
2.9494948387146 2.9494948387146
3.03030300140381 3.03030300140381
3.11111116409302 3.11111116409302
3.19191908836365 3.19191908836365
3.27272725105286 3.27272725105286
3.35353541374207 3.35353541374207
3.4343433380127 3.4343433380127
3.5151515007019 3.5151515007019
3.59595966339111 3.59595966339111
3.67676758766174 3.67676758766174
3.75757575035095 3.75757575035095
3.83838391304016 3.83838391304016
3.91919183731079 3.91919183731079
4 4
};
\end{axis}

\end{tikzpicture}
    }\\ 
    & \hspace{0.27cm} \textbf{\small(d)} & \hspace{0.38cm} \textbf{\small(e)}\\
    \end{tabular}
    \caption{Illustration of DiTAC versions. We display (a) $v^{\btheta}$, the CPA velocity field used in all DiTAC version illustrations, (b) DiTAC, (c) GE-DiTAC, (d) L-DiTAC, and (e) inf-DiTAC}    
    \label{fig:ditac_versions}
\end{figure}

In the paper, in section 3.2, we present how DiTAC is built. 
DiTAC uses a CPAB transformation $T^\btheta$, which is defined on a finite interval, $\Omega=[a,b]\subset \RR$, and its co-domain is also a finite interval. In order to handle input data that fall outside of $[a,b]$, we combine $T^\btheta$ with GELU, a recent widely-used AF in state-of-the-art models. Other versions of DiTAC can also be built by combining $T^\btheta$ with a variety of other AFs, or defining a certain function, not necessarily a known AF, outside of CPAB's map. Here we present additional DiTAC versions (definitions and illustrations), share our insights about their quality, and evaluate their performance on a simple regression task. All DiTAC versions are illustrated in \autoref{fig:ditac_versions}.

We note, that conceptually it is possible to apply CPAB transformation on the entire input data by first apply a normalization that maps the data into $\Omega$, perform the CPAB transformation, and then rescale the transformed data back to its original range. However, we will then have to extract the minimum and maximum values of the entire input data, something that is hard to achieve during training as these values depend on the network's parameters learning. Therefore, we set the $[a,b]$-interval before training, which usually includes a large portion of the data, and apply a different behavior on the data that is outside of that range.

\subsubsection{GELU-like DiTAC (DiTAC).} 
This is the main DiTAC version, which is also used in all of our experiments. GELU is a natural choice given its popularity and success in state-of-the-art architectures. Input data that fall outside of the $[a,b]$-interval inherit GELU's behavior, while the input data inside $[a,b]$ first go through CPAB transformation and then go through the GELU function.
The GELU-like DiTAC is defined as follows:
\begin{align}
    \mathrm{DiTAC}(x) = \Tilde{x}\cdot\Phi(x)\,,
   \qquad
      \Tilde{x}   =    \begin{cases}
       T^\btheta(x) & \quad \text{If } a \leq x \leq b\\
       x & \text{Otherwise}\\
     \end{cases}\, 
\end{align}
where $\Phi$ is the cumulative distribution function (CDF) of a standard normal distribution, $T^\btheta$ is a CPAB transformation and $\Omega=[a,b]$, the domain of $T^\btheta$, is user-defined. \\

\subsubsection{GELU-DiTAC (GE-DiTAC).} 
This activation is similar to DiTAC's main version, except that here we apply GELU only on negative input values, whereas a pure CPAB transformation is performed on the input-data range $[0, b]$. In order to keep the function continuous
(in case we impose zero-boundary conditions
on $\bv^\btheta$; see~\cite{Freifeld:PAMI:2017:CPAB}),
we apply the identity function on values that are larger than $b$. The GE-DiTAC is defined as follows:
\begin{align}
      \mathrm{GE-DiTAC}(x)   =    \begin{cases}
                   x\cdot\Phi(x) & \quad \text{If } x < 0 \\
                   T^\btheta(x) & \quad \text{If } 0 \leq x \leq b\\
                   x & \quad \text{If } x > b\\
     \end{cases}\, 
\end{align}

where $\Phi$ is the CDF of a standard normal distribution, $T^\btheta$ is a CPAB transformation and $\Omega=[0,b]$, the domain of $T^\btheta$, is user-defined. \\
Note that GE-DiTAC allows CPAB transformation's capability to be more evident, as this part of the transformed data is not composed with any other function. Empirically, it performs similar to DiTAC in most of our experiments, while its advantage is mainly demonstrated in simple regression tasks using simple networks.

\subsubsection{Leaky DiTAC (L-DiTAC).} 
Here $T^\btheta$ is applied on $[a,b]$ while the rest of the data goes through a Leaky-ReLU (LReLU) function. That is, 
\begin{align}	  
\mathrm{Leaky \ DiTAC}(x) = 
     \begin{cases}
        T^\btheta(x) & \quad \text{If } a \leq x \leq b\\ \mathrm{LReLU}(x) & \text{Otherwise}\\
     \end{cases}
\end{align}
where $T^\btheta$ is a CPAB transformation and $\Omega[a,b]$, the domain of $T^\btheta$, is user-defined. This version can be  as an expressive version of ReLU. 
As shown in~\cite{Dubey:2022:survey_AFs_1}, different AFs are suitable to different types of data and tasks.
This DiTAC version may improve problems in which the ReLU function performs better than any other existing AF.\\

\subsubsection{Infinite-edges DiTAC (inf-DiTAC).} 
Recall that CPAB transformations are obtained via an integration of elements in $\Vcal$, which is a space of continuous functions from $\Omega$ to $\RR$, that are piecewise-affine w.r.t. some fixed partition of $\Omega$ into sub-intervals.
In inf-DiTAC, similarly to GE-DiTAC and L-DiTAC, $T^\btheta$ is applied on $[a,b]$-interval. As for the input data that fall outside of that range, we apply the affine transformations learned in both edges of $\Omega$'s tessellation (most-right and most-left cells), yielding a continuous AF which is controlled solely by CPAB transformation parameters. The inf-DiTAC is defined as follows:
\begin{align}	  
\mathrm{inf-DiTAC}(x) = 
     \begin{cases}
        A_{l}^\btheta x &  \quad \text{If } x < a \\
        T^\btheta(x) & \quad \text{If } a \leq x \leq b \\ 
        A_{r}^\btheta x & \quad \text{If } x > b \\
     \end{cases}
\end{align}
where $T^\btheta$ is a CPAB transformation, $A_{l}^\btheta$ and $A_{r}^\btheta$ are the affine transformations in the most-left and most-right cells of the tessellation respectively, and $\Omega[a,b]$, the domain of $T^\btheta$, is user-defined.\\

\begin{table}[hbt!] 
    \centering
    \setlength{\tabcolsep}{5pt}
    \caption{Regression-task results of learning a two-dimensional target function, using various DiTAC versions on a simple MLP.}
    \begin{tabular}{lcc}
         \toprule
         DiTAC Version & MSE $\downarrow$ & R2 $\uparrow$ \\
         \midrule
         DiTAC & 0.004 & 98.4  \\
         GE-DiTAC & 0.001 & 99.4 \\
         L-DiTAC & 0.006 & 98.7 \\
         inf-DiTAC & 0.004 & 97.6 \\
         \bottomrule
    \end{tabular}
    \label{tab:ditac_versions}
\end{table}

In \autoref{tab:ditac_versions} we show performance evaluation of all aforementioned DiTAC versions, on the two-dimensional function-reconstruction task presented in the paper (training details are provided in~\autoref{Sec:Regression:Subsec:func}). It can be shown that GE-DiTAC provides the best performance on this specific task.


\section{Computational Cost} \label{submat:computational_cost}

In section 3.3 in the paper we extensively describe DiTAC's computational cost, and the use of a lookup table in both training and inference phases.
In \autoref{tab:rebuttal_flops} we measure the computational cost by comparing the \# of parameters, FLOPs and latency during inference, along with top-1 accuracy on ImageNet-1K, where DiTAC is integrated in two model types: mobile-oriented (\eg MobileNet-V3), and massive architectures (\eg ConvNeXT-S).
Evidently, DiTAC consistently improves accuracy with a moderate increase in latency, and no significant change in FLOPs or Params.
\begin{table}[hbt!]  
    \centering
    \setlength{\tabcolsep}{5pt}
    \caption{Computational cost/accuracy (on NVIDIA Tesla P100 GPU, batch size of 1). Models using DiTAC are marked with $^+$.}
    \small
    \begin{tabular}{lcccc}
         \toprule
         Configuration & Params. & FLOPs & Latency [sec] & Top-1 \\
         \midrule
         MobileNet-V3 & 1.57M & 57.2M & 0.009$\pm$0.004  & 61.1 \\
         MobileNet-V3$^+$ & 1.57M & 60.4M & 0.016$\pm$0.001  & \textbf{61.3} \\
         \midrule
         ConvNeXT-S & 49.5M  & 8.6B & 0.016$\pm$0.002  & 83.1 \\
         ConvNeXT-S$^+$ & 49.5M  & 8.7B & 0.027$\pm$ 0.001 & \textbf{83.3} \\
         \bottomrule
    \end{tabular}
    \label{tab:rebuttal_flops}
\end{table}

\end{document}